\documentclass[10pt,twocolumn]{article}

\usepackage{natbib}
\usepackage{xurl}      
\usepackage{hyperref}  
\usepackage{comment}
\usepackage{tikz}
\usepackage{amssymb}
\usepackage{multicol}
\usepackage{caption}           
\usepackage{subcaption} 
\usepackage{amsmath, amsthm} 
\usepackage{mathtools} 
\DeclareMathAlphabet{\mathbbold}{U}{bbold}{m}{n}
\usepackage{siunitx}
\usepackage{booktabs}
\usepackage{multirow}

\newcommand{\Binom}[2]{\binom{#1}{#2}}
\newcommand{\M}{\,{}_1F_1}

\usepackage{array}    
\usepackage{graphicx} 
\usepackage[margin=1in]{geometry}

\usetikzlibrary{positioning}
\usetikzlibrary{backgrounds}
\newcommand{\yesSymbol}{\tikz[scale=0.2] \draw[green!80!black, thick] (0,-0.1) -- (0.2,-0.4) -- (0.6,0.2);}
\newcommand{\noSymbol}{\tikz[scale=0.2] \draw[red!80!black, thick] (0,0) -- (0.6,-0.6) (0,-0.6) -- (0.6,0);}

\tikzset{every picture/.style={every node/.style={font=\small}}}

\tikzset{infoBox/.style={
    draw, rounded corners=2pt,
    inner sep=5pt, outer sep=0pt, text width=3.2cm, align=center
  }}

\tikzset{nodrawbox/.style={
    rounded corners=2pt,
    inner sep=3pt, text width=3.2cm, align=center
  }}

\newcolumntype{C}[1]{>{\centering\arraybackslash}m{#1}}

\newtheorem{theorem}{Theorem}[section]
\newtheorem{corollary}[theorem]{Corollary}
\newtheorem{proposition}[theorem]{Proposition}

\title{Fisher-Bingham-like normalizing flows on the sphere}
\author{Thorsten Glüsenkamp\thanks{thorsten.glusenkamp@fysik.su.se}}

\begin{document}
\maketitle

\begin{abstract}
A generic D-dimensional Gaussian can be conditioned or projected onto the D-1 unit sphere, thereby leading to the well-known
Fisher-Bingham (FB) or Angular Gaussian (AG) distribution families, respectively.
These are some of the most fundamental distributions on the sphere, yet cannot straightforwardly be written as a normalizing flow except in two special cases: the von-Mises Fisher in D=3 and the central angular Gaussian in any D. In this paper, we describe how to generalize these special cases to a family of normalizing flows that behave similarly to the full FB or AG family in any D. We call them "zoom-linear-project" (ZLP)-Fisher flows. Unlike a normal Fisher-Bingham distribution, their composition allows to gradually add complexity as needed. Furthermore, they can naturally handle conditional density estimation with target distributions that vary by orders of magnitude in scale - a setting that is important in astronomical applications but that existing flows often struggle with. A particularly useful member of the new family is the Kent analogue that can cheaply upgrade any flow in this situation to yield better performance.
\end{abstract}

\section{Introduction}

Normalizing flows \cite{flows_paper} and in particular conditional normalizing flows are an emerging new reconstruction tool in statistical inference. In contrast to their usage purely as a generative model \cite{glow_paper} in high dimensions, they are also exceedingly used in variational inference \cite{vae_paper} \cite{hyperspherical_vaes_paper} or in neural posterior estimation (NPE) \cite{npe_paper} in lower-dimensional setups.
In the context of directional posterior estimation it can be vital to use distributions defined naturally on the 2-sphere which fall into the sub-category of manifold normalizing flows \cite{tori_and_spheres_paper}. In the context of NPE these manifold normalizing flows have to be efficiently made conditional and stay numerically stable. In particular, we are interested in situations where the involved posteriors vary orders of magnitude in scale. Such situations can appear, for example, in gravitational-wave or neutrino astronomy.

\begin{figure*}[htb!]
\begin{subfigure}{0.45\linewidth}
\centering
\begin{tikzpicture}

\node[] (img2) 
{\includegraphics[width=2cm]{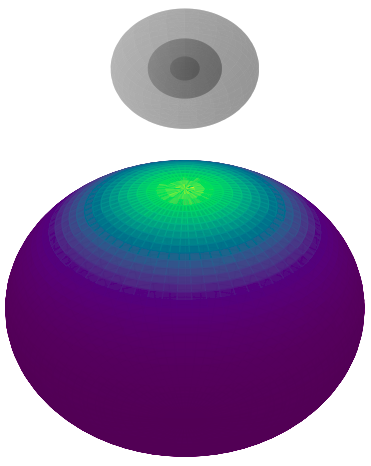}};

\node[infoBox, below left=0.5cm and -1cm of img2] (tab3) {
    \textbf{von-Mises-Fisher}\\
    \rule{\linewidth}{0.4pt}\\
    simple functions \yesSymbol\\
    normalizing flow \yesSymbol \\
    ("Fisher zoom" $\Phi_\mathrm{Z}$)
  };

\node[nodrawbox, below right=0.5cm and -1.0cm of img2] (tab4) {
    \textbf{AG (offset/symm.)}\\
    \rule{\linewidth}{0.4pt}\\
    simple functions \yesSymbol\\
    normalizing flow \noSymbol
  };

\draw[->] (img2) -- (tab3.north) node[above left, xshift=0.5cm, yshift=0.1cm] {\begin{minipage}{2cm}
                \centering
                conditional\\
                on $S^2$
        \end{minipage}};
\draw[->] (img2) -- (tab4.north) node[above right,xshift=-0.5cm,yshift=0.1cm] {\begin{minipage}{2cm}
                \centering
                marginal\\
                on $S^2$
        \end{minipage}};
\end{tikzpicture}
\caption{}
\label{fig:fvm_as_conditional}
\end{subfigure}
\centering
\begin{subfigure}[b]{0.45\linewidth}
\centering
\begin{tikzpicture}

\node[] (img1) {\includegraphics[width=3cm]{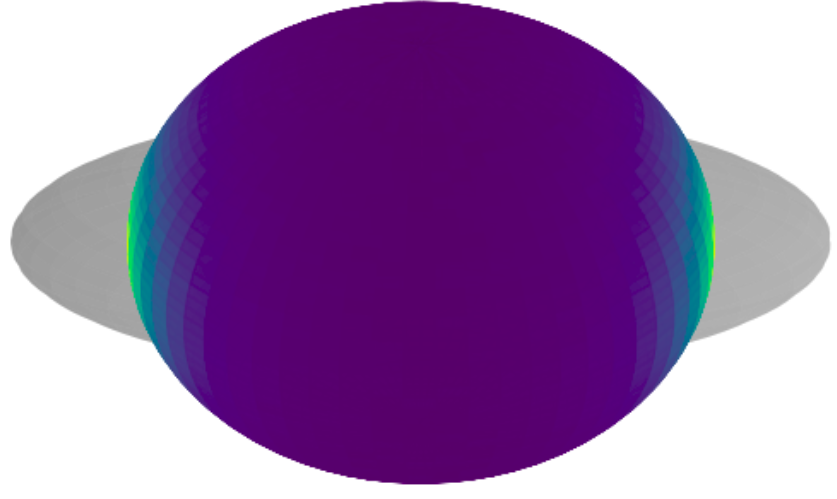}};

\node[nodrawbox, below left=0.5cm and -1.5cm of img1] (tab1) {
    \textbf{ Bingham/Watson}\\[-0.2em]
    \rule{\linewidth}{0.4pt}\\[0.4em]
    simple functions  \noSymbol\\
    normalizing flow  \noSymbol 
   
  };

\node[infoBox, below right=0.5cm and -1.5cm of img1,text width=3.6cm] (tab2) {
    \textbf{AG (central/asymm.)}\\[-0.2em]
    \rule{\linewidth}{0.4pt}\\[0.4em]
    simple functions \yesSymbol\\
    normalizing flow \yesSymbol \\
    ("linear project" $\Phi_{\mathrm{LP}}$)
  };

\draw[->] (img1) -- (tab1.north) node[above left, xshift=0.5cm, yshift=0.1cm] {\begin{minipage}{2cm}
                \centering
                conditional\\
                on $S^2$
        \end{minipage}};
\draw[->] (img1) -- (tab2.north) node[above right,xshift=-0.5cm,yshift=0.1cm] {\begin{minipage}{2cm}
                \centering
                marginal\\
                on $S^2$
        \end{minipage}};
\end{tikzpicture}
\caption{}
\end{subfigure}
\begin{subfigure}{0.45\linewidth}
\centering
\begin{tikzpicture}
\node[] (img3) {\includegraphics[width=2cm]{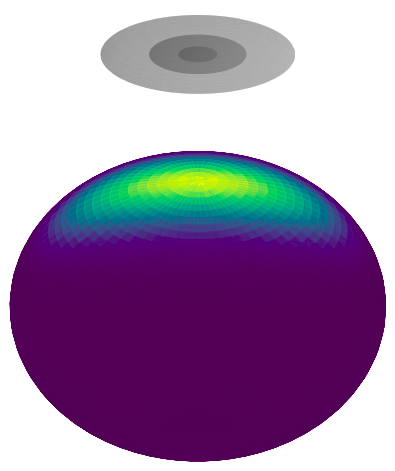}};
\node[nodrawbox, below left=0.5cm and -1cm of img3] (tab5) {
    \textbf{Kent ($\mathrm{FB}_5$)}\\[-0.2em]
    \rule{\linewidth}{0.4pt}\\[0.4em]
    simple functions  \noSymbol\\
    normalizing flow  \noSymbol 
   
  };
\node[nodrawbox, below right=0.5cm and -1cm of img3, text width=3.3cm] (tab6) {
    \textbf{AG (offset/asymm.)}\\[-0.2em]
    \rule{\linewidth}{0.4pt}\\[0.4em]
    simple functions \yesSymbol\\
    normalizing flow \noSymbol
  };

\draw[->] (img3) -- (tab5.north) node[above left, xshift=0.5cm, yshift=0.1cm] {\begin{minipage}{2cm}
                \centering
                conditional\\
                on $S^2$
        \end{minipage}};
\draw[->] (img3) -- (tab6.north) node[above right,xshift=-0.5cm,yshift=0.1cm] {\begin{minipage}{2cm}
                \centering
                marginal\\
                on $S^2$
        \end{minipage}};

\end{tikzpicture}
\caption{}
\end{subfigure}
\centering
\begin{subfigure}{0.45\linewidth}
\centering
\begin{tikzpicture}
    \node (graphic1) at (0, 2) {\includegraphics[width=1.5cm]{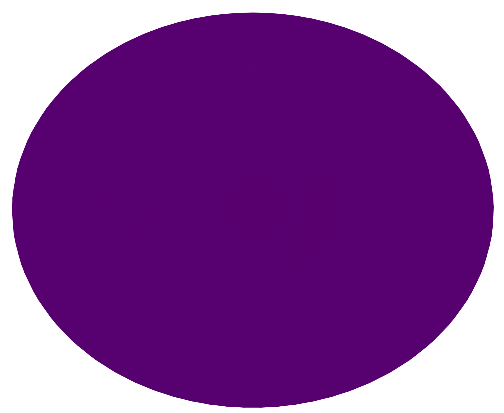}}; 
    
    \node (graphic2) at (3.2, 2) {\includegraphics[width=1.5cm]{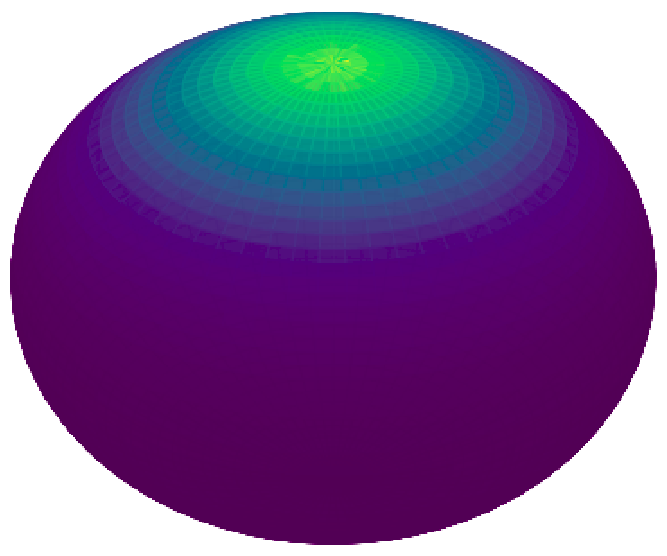}}; 
    
    \node (graphic3) at (6.4, 2) {\includegraphics[width=1.5cm]{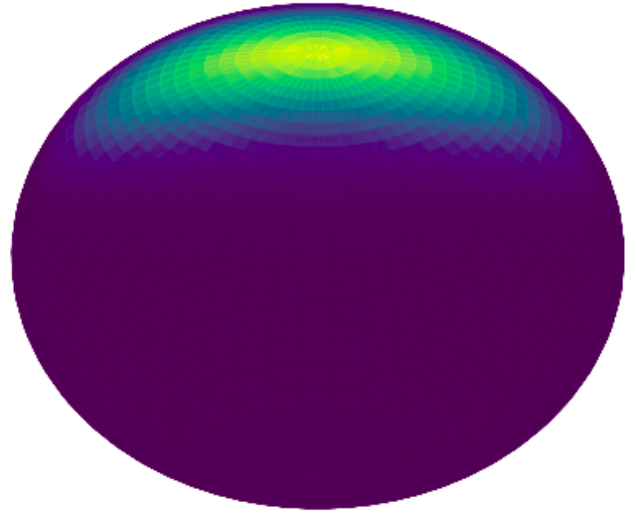}};
    

    \draw[->]
    (graphic1.south east) 
    to[out=-45,in=-135]
      node[midway,below,yshift=-2pt]{\begin{minipage}{4cm}
                \centering
                Fisher zoom\\
               $\Phi_\mathrm{Z}$
        \end{minipage}}
    (graphic2.south west);
    
    \draw[->]
    (graphic2.south east) 
    to[out=-45,in=-135]
      node[midway,below,yshift=-2pt]{\begin{minipage}{4cm}
                \centering
                linear-project\\
               $\Phi_\mathrm{LP,S_c}$
        \end{minipage}}
    (graphic3.south west);

    \draw[->]
    (graphic1.south) 
    to[out=-85,in=-91]
node[infoBox,midway,below,yshift=-2pt,text width=3.5cm]{
       \textbf{Kent-like flow}\\[-0.2em]
    \rule{\linewidth}{0.4pt}\\[0.4em]
      $\Phi_K=\Phi_\mathrm{LP,S_c} \circ \Phi_\mathrm{Z}$ (with constraints)}
    (graphic3.south);
    
\end{tikzpicture}
\caption{}
\end{subfigure}
\caption{Results of specific 3-d Gaussians (contours illustrated by gray ellipsoids) conditioned or marginalized onto the surface of the 2-sphere (a-c). These yield members of the Fisher-Bingham (FB) or angular Gaussian (AG) family. For generic Gaussians (c), no normalizing-flow description exists. Picking the "Fisher zoom" from a) and the "linear-project" from b) with specific parameter constraints ($\Phi_\mathrm{LP,S_c}$, see section \ref{section:combining_flows}) leads to a dynamical normalizing-flow construction of a density with similar properties as the Kent distribution (d) - a bivariate unimodal distribution with a Gaussian limit in tangent space for large concentrations.}
\label{fig:nf_overview}
\end{figure*}
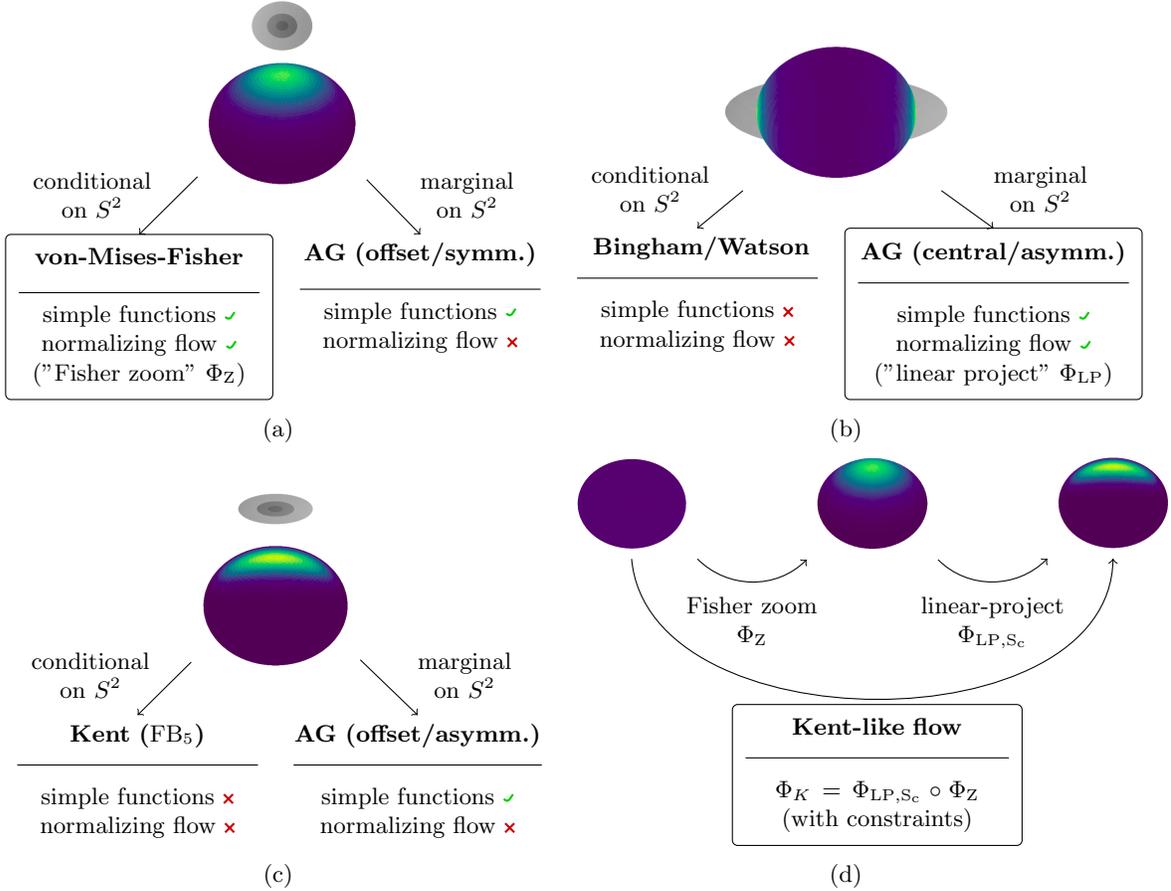

In the following, we discuss a strategy to construct arguably the "simplest" normalizing flows on the sphere that have so far have been neglected in the literature: spherical analogues of the multivariate Normal distribution. In order to do so, we first give an overview of the two families of distributions of this kind that are known. The first such distribution is the \textbf{F}isher-\textbf{B}ingham (FB) family \cite{mardia_fb8_1975} \cite{fb5_paper} which is obtained by conditioning a generic Gaussian distribution in embedding space onto the unit sphere \cite{watson_book}, i.e. $|\vec{x}|=1$. An alternative to the Fisher-Bingham family is obtained via marginalization of the radial coordinate instead of conditioning on it, which results in the \textbf{A}ngular \textbf{G}aussian (AG) family of distributions \cite{generic_ag_paper}. 

As we discuss in section \ref{section:combining_flows},
these distributions are not in general writable as a normalizing flow. However, two specific subclasses of them are: The von-Mises-Fisher distribution sub-class on the 2-sphere \cite{vmf_distribution} is a known normalizing flow, and the central angular Gaussian \cite{ag_bingham} in any dimension is also a known normalizing flow (see. Fig. \ref{fig:nf_overview}). We show that the vMF normalizing flow  "zooms in" onto a region, which is why we also call it "Fisher zoom", while the central angular Gaussian part allows to induce covariance structure into the problem with a linear transformation in embedding space followed by a projection. We call this part "linear-project".
Remarkably, different orderings of the "zoom" or the "linear-project" step re-create qualitatively all the distributions in either of these families (see table \ref{tab:equivalent_fb_distributions}), including the most general one of $\mathrm{FB}_8$ type.
We therefore call these flows "\textbf{Z}oom-\textbf{L}inear-\textbf{P}roject" (ZLP) Fisher flows. Certain important properties are fulfilled, i.e. we show that the "Kent" version of the ZLP-Fisher flow also approaches a multivariate Gaussian in tangent space as the concentration parameter becomes large.
Further we discuss how to sample from the vMF distribution in any dimension without rejection sampling using a proper diffeomorphism. This defines the resulting normalizing-flow in any dimension, even though it is most efficient for the 2-sphere since only there closed-form inverses exist.

Several of these basic flows can be chained together, intertwined by rotations, to create more complex distributions, and we test these in a conditional density estimation setting on the 2-sphere in section \ref{sec:conditional_density_test}, where ZLP Fisher flows are seen to behave very stable compared to alternatives when the entire flow parameters are the output of a neural network. In particular, the Kent-like version is shown to efficiently upgrade established flows to add better first and second moment capabilities in the situation where the conditional target distributions vary by order of magnitudes in scale.

\section{The Fisher-Bingham and Angular Gaussian Families}

The \textbf{F}isher-\textbf{B}ingham (FB) family \cite{mardia_fb8_1975}\cite{fb5_paper} is a well-known fundamental family of distributions on the sphere. It is obtained by conditioning a generic Gaussian distribution in embedding space onto the unit sphere \cite{watson_book}, i.e. $|\vec{x}|=1$.
Special cases include the unimodal vMF distribution \cite{vmf_distribution} which comes from an isotropic covariance matrix, the Watson distribution \cite{watson_distribution_great_dircle} which defines a density along the great circle of the equator (in $D=3$) and comes from a Gaussian at the origin with a covariance matrix where two diagonal entries are equal, and the Bingham distribution \cite{bingham_distribution_paper}, which generalizes the Watson to include bimodal densities by generalizing the allowed covariance structure. Furthermore there are the "small circle" $\mathrm{FB}_4$ distribution \cite{fb4_paper}, which comes from a Gaussian offset from the origin with a covariance similar to the symmetric Bingham/Watson case, and the $\mathrm{FB}_5$ or "Kent" distribution \cite{fb5_paper}, which is a bivariate (multivariate in higher $D$) generalization of the vMF distribution. Here, the covariance parameters are restricted so the distribution is always unimodal. There is also a $\mathrm{FB}_6$ distribution \cite{rivest_fb6}, which unifies both the $\mathrm{FB}_4$ and $\mathrm{FB}_5$ in a single parametrization. It contains other special cases like bimodal small-circle distributions. Finally, for a generic mean and covariance matrix of the Gaussian one obtains the most general $\mathrm{FB}_8$ form \cite{fb8_paper}, which includes unimodal and bimodal asymmetric distributions and contains all the others as special cases. 

An alternative to the Fisher-Bingham family is obtained via marginalization of the radial coordinate instead of conditioning on the unit sphere, which results in the \textbf{A}ngular \textbf{G}aussian (AG) family of distributions \cite{generic_ag_paper}. They are also known as projected Gaussians since marginalization is equivalent to projection. These distributions are less often mentioned in the literature, but in principle the family matches qualitatively the Fisher-Bingham distributions - for example there is a Bingham/Watson version \cite{ag_bingham} and a "Kent" version \cite{ag_kent_paper} of the angular Gaussian and so on. We will stick with the FB-nomenclature for this paper for simplicity. 

As is indicated in Fig. \ref{fig:nf_overview} a) and b), only the  vMF and central AG subsets allow a normalizing flow description. As soon as one has the generic Gaussian in embedding space as in Fig. \ref{fig:nf_overview} c), neither class does so anymore, but specific combinations do (see Fig. \ref{fig:nf_overview} d), which we explore in more detail in the following.

\section{Normalizing flows on the sphere}

Normalizing flows allow to define a probability density $p(\vec{x})$ on a target space $\vec{x}$  via a flow-defining diffeomorphism $\Phi_\theta(\vec{x}_b)$ and base distribution $p_0(\vec{x}_b)$ defined on an "auxiliary" base pace $\vec{x}_b$. The generic change of variable formula
\begin{equation}
    p_{\theta}(\vec{x})=p_0(\Phi_\theta^{-1}(\vec{x})) \cdot D_{upd.}(\Phi_\theta^{-1}(\vec{x}))\label{eq:sphere_nf_def},
\end{equation} allows to evaluate that target space density exactly. For Euclidean space, the density update is just the Jacobian determinant. For manifolds like the sphere, the density update $D_{upd.}(\Phi_\theta(\vec{x}))$ for an arbitrary diffeomorphism $\Phi_\theta(\vec{x})$ \footnote{For density functions, the diffeomorphisms $\Phi_\theta(x)$ are really the inverses of the flow function and the corresponding density update involves the Jacobian of the inverse.} can be defined as \cite{tori_and_spheres_paper}
\begin{align}
    D_{upd.}(\Phi_\theta(x))&=\sqrt{\mathrm{Det} \left[E^T(\vec{x}) \cdot J_{\Phi_\theta}^{T}(\vec{x}) \cdot J_{\Phi_\theta}(\vec{x}) \cdot E(\vec{x})\right]} \nonumber  \\
    & =\sqrt{\mathrm{Det} \left[\tilde{J}_{\theta}^{T}(\vec{x}) \cdot \tilde{J}_{\theta}(\vec{x})\right]}
\end{align} where the Jacobian is calculated treating $\vec{x}$ as the embedding coordinates and $E(\vec{x})$ is an orthogonal projection matrix that projects the Jacobian into the tangent space of manifold at $\vec{x}$. We define $\tilde{J}=J \cdot E$ to simplify notation.  For the 2-sphere in particular, the embedding coordinates would be $(x, y, z) \in S^2$. On top of probability evaluation we can also sample from $p_{\theta}(x)$ by first sampling from $p_0(\vec{x}_b)$, for example the flat distribution on the sphere, and passing those samples through the function $\Phi_\theta(\vec{x}_b)$.

\begin{table*}[ht]
  \centering
  \begin{tabular}{|C{3cm}|C{5.6cm}|C{6.5cm}|}
    \hline
    \textbf{Type} & \textbf{Flow function}  & \textbf{PDF examples} \\
    \hline
    von-Mises-Fisher & $\Phi_\mathrm{F}(\vec{x})=[\Phi_\mathrm{R} \circ \phi_{\mathrm{Z}}](x)$  & 
      \includegraphics[height=1.5cm]{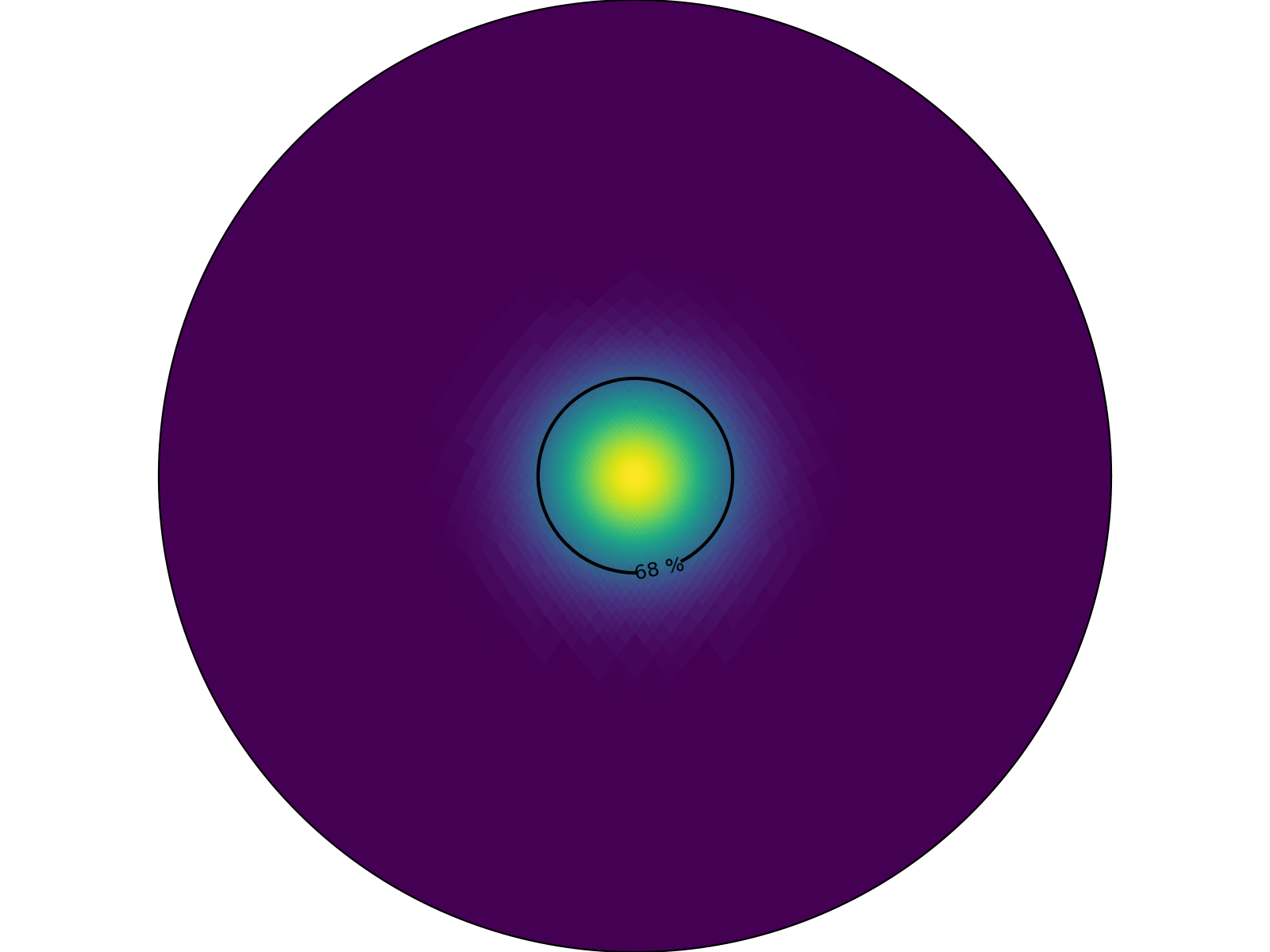}\hspace{3pt}%
      \includegraphics[height=1.5cm]{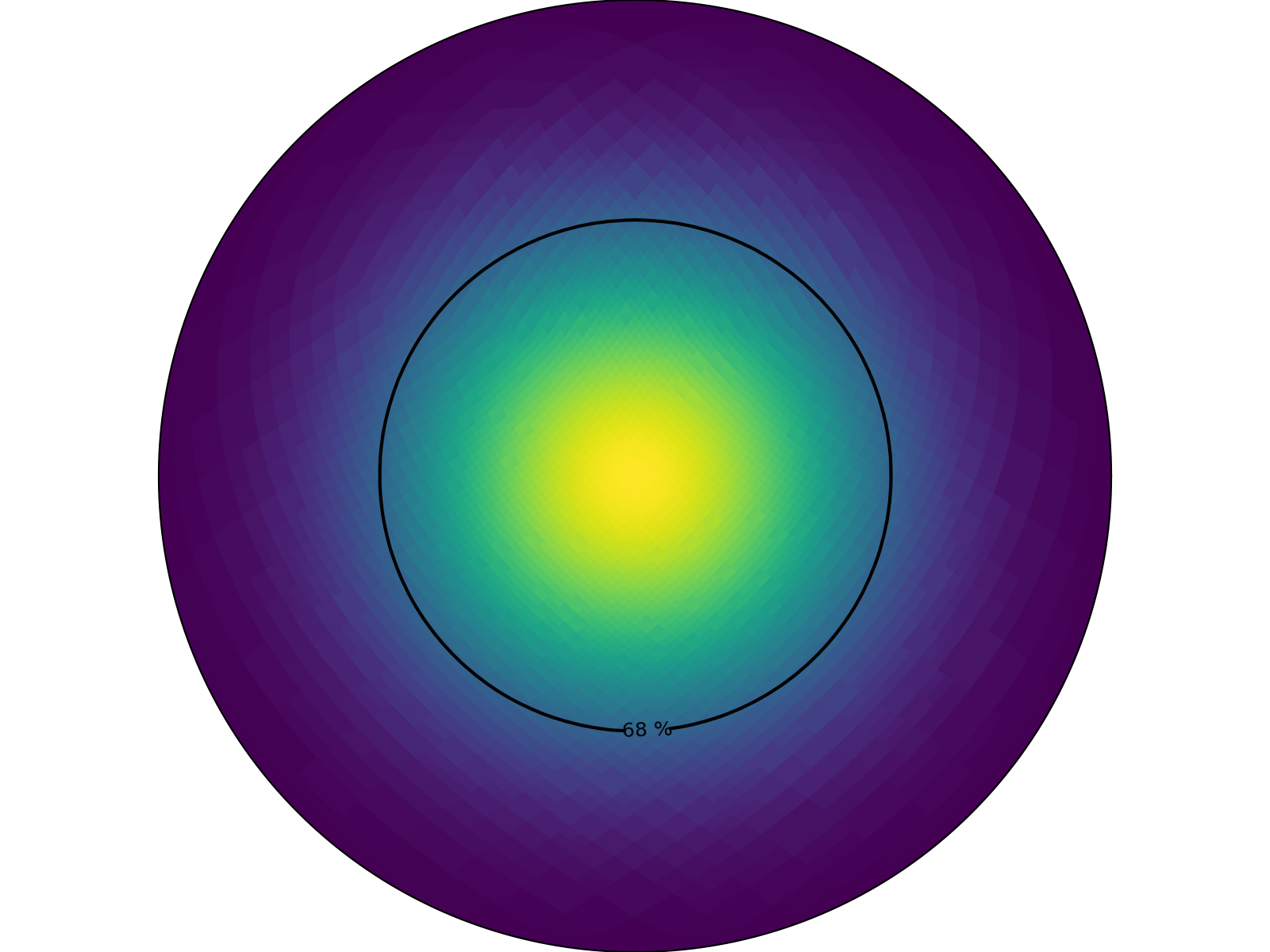} \\ \hline
    Bingham/Watson-like (central angular Gaussian) & $\Phi_\mathrm{B}(\vec{x})=[ \Phi_{\mathrm{LP}}](x)$ or  $\Phi_\mathrm{B}(\vec{x})=[\Phi_{\mathrm{R}} \circ \Phi_{\mathrm{LP,S}}](x)$ &
    \includegraphics[height=1.5cm]{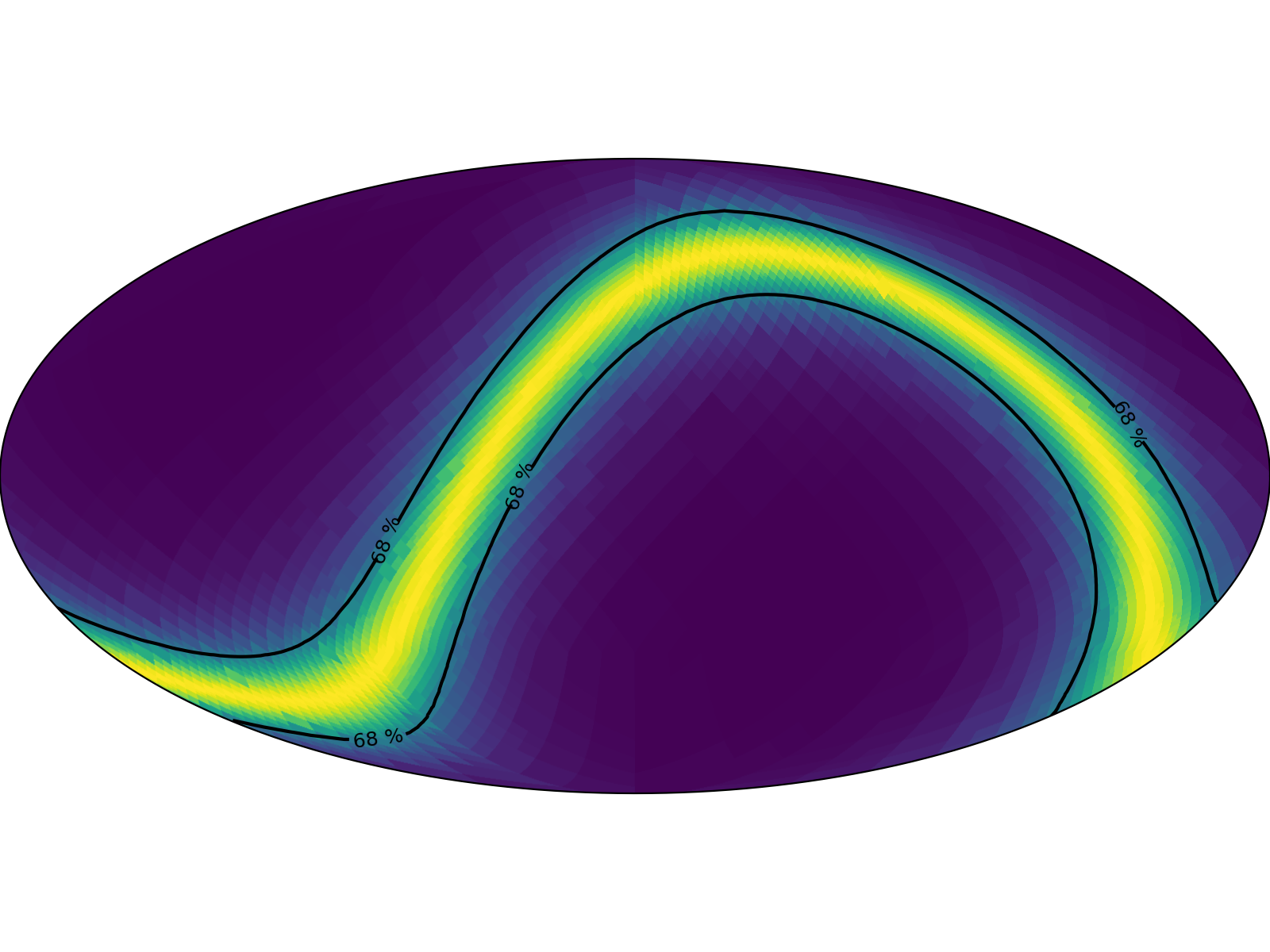}
      \includegraphics[height=1.5cm]{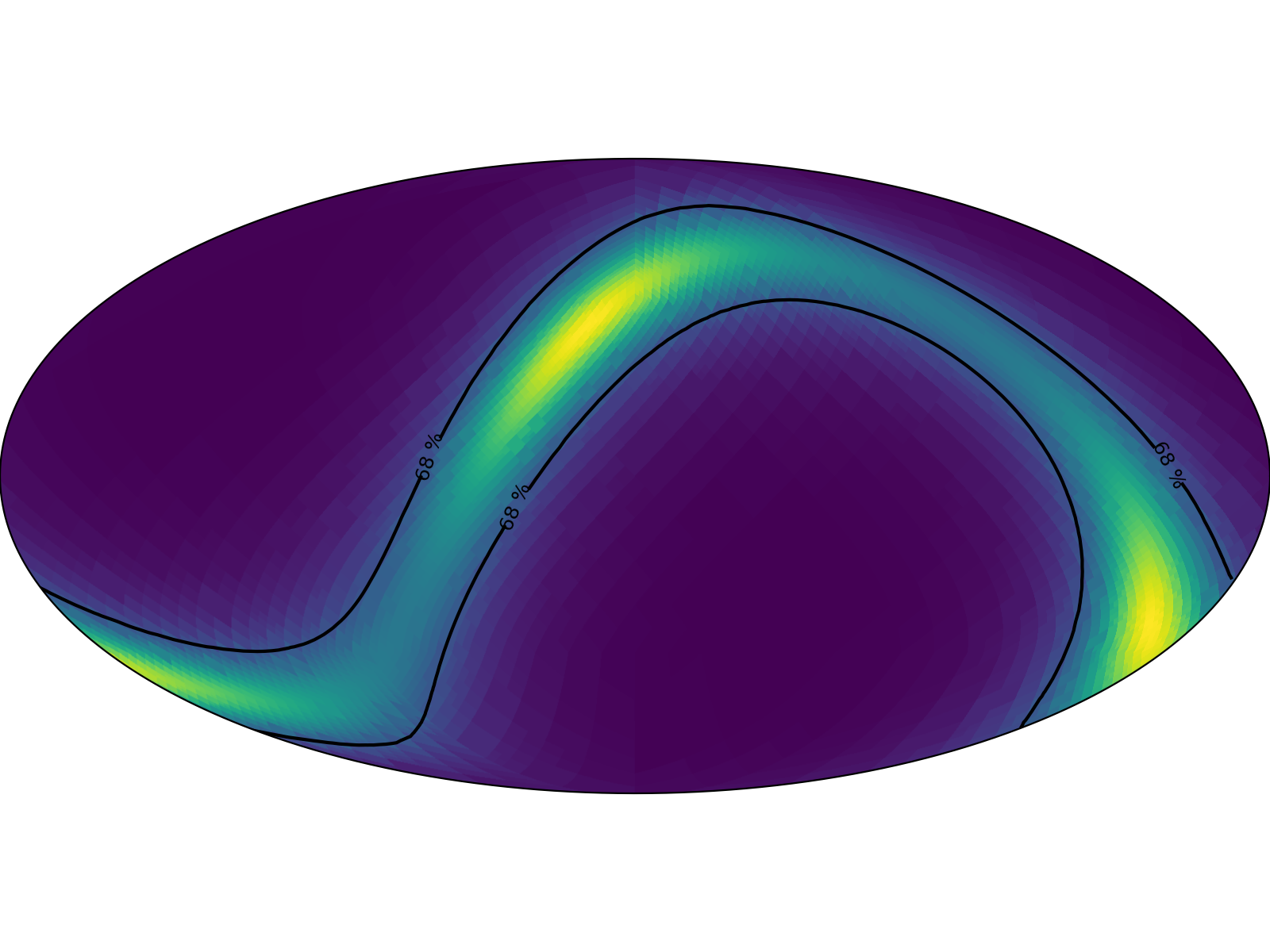}\hspace{3pt}%
      \includegraphics[height=1.5cm]{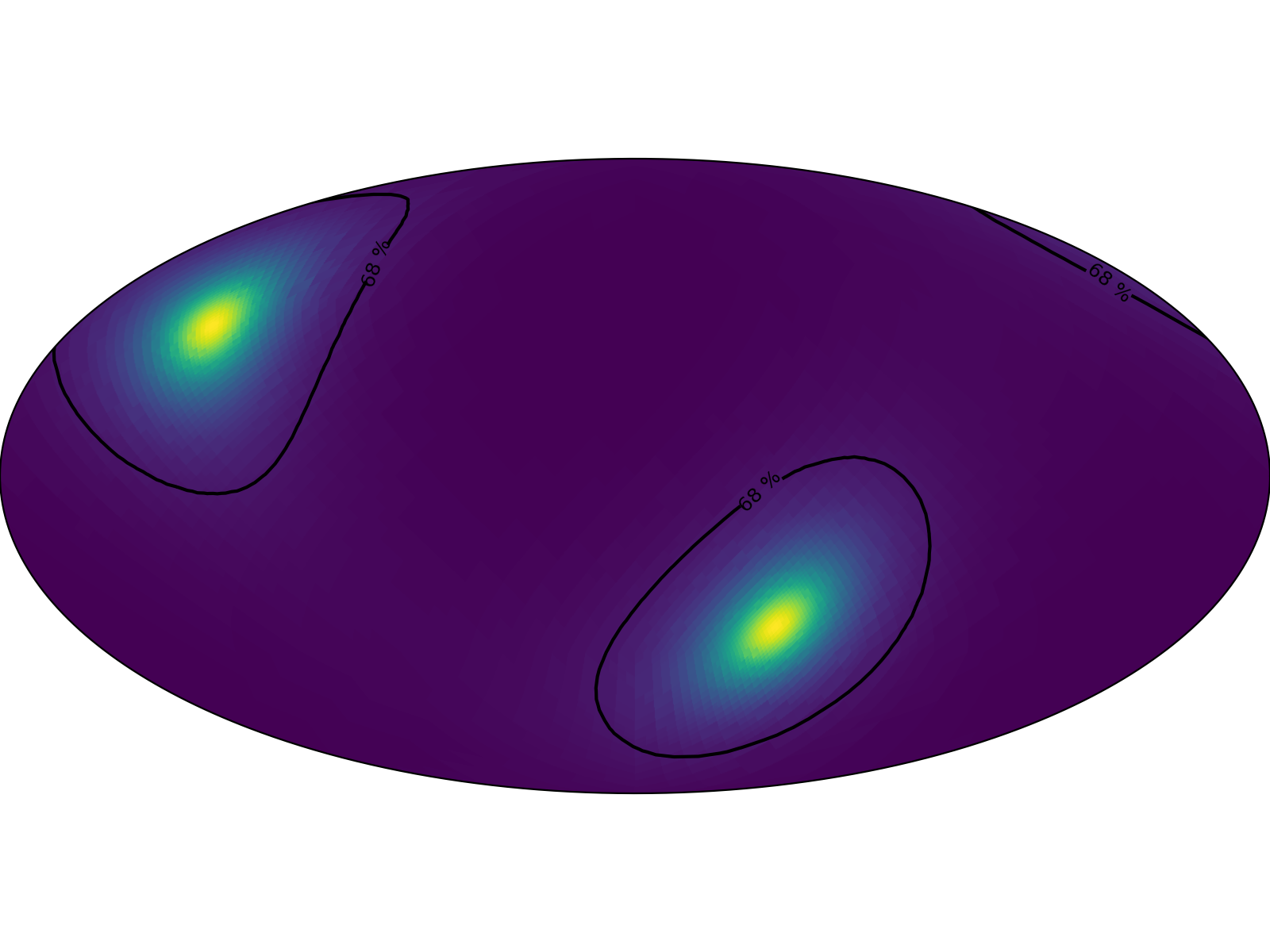} \\ \hline
    $\mathrm{FB}_4$-like (small circle + bimodal $\mathrm{FB}_6$ subset) & $\Phi_{\mathrm{FB}_4}(\vec{x})=[\Phi_{\mathrm{R}} \circ \Phi_\mathrm{Z} \circ \Phi_{\mathrm{LP,S}}](x)$  &
    \includegraphics[height=1.5cm]{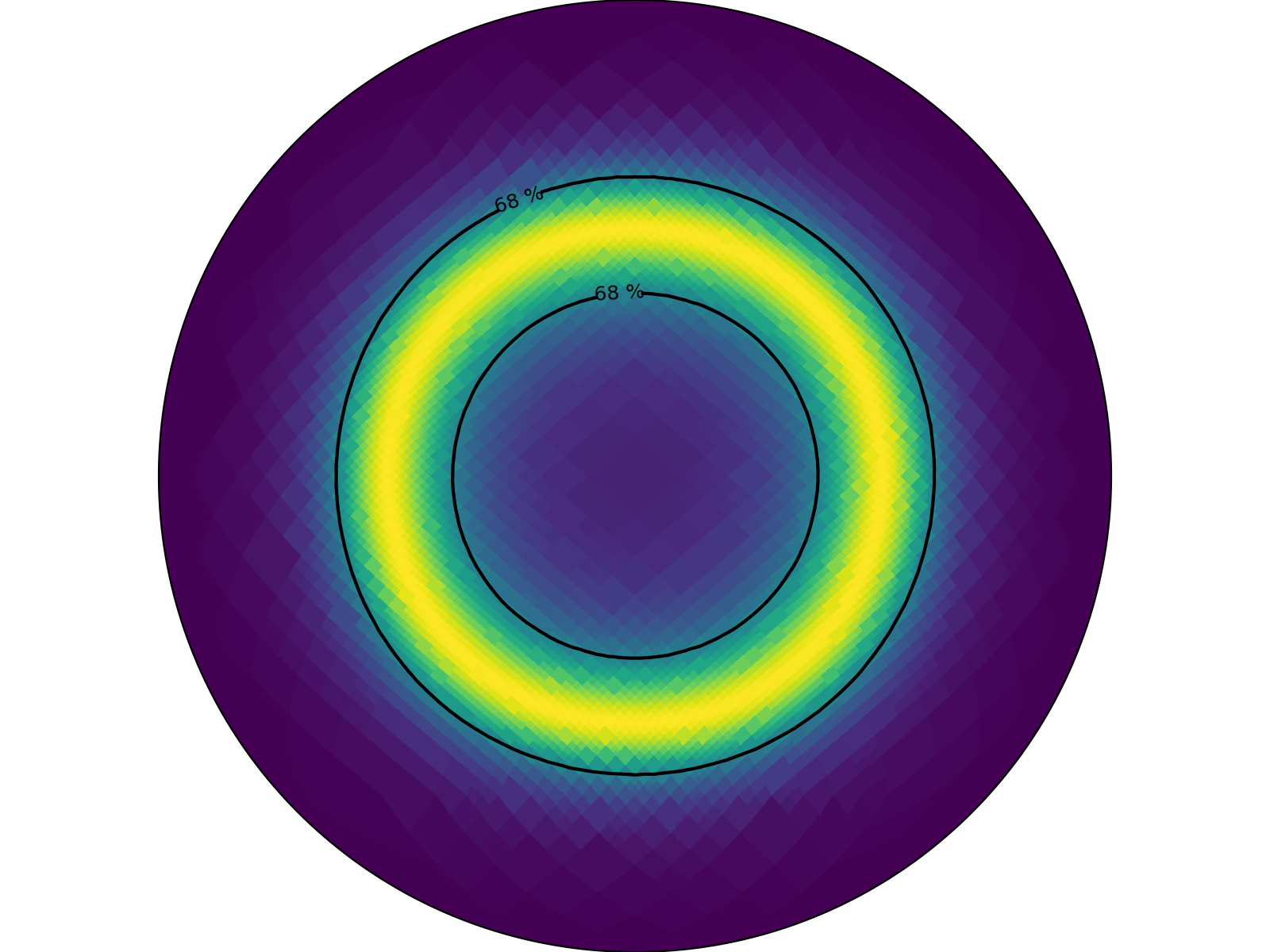}
      \includegraphics[height=1.5cm]{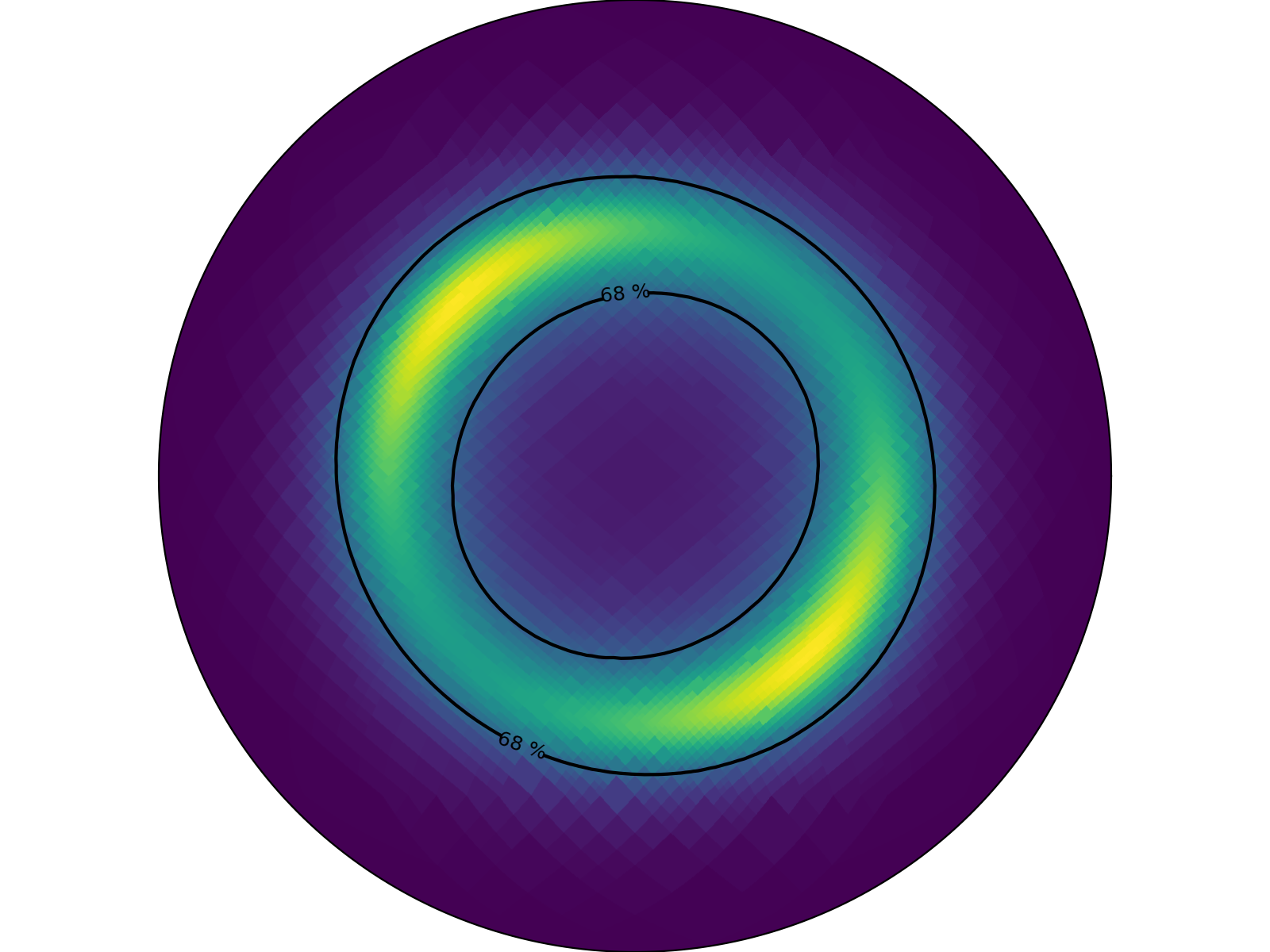}\hspace{3pt}%
      \includegraphics[height=1.5cm]{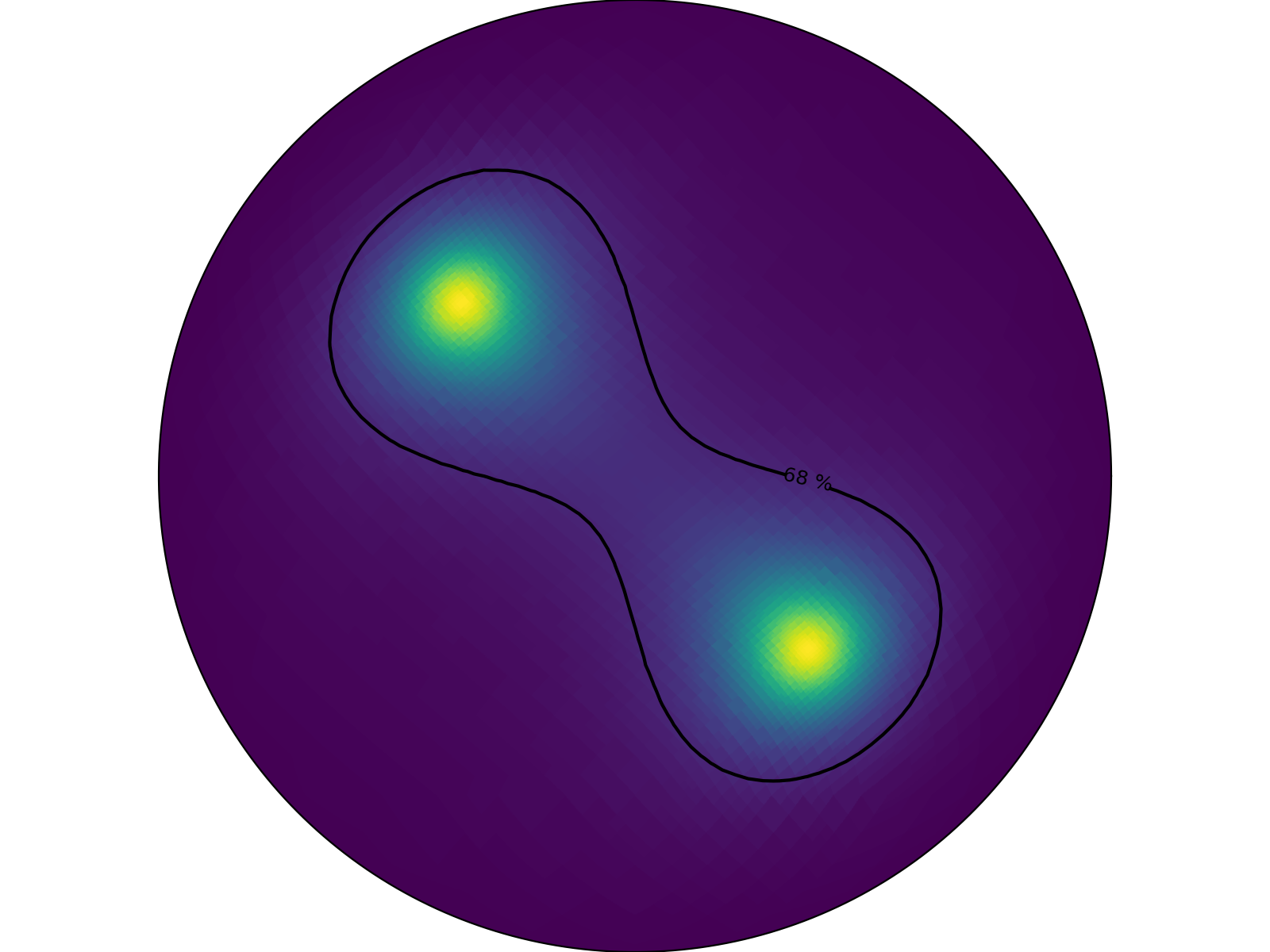} \\ \hline
    $\mathrm{FB}_5$/Kent-like & $\Phi_K(\vec{x})=[\Phi_{\mathrm{R}} \circ \Phi_{\mathrm{LP,S_c}} \circ \Phi_\mathrm{Z} ](x)$  &
      \includegraphics[height=1.5cm]{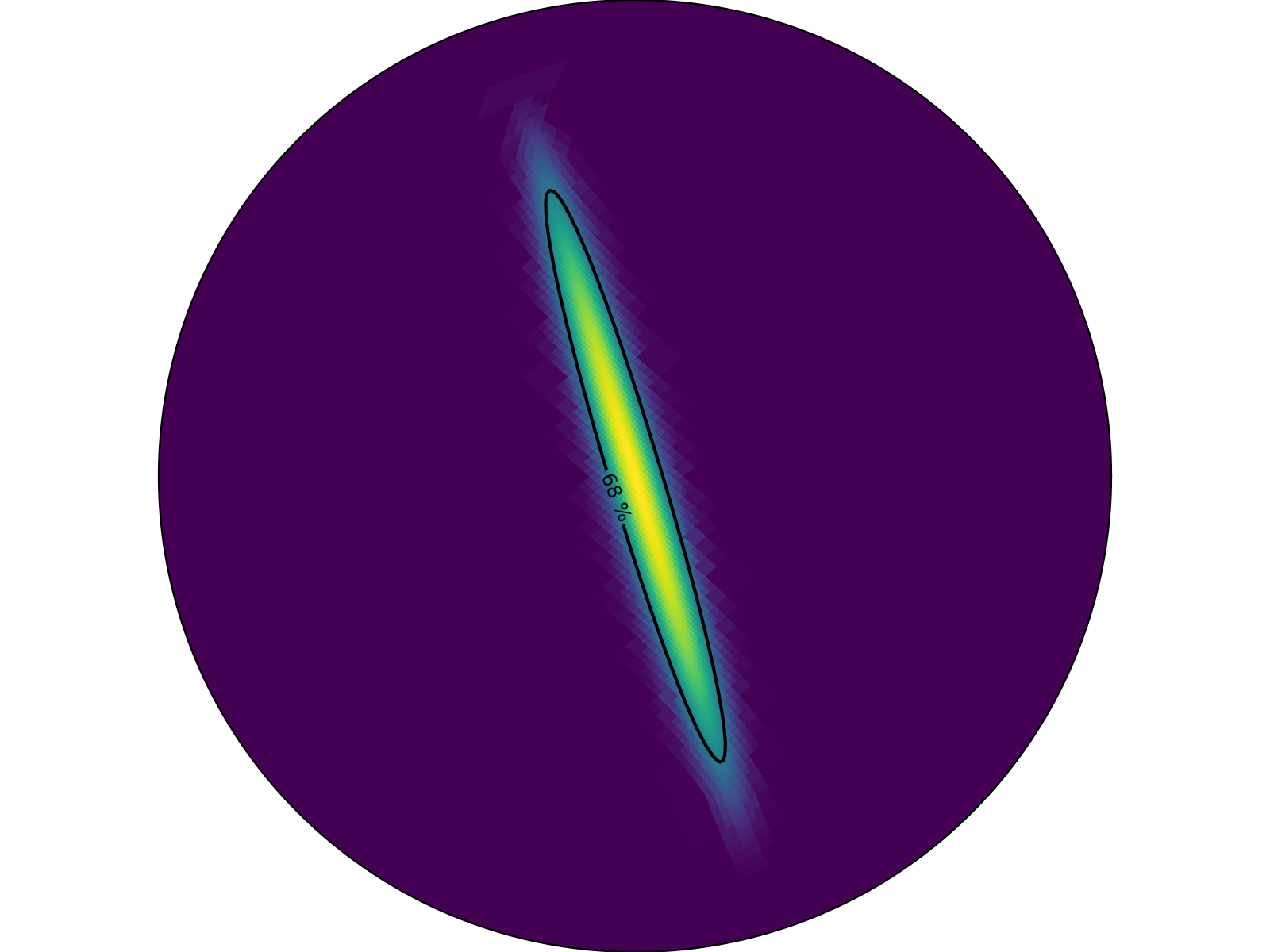}\hspace{3pt}%
      \includegraphics[height=1.5cm]{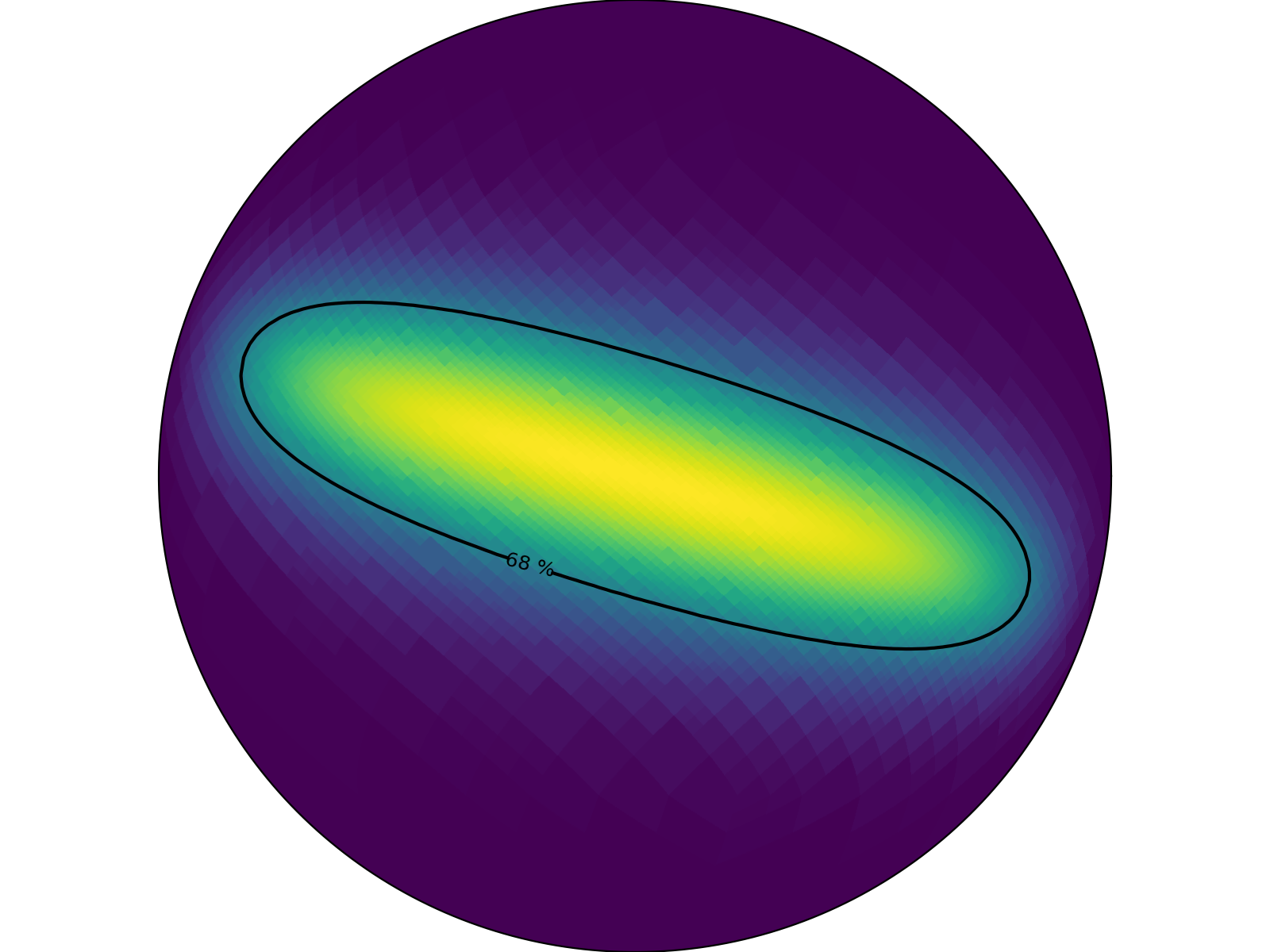}\\ \hline
    $\mathrm{FB}_6$-like & $\Phi_{\mathrm{FB}_6}(\vec{x})=[\Phi_\mathrm{R} \circ \Phi_{\mathrm{LP,S_c}} \circ \Phi_\mathrm{Z} \circ \Phi_{\mathrm{LP_,S}}](x)$  &
      \includegraphics[height=1.5cm]{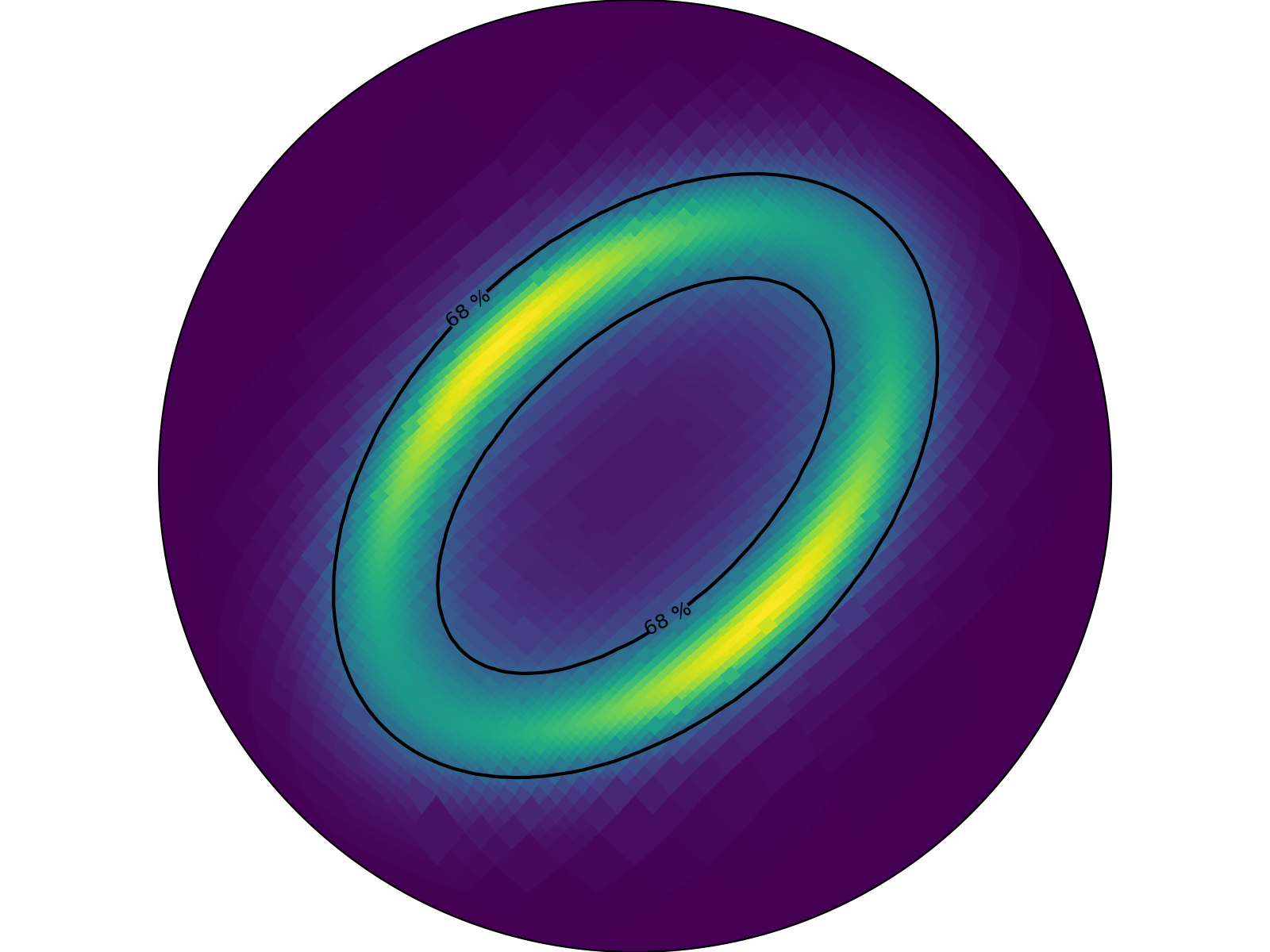}\hspace{3pt}%
      \includegraphics[height=1.5cm]{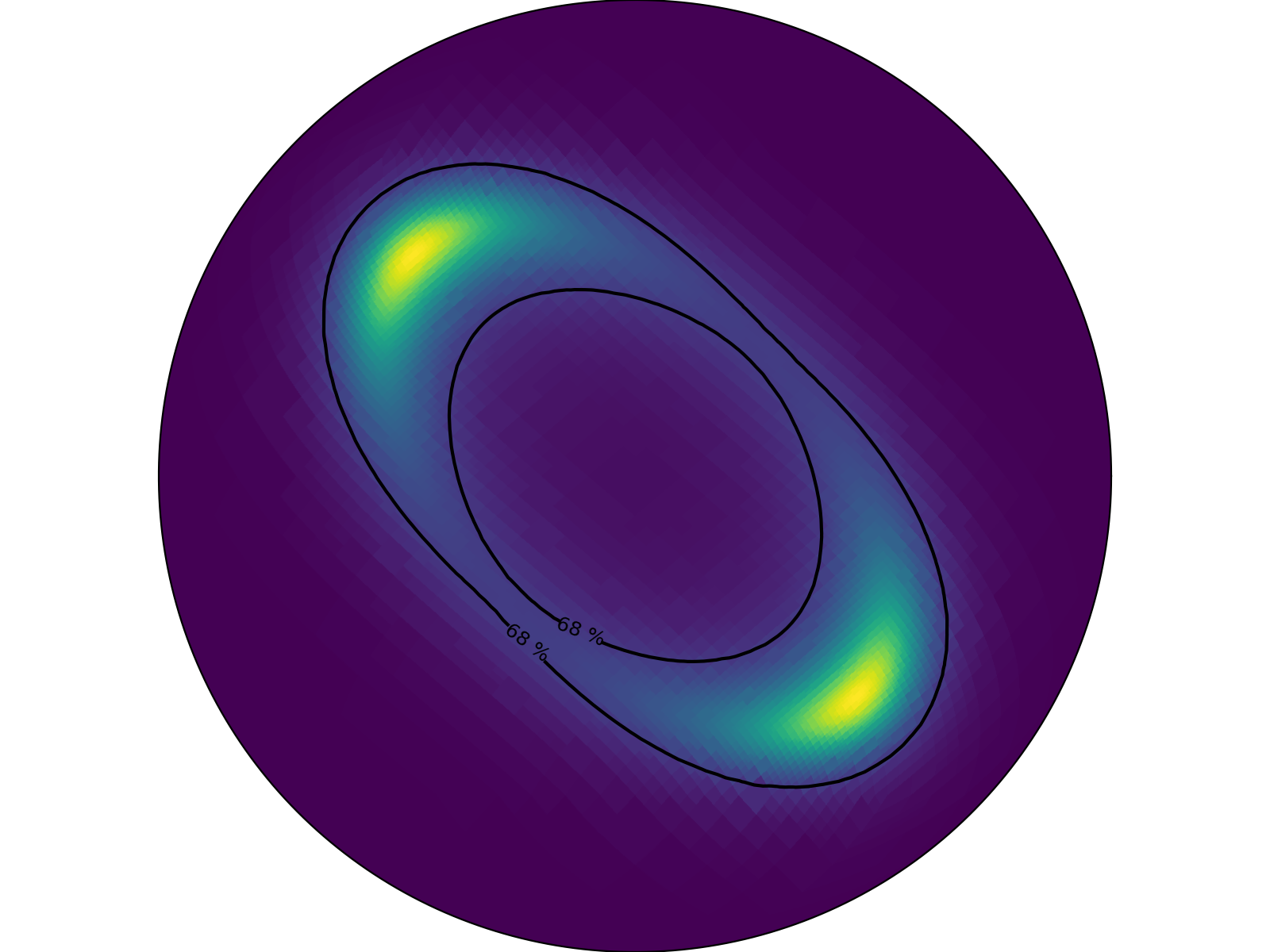} \\ \hline
    $\mathrm{FB}_8$-like & $\Phi_{\mathrm{FB}_8}(\vec{x})=[\Phi_\mathrm{R} \circ \Phi_{\mathrm{LP,S_c}} \circ \Phi_\mathrm{Z} \circ \Phi_{\mathrm{LP}}](x)$  &
      \includegraphics[height=1.5cm]{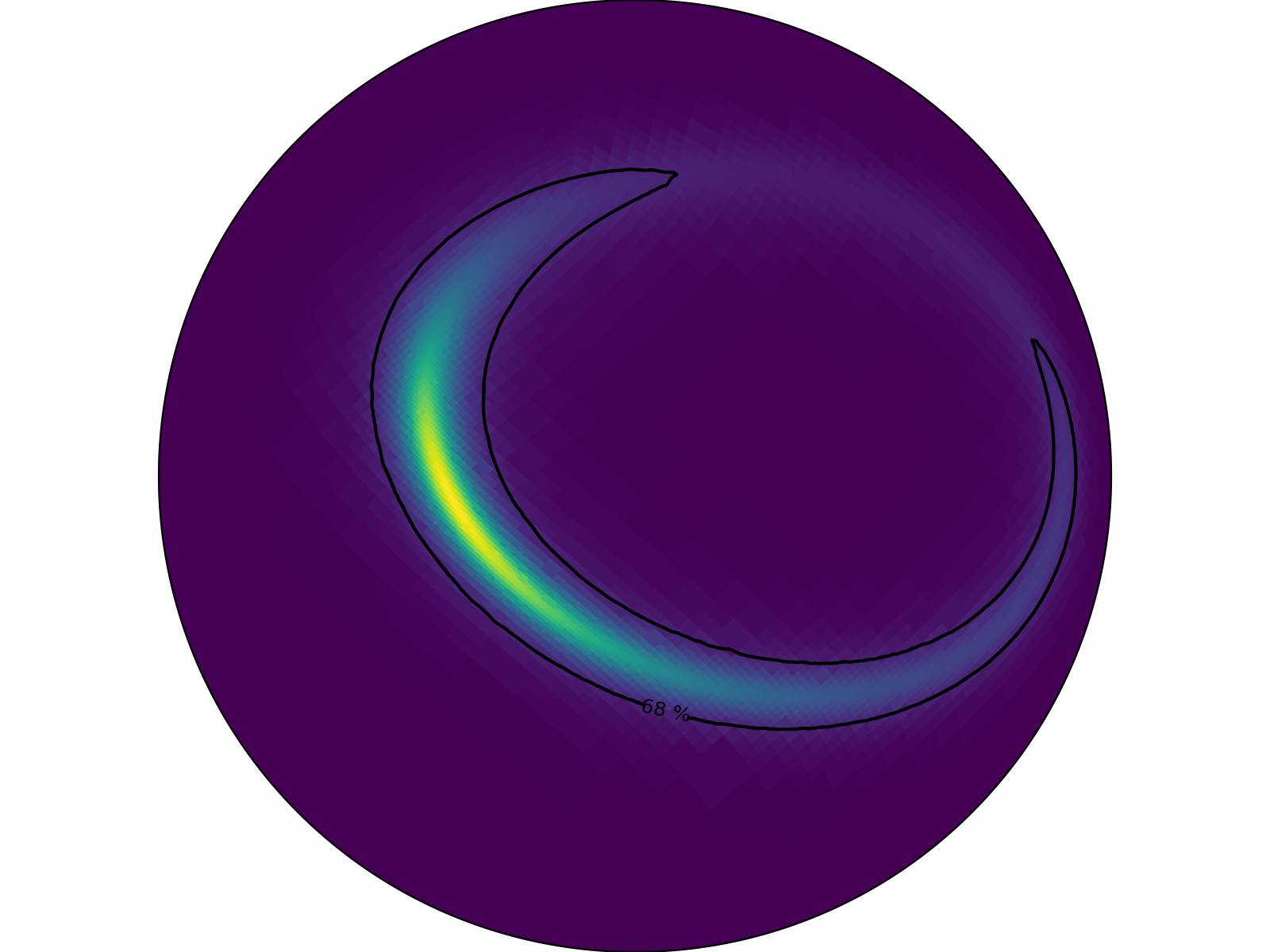}\hspace{3pt}
      \includegraphics[height=1.5cm]{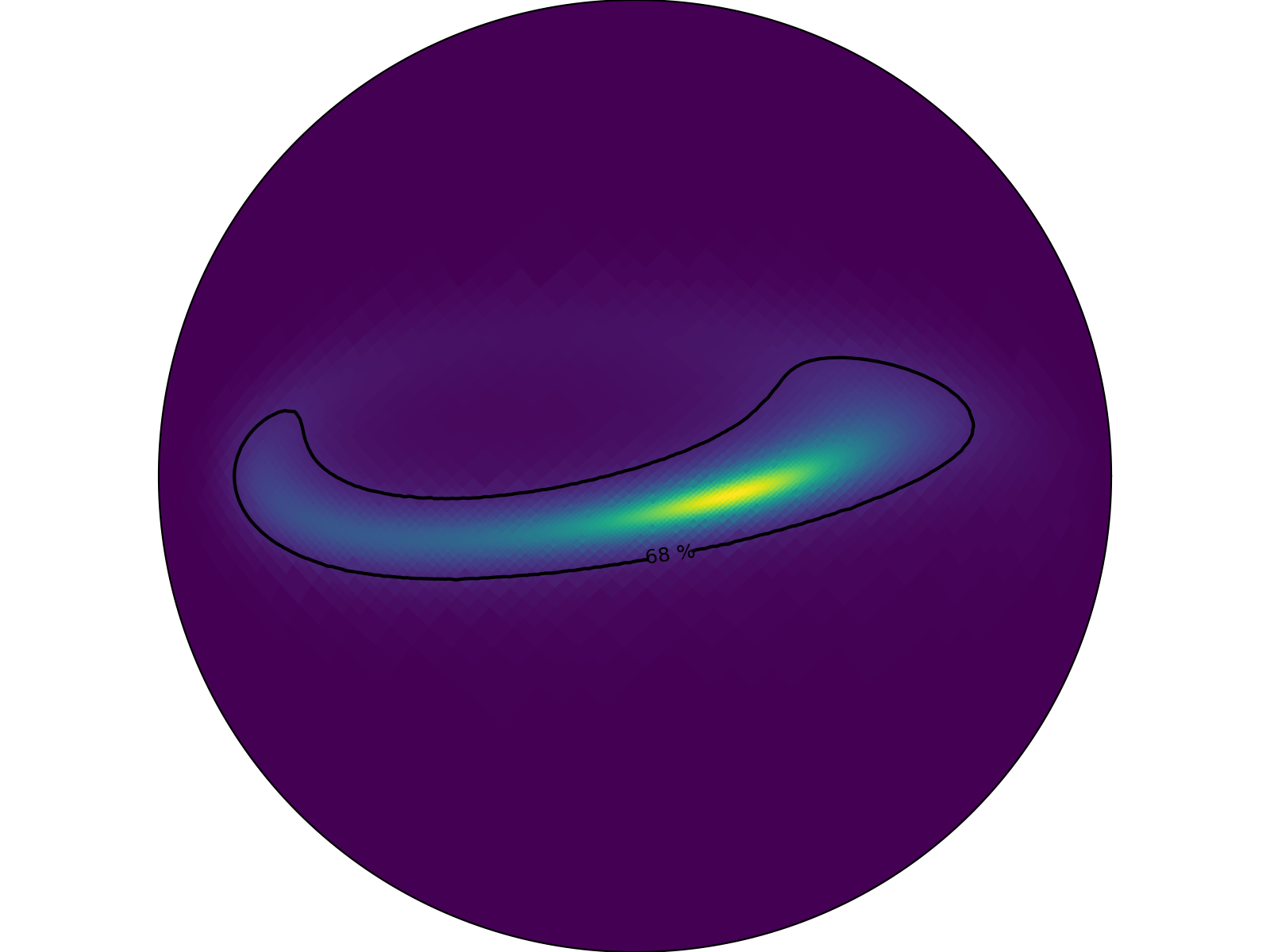}\\
      \hline
    generic & $\Phi_G(\vec{x})=[\Phi_\mathrm{R}  \circ \Phi_\mathrm{Z} \circ \Phi_{\mathrm{LP}}]^N(\vec{x})$  &
      \includegraphics[height=1.5cm]{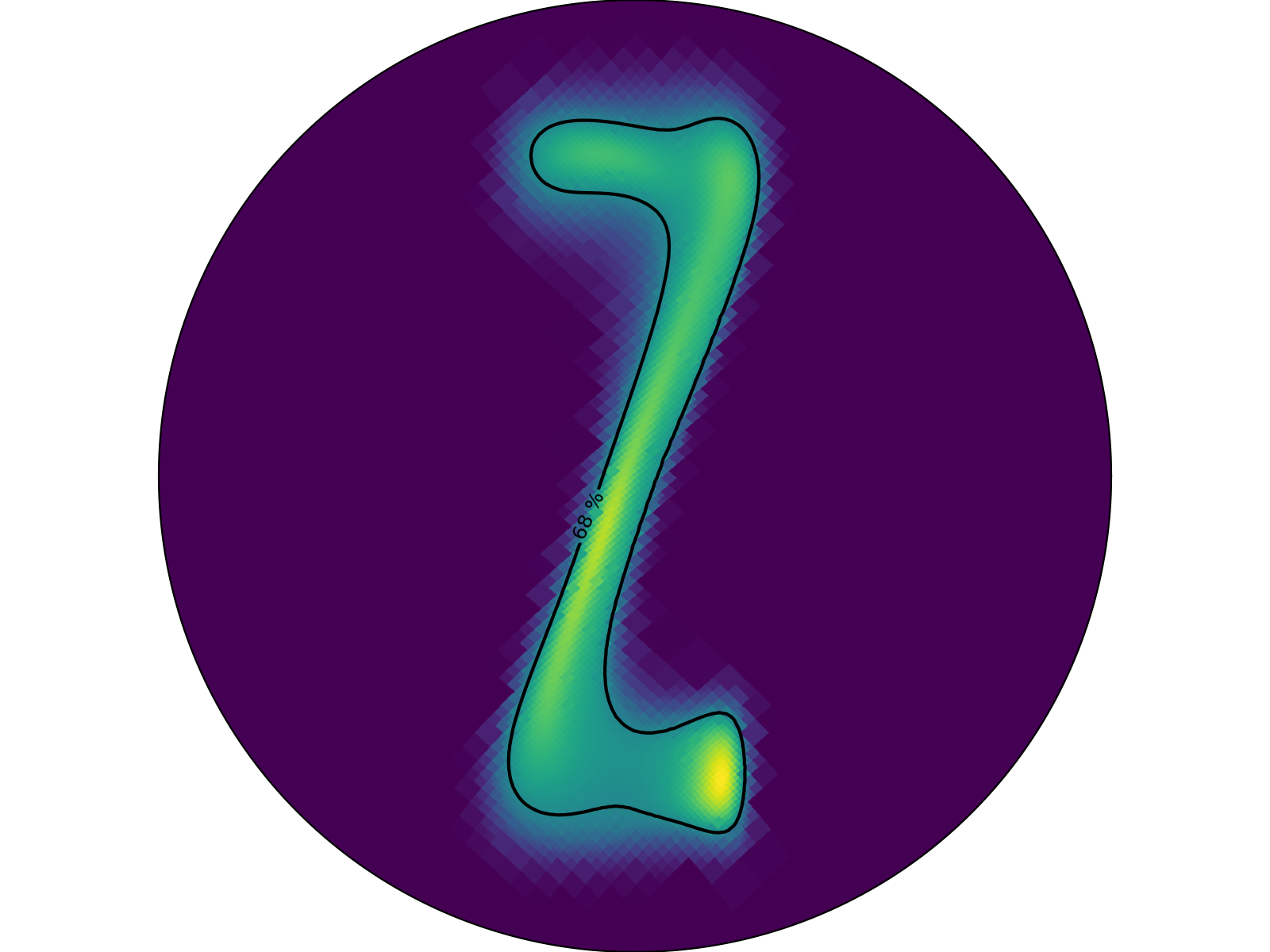}\hspace{3pt}%
      \includegraphics[height=1.5cm]{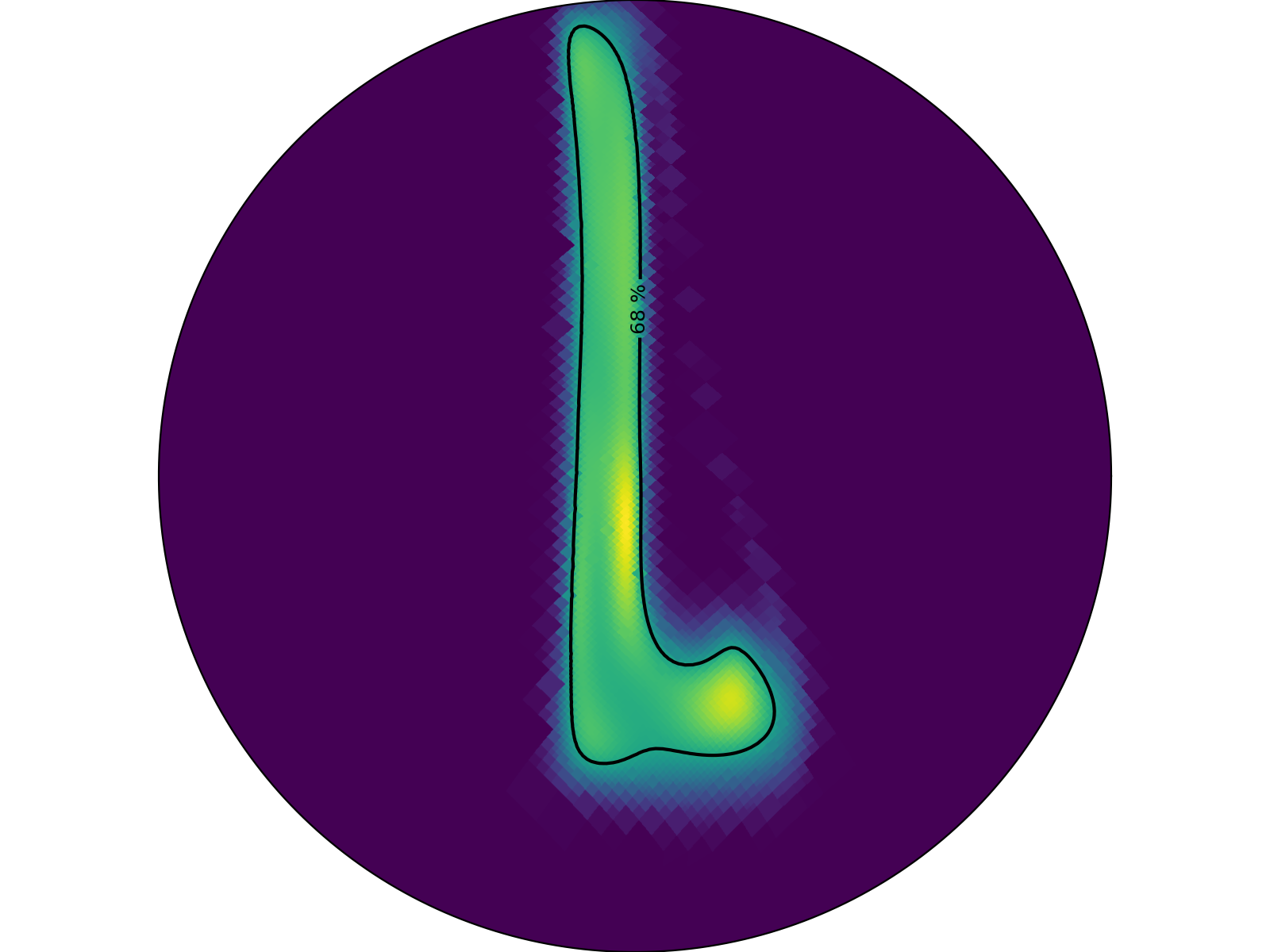}\hspace{3pt}
      \includegraphics[height=1.5cm]{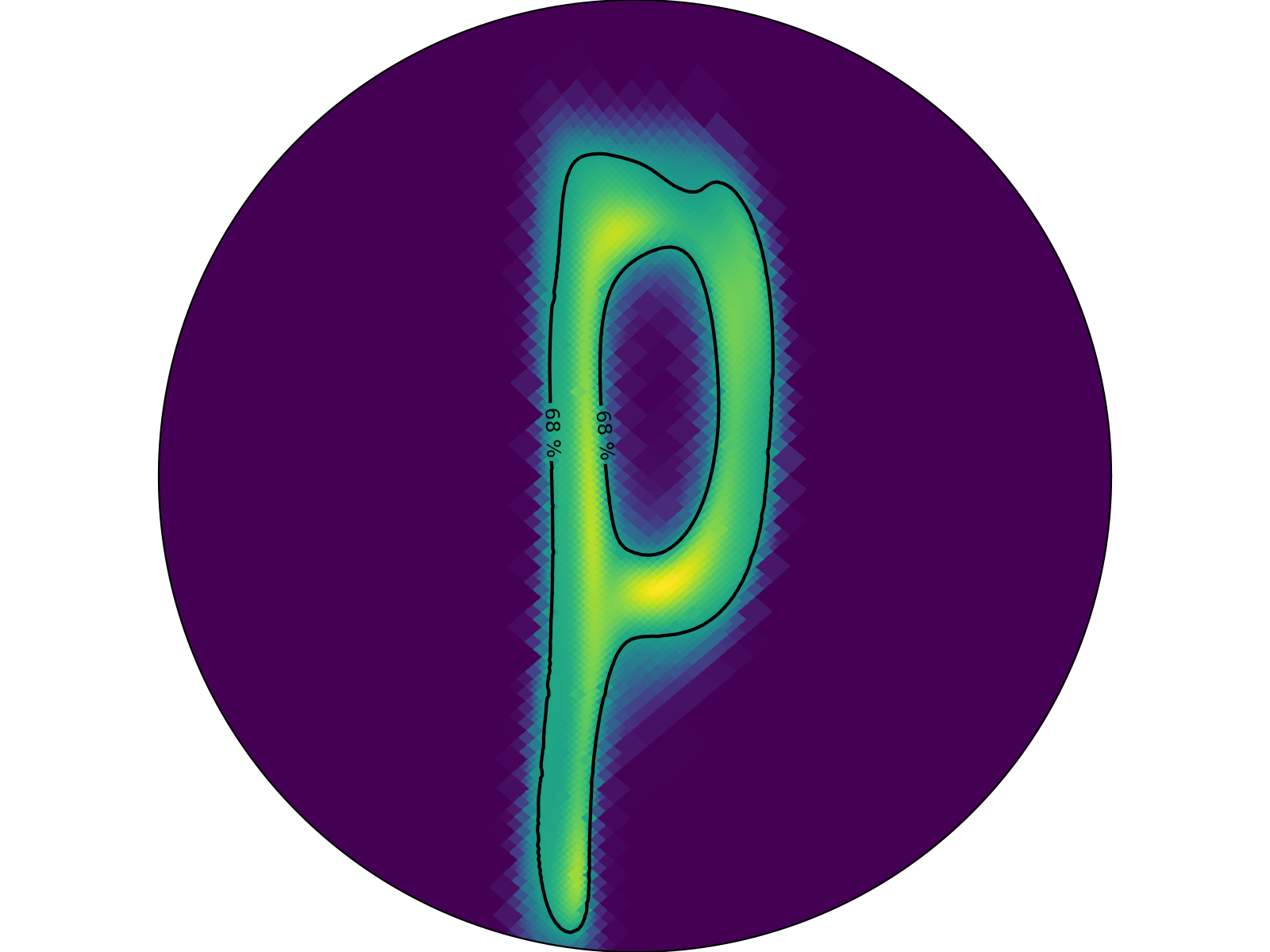} \\ \hline
  \end{tabular}
  \caption{The dictionary how different combinations of the "Fisher-zoom" (Z), the "linear-project" (LP) flow, and rotations (R) generate the various distributions of the "ZLP"-Fisher family. The last row indicates how iterative combinations of these flows also gives rise to more complex distributions, using $N=15$. The distributions are visualized using orthographic projection, where only a (zoomed-in) patch of the sphere is visible. An exception is the Bingham-like case which uses Mollweide projection and shows the whole sphere.}
  \label{tab:equivalent_fb_distributions}
\end{table*}

\subsection{"Fisher zoom" - the von-Mises-Fisher distribution}
\label{sec:fisher_zoom}
In order to define the von-Mises-Fisher distribution as a normalizing flow, it is useful to define the following proposition first.
\begin{proposition}[Simple embedding-based density update]
\label{proposition:simple_sphere_flow}
Let \(\vec{x} = (x_1,\ldots,x_D)\in S^{D-1}\) be a point on the D-1 sphere. Let \(h\colon [-1,1]\to\mathbb{R}\) be a diffeomorphism on the interval $[-1,1]$. Then \(\Phi (\vec{x})=(x_1\cdot \sqrt{\frac{1-h(x_i)^2}{1-x_i^2}},\ldots,h(x_i),\ldots,x_D \cdot \sqrt{\frac{1-h(x_i)^2}{1-x_i^2}}) \) is the corresponding diffeomorphism on the D-1 sphere and \( \sqrt{\mathrm{Det} \left[\tilde{J}_{\Phi}^{T}(x) \cdot \tilde{J}_{\Phi}(x)\right]}=h^{'}(x_i) \sqrt{ \left[\frac{1-h(x_i)^2}{1-x_i^2}\right]^{D-3} }  \) is its density update.
\end{proposition}
Proof in appendix. We can see that any 1-d transformation $h(x)$ that is a diffeomorphism on $[-1,1]$ will generate an associated diffeomorphism on the sphere that is symmetric around the axis $x_i$ on which it is applied. No transformation to spherical coordinates is required. The following theorem states the corresponding $h_{F,D}(x)$ that leads to the (D-1) dimensional vMF distribution.
\begin{theorem}[Diffeomorphism to generate vMF flow in any dimension]
\label{theorem:any_d_vmf}
    Let
    \begin{equation}
        U(x_D)\equiv  I_{\frac{x_D+1}{2}}\left(\frac{D-1}{2}, \frac{D-1}{2}\right)  \in [0,1]
    \end{equation} and
    \begin{align}
        F_{\kappa}(x_D)\equiv\frac{\int\limits_{-1}^{x_D}\mathrm{e}^{\kappa \cdot t} \cdot (1-t^2)^{(D-3)/2} dt}{I_{D/2-1}(\kappa) \cdot \sqrt{\pi} \cdot \Gamma(\frac{D-1}{2}) \cdot (2/\kappa)^{D/2-1}} \in [0,1]
    \end{align}
    where $\mathrm{I_\frac{x_D+1}{2}(D+1,D+1)}$ is the symmetric regularized incomplete Beta function and $\Gamma$ the Gamma function. 
    The transformation $h_{F,D}=F_\kappa^{-1}\left(U(x_D)\right)$ leads via proposition \ref{proposition:simple_sphere_flow} to a global diffeomorphism $\Phi_Z(\vec{x})$ on the (D-1) sphere. Using the change-of-variable formula in eq. \ref{eq:sphere_nf_def}, its inverse $\Phi^{-1}_Z(\vec{z})$ with a uniform base distribution defines the vMF density with mean $(0,\ldots,1)$ on the (D-1) sphere. 
\end{theorem}
A proof is found in the appendix, which involves solving an ODE based on proposition \ref{proposition:simple_sphere_flow}. One can identify $U$ and $F$ with the CDF of the marginals of the uniform and vMF distribution along its symmetry axis, respectively. The functions admit finite-sum or series representations in various regimes, which can be useful in practice. 

\begin{corollary}[Finite sums for U and F]
\label{vmf_trafo_corollary}
For even-dimensional spheres (odd D), we can simplify $U(x_D)$ and $F(x_D)$ from theorem \ref{theorem:any_d_vmf} to the following finite sum representations:
\begin{equation}
    U(y[x_D])= \sum_{i=A}^{n} \Binom{n}{i}\, y^{i}(1-y)^{n-i}
\end{equation}
and
\begin{equation}
    F_\kappa(y[x_D]) \;=\; \sum_{i=A}^{n} \Binom{n}{i}\, y^{i}(1-y)^{n-i} \cdot w_{A,i,y,\kappa}
\end{equation}
with $\displaystyle A=\frac{D-1}{2}$, $\displaystyle y=\frac{x_D+1}{2}$, $\displaystyle n=D-2$ and $\displaystyle w_{A,i,y,\kappa}=\frac{\M(A;\,i+1;\,2 \kappa y)}{\M(A;\,2A;\,2 \kappa)}$ where $\M(a,b,x)$ is the Kummer confluent hypergeometric function.
\end{corollary}
Proof in appendix. We found this representation to be stable under inversion within $h_{F,D}(x)$ in tests up to at least $D=100$ and $\kappa > \num{1e6}$. For odd-dimensional spheres (even D), the expressions are typically more challenging. For $D=3$, we have $A=i=1$, yielding $U(z)=\frac{z+1}{2}$ and with $\displaystyle \M(1,2,x)=\frac{e^x-1}{x}$ we get $\displaystyle F_\kappa(z)=\frac{e^{2\kappa \cdot (z+1)/2}-1}{e^{2\kappa}-1}$. Via theorem \ref{theorem:any_d_vmf} these combine to
\begin{align}
    h_{F,3}(z)=1 + \frac{1}{\kappa} \mathrm{ln}\left[ \frac{1 + z}{2} + \left(1 - \frac{1 + z}{2}\right) \cdot e^{-2\kappa} \right] \label{eq:fisher_zoom_single}
\end{align}, a scaled version of the well-known transformation to generate samples for the vMF distribution on the 2-sphere \cite{fisher_sampling}. 
\begin{corollary}[Scaling behavior]
\label{vmf_fisher_zoom}
Let 
\begin{align}
\Phi_{Z}(\vec{x};h_{F,D})= \begin{pmatrix} x_1 \cdot \sqrt{\frac{1-h_{F,D}(x_D)^2}{1-x_D^2}} \\ x_2 \cdot \sqrt{\frac{1-h_{F,D}(x_D)^2}{1-x_D^2}} \\ \vdots \\ h_{F,D}(x_D)  \end{pmatrix}   \label{eq:fisher_zoom}
\end{align} be the  diffeomorphism on the sphere that generates the (D-1)-dimensional vMF density. For $x_D\rightarrow 1$ and $\kappa\gg1$, it behaves to leading order like a linear scaling in which all coordinates except the last are scaled as
\begin{equation}
    x_i \rightarrow x_i \cdot \frac{C}{\sqrt{\kappa}}
\end{equation}
with $\displaystyle C=\left( \frac{ (2 \pi)^{\frac{D-1}{2}} }{ S_{D-1} } \right)^{\frac{1}{D-1}}$ where $S_{D-1}$ is the surface volume of the $D-1$ sphere.
\end{corollary} 
A proof is given in the appendix. In practice, $\Phi_{Z}(\vec{x})$ generates a zoomed version of the input at the pole where $x_D \rightarrow 1$, and we call it "Fisher-zoom".

\subsection{"linear-project" - the bimodal central angular Gaussian distribution}
\label{sec:linear_project}

The only alternative normalizing flow to the Fisher zoom from the FB or AG family is related to the central angular Gaussian (see fig. \ref{fig:nf_overview} b) and consists of a linear transformation with subsequent projection onto the sphere. In contrast to the Fisher-zoom, which comes from an isotropic Gaussian that can be offset from the origin, the central angular Gaussian can have any covariance structure but must be at the origin.
In general, a central angular Gaussian distribution has density in standard form\cite{ag_bingham}
\begin{align}
p(\vec{x};\Lambda)=\frac{1}{ S^{D-1} \sqrt{\mathrm{Det}(\Lambda)}} \cdot (\vec{x} \Lambda^{-1} \vec{x})^{-D/2}, \label{eq:angular_gaussian_generic}
\end{align} 
where $S^{D-1}$ is the surface volume of the D-1 sphere and $\Lambda$ corresponds to the covariance matrix of the Gaussian in embedding space that is projected on the sphere.
It can be generated via the following inverse diffeomorphism that transforms this distribution back into a uniform distribution on the hypersphere \cite{ag_bingham}
\begin{align}
    \label{eq:ap_inverse}
    \phi_{\mathrm{LP}}^{-1}(\vec{x})=\frac{\Lambda^{-1/2} \cdot \vec{x}}{\sqrt{\vec{x}^T\Lambda^{-1} \vec{x}}} = \frac{A^{-1} \cdot \vec{x}}{ |A^{-1} \cdot \vec{x}|} 
\end{align}
where, and $A$ is a suitable decomposition of the covariance. In contrast to a standard linear transformation, the result is projected back to the sphere in a second step, hence we call it "linear-project". In the generic case we use a parametrization for $A$ based on a scaling matrix and Cholesky decomposition. We also use a reduced transformation where $A$ is a diagonal "scaling" matrix $S$, i.e. $A=S$, which we denote with $\Phi_{\mathrm{LP,S}}$. It is used for certain sub-families of the full ZLP-Fisher. Details of the decomposition and parametrizations are given in the supplementary material.
It can be checked that the forward diffeomorphism that defines the flow, i.e. the inverse of eq. \ref{eq:ap_inverse}, is
\begin{align}
    \phi_{\mathrm{LP}}(\vec{x}) = \frac{A \cdot \vec{x}}{ |(A \cdot \vec{x})|}.
\end{align}
The transformation is invariant under overall rescaling of $A$ or $\Lambda$ by a scalar. Using a comparison of eq. \ref{eq:sphere_nf_def} and eq. \ref{eq:angular_gaussian_generic}, we can deduce that the density update due to the linear project step with matrix $H$ is $\sqrt{\mathrm{Det}\left[\tilde{J}_{\Phi_{LP,H}}^{T}(x) \cdot \tilde{J}_{\Phi_{LP,H}}(x)\right]}=\mathrm{Det}(H) \cdot |H \vec{x}|^{-D}$, where we use $H=A^{-1}$ to obtain
\begin{align}
    p(\vec{x};A)=\frac{1}{S^{D-1}} \frac{|A^{-1} \cdot \vec{x}|^{-D}}{\mathrm{Det}(A)}
\end{align}
which is the standard angular gaussian in eq. \ref{eq:angular_gaussian_generic} rewritten using the matrix $A$ that actually defines the flow.

\subsection{ZLP - Combining "Z" and "LP" flows}
\label{section:combining_flows}




    \begin{figure*}[htb!]
      \centering
      \setlength{\tabcolsep}{3pt}
      \renewcommand{\arraystretch}{1.1}
      \newcommand{\thumbwidth}{0.95\linewidth}

      \begin{tabular}{%
        >{\raggedright\arraybackslash}p{0.030\linewidth}>{\raggedright\arraybackslash}p{0.030\linewidth}>{\centering\arraybackslash}m{0.097\linewidth}>{\centering\arraybackslash}m{0.063\linewidth}>{\centering\arraybackslash}m{0.063\linewidth}>{\centering\arraybackslash}m{0.063\linewidth}>{\centering\arraybackslash}m{0.063\linewidth}>{\centering\arraybackslash}m{0.063\linewidth}>{\centering\arraybackslash}m{0.063\linewidth}>{\centering\arraybackslash}m{0.063\linewidth}>{\centering\arraybackslash}m{0.063\linewidth}>{\centering\arraybackslash}m{0.063\linewidth}
      }

        \multicolumn{2}{l}{\textbf{}} &
        \multicolumn{10}{c}{\textbf{average true PDF extent [deg.]}} \\
        \cmidrule(lr){3-12}
        &  & \textbf{$180^{\circ}$} & \textbf{$83^{\circ}$} & \textbf{$38^{\circ}$} & \textbf{$18^{\circ}$} & \textbf{$8^{\circ}$} & \textbf{$3.9^{\circ}$} & \textbf{$1.8^{\circ}$} & \textbf{$0.8^{\circ}$} & \textbf{$0.4^{\circ}$} & \textbf{$0.2^{\circ}$} \\
        \midrule
\multicolumn{2}{c}{\textbf{ZLP} }   & \includegraphics[width=\thumbwidth]{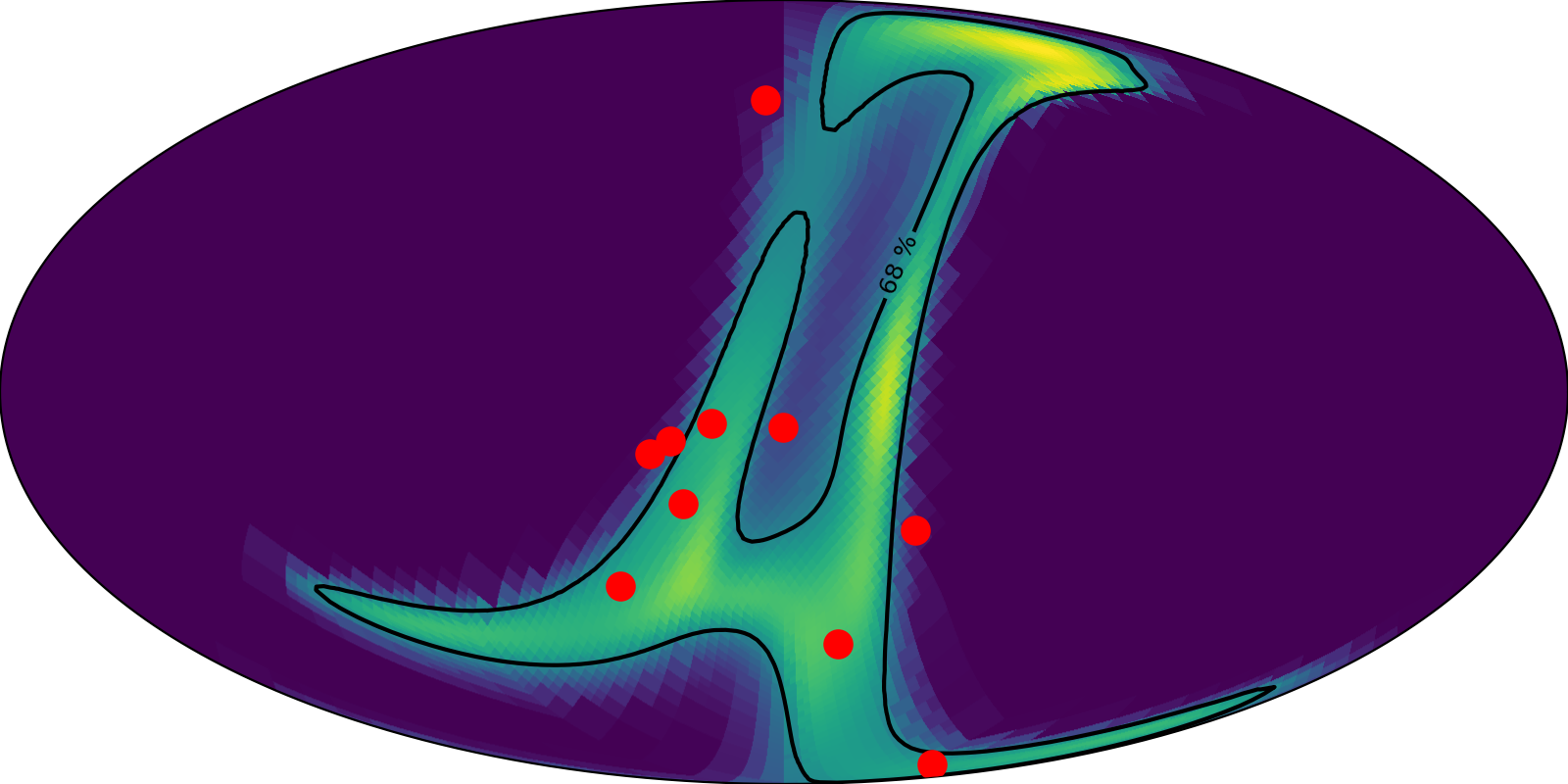} & \includegraphics[width=\thumbwidth]{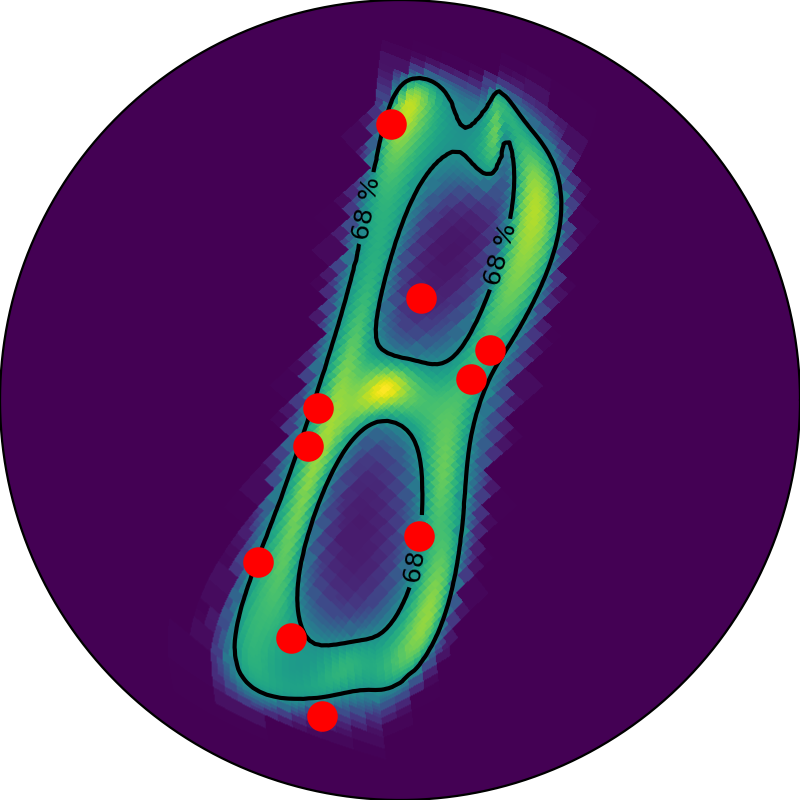} & \includegraphics[width=\thumbwidth]{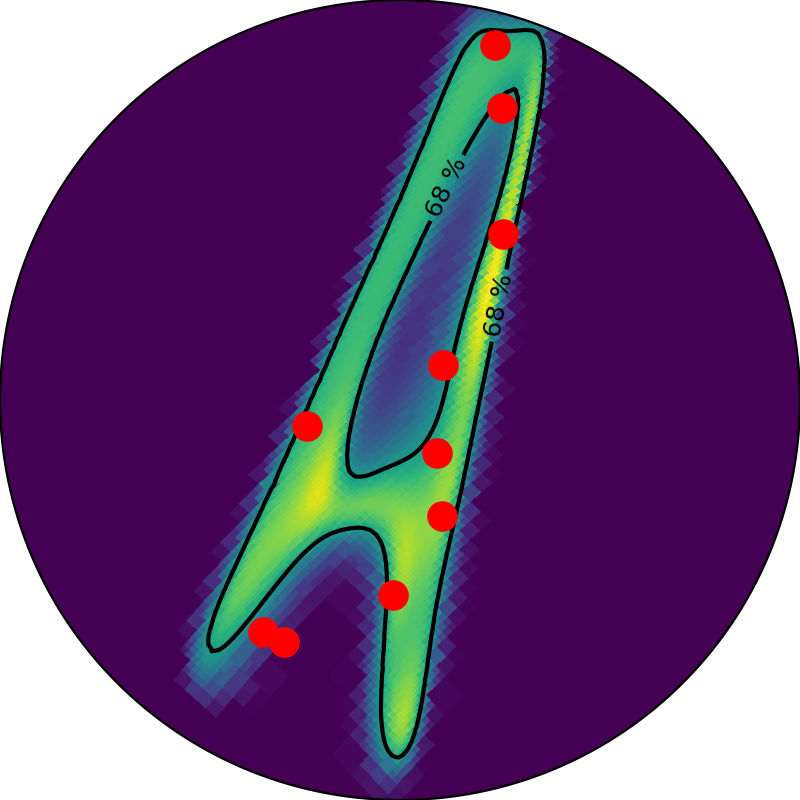} & \includegraphics[width=\thumbwidth]{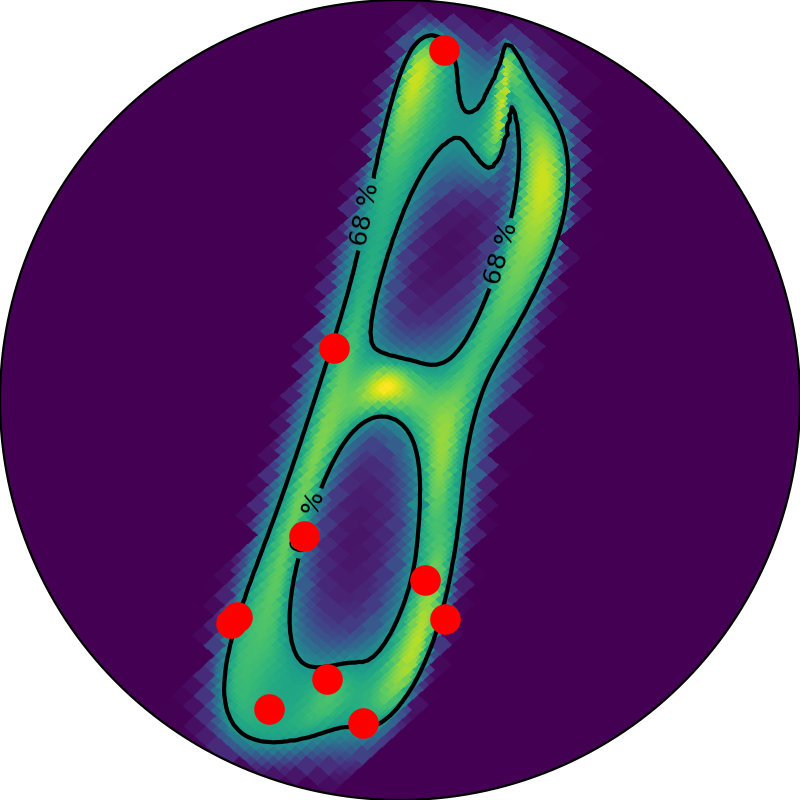} & \includegraphics[width=\thumbwidth]{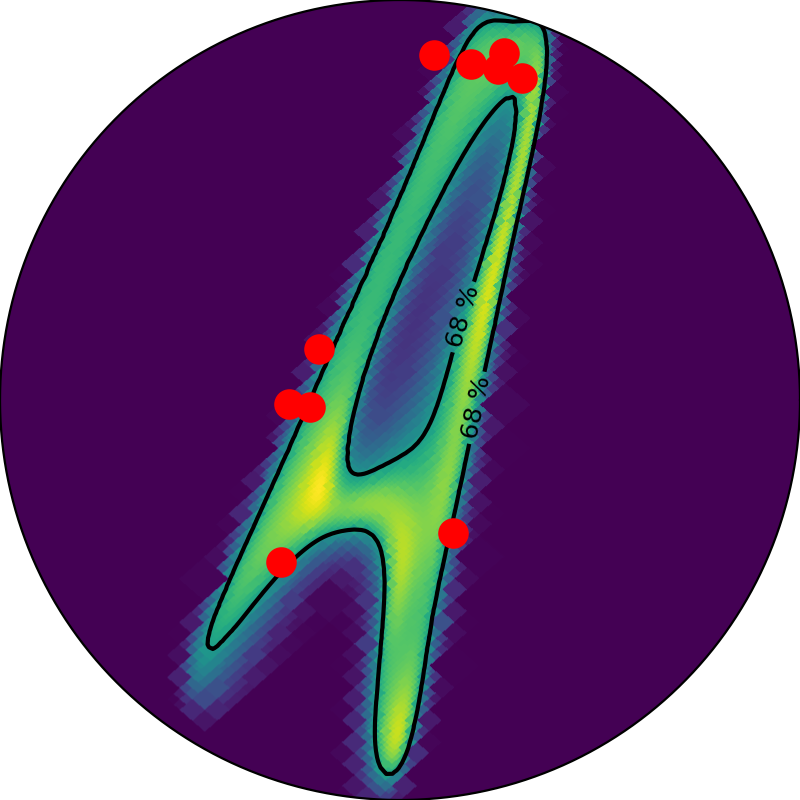} & \includegraphics[width=\thumbwidth]{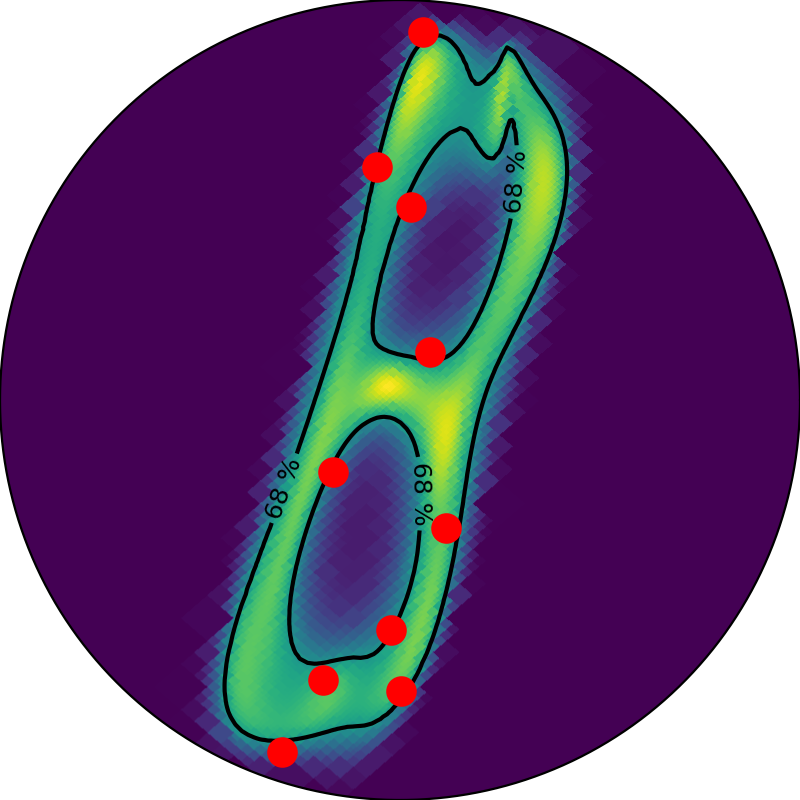} & \includegraphics[width=\thumbwidth]{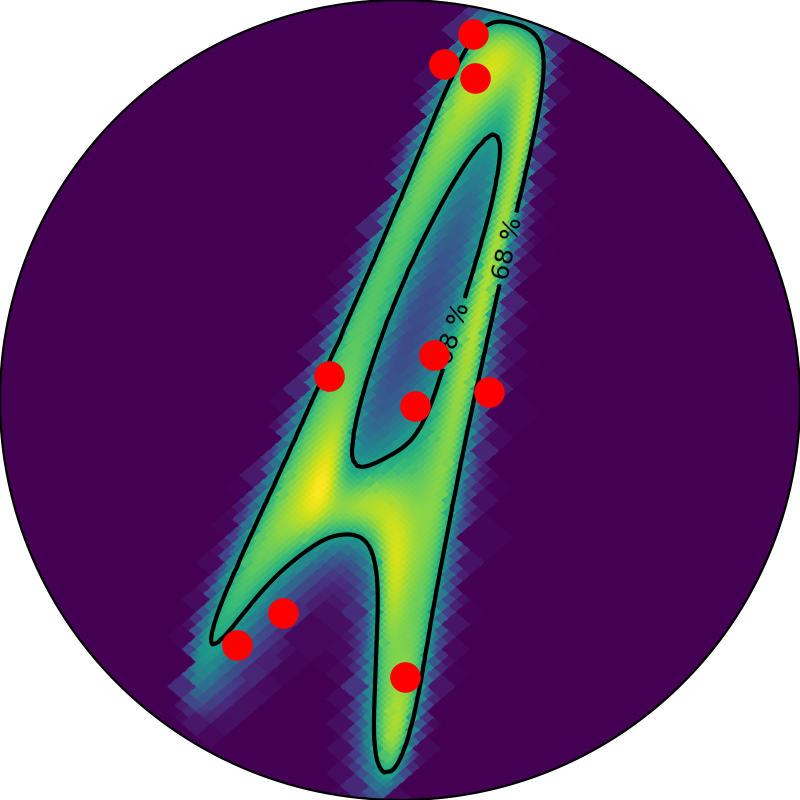} & \includegraphics[width=\thumbwidth]{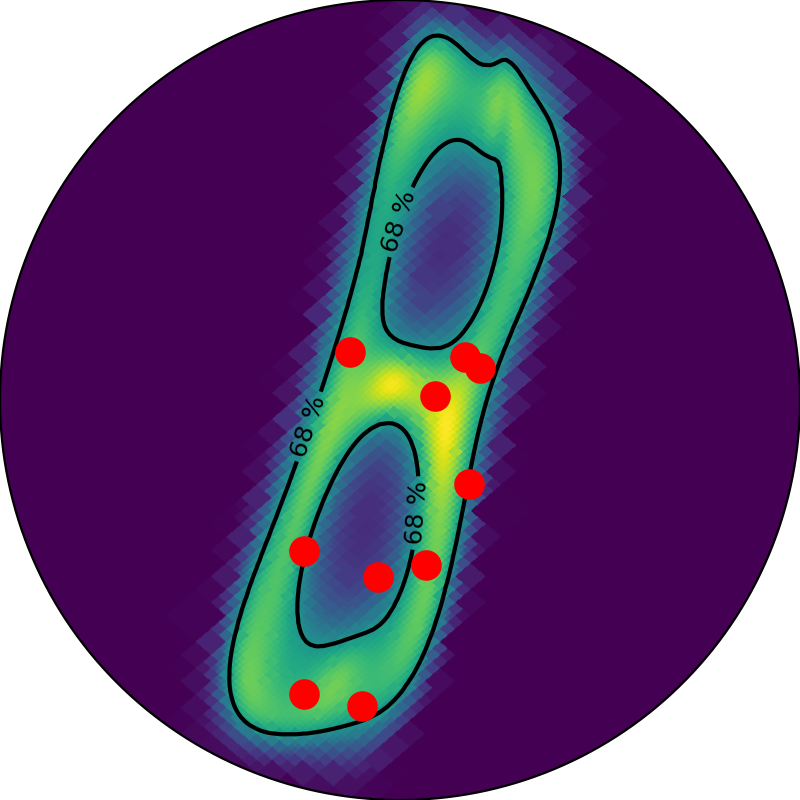} & \includegraphics[width=\thumbwidth]{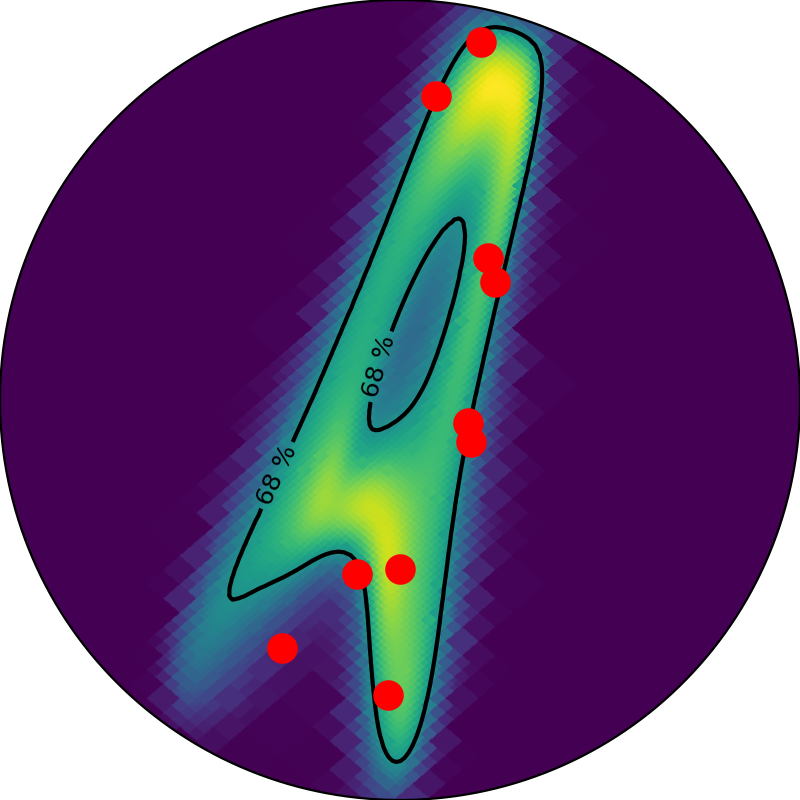} & \includegraphics[width=\thumbwidth]{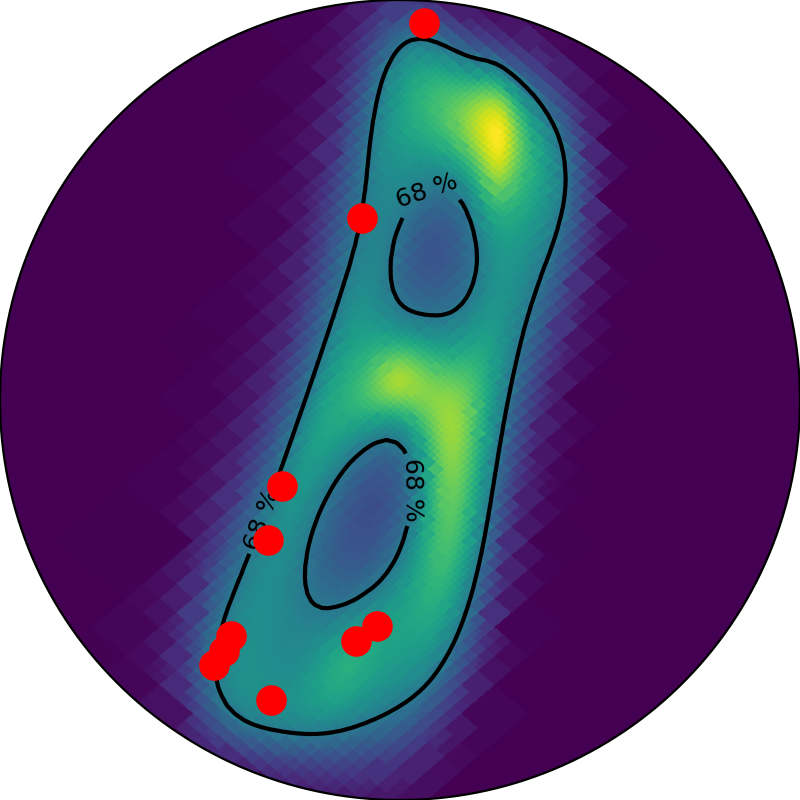} \\
\midrule
\multicolumn{2}{c}{\textbf{RQS-M (big)} }   & \includegraphics[width=\thumbwidth]{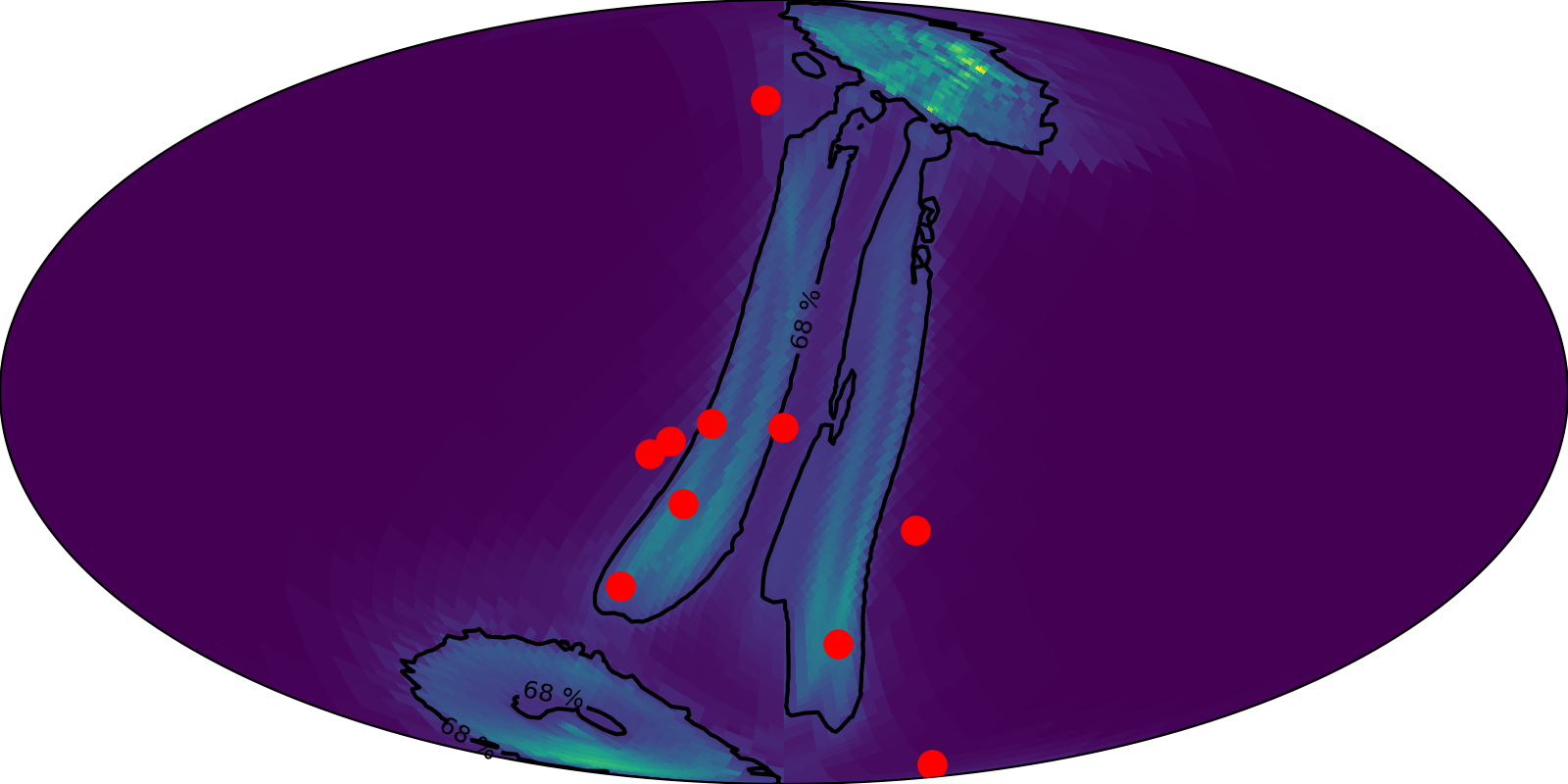} & \includegraphics[width=\thumbwidth]{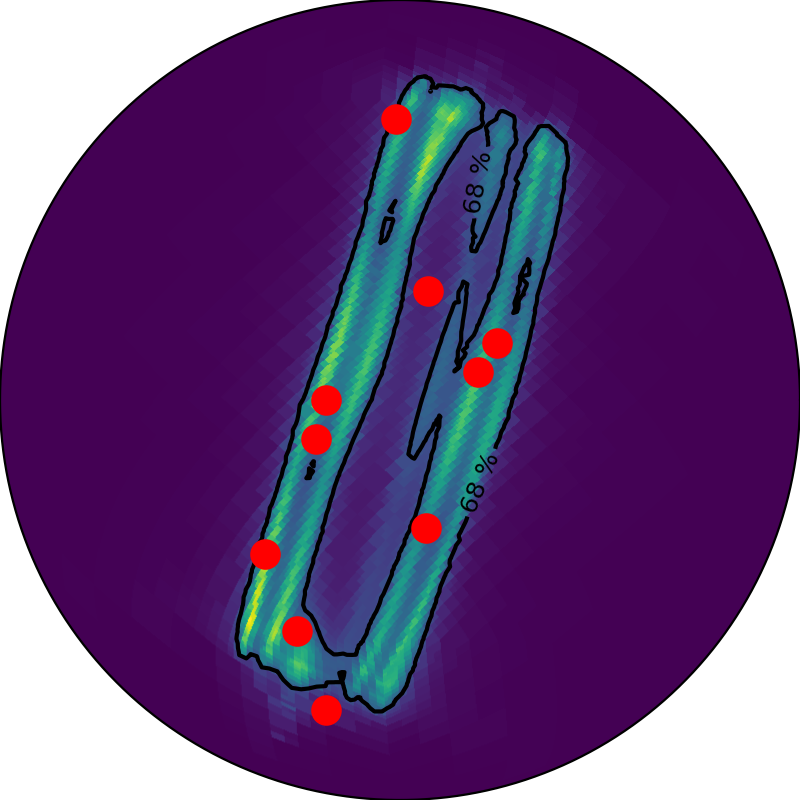} & \includegraphics[width=\thumbwidth]{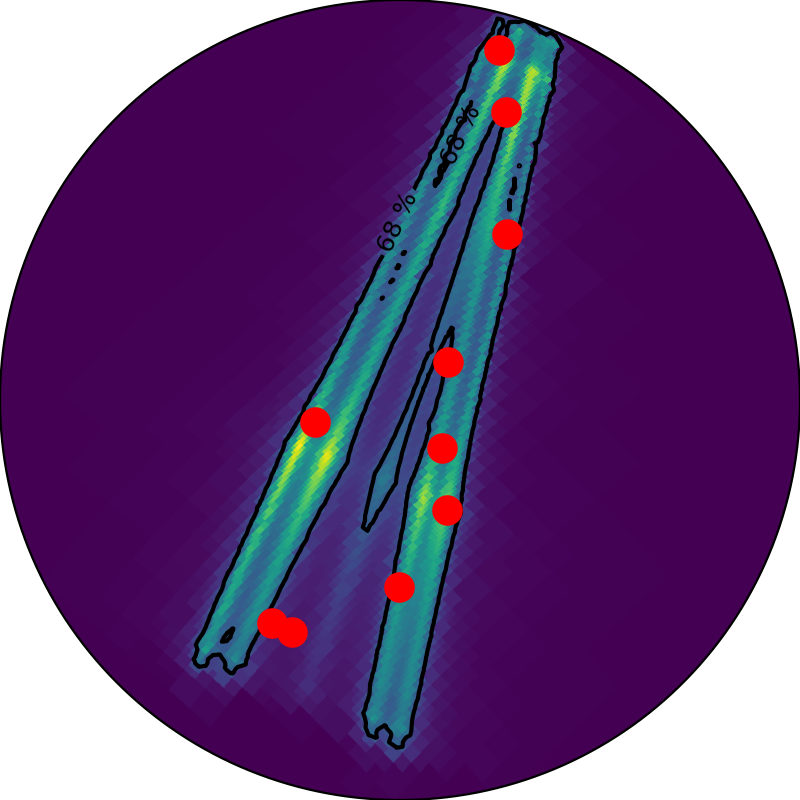} & \includegraphics[width=\thumbwidth]{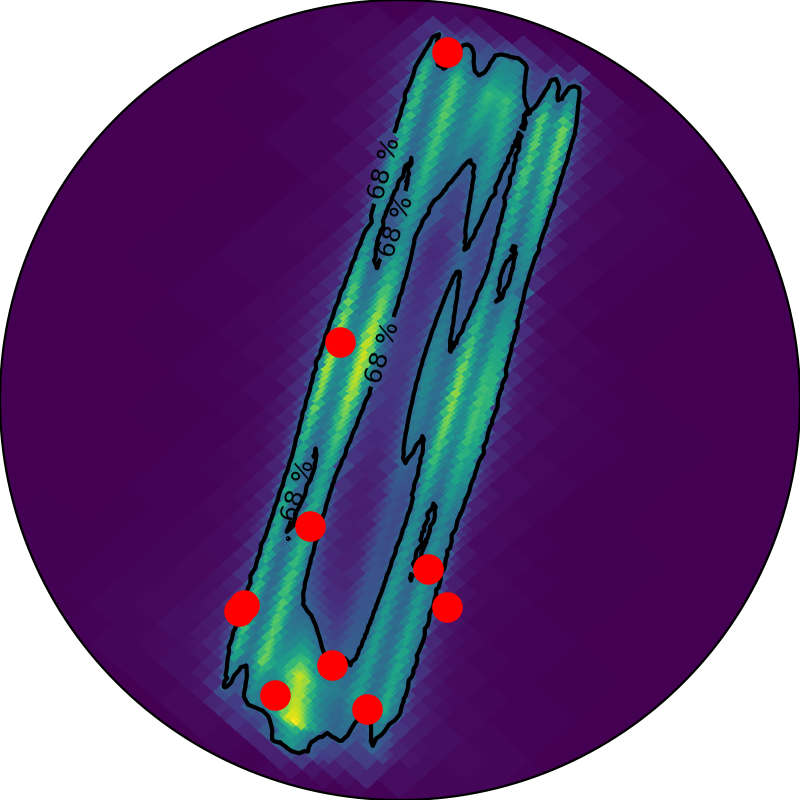} & \includegraphics[width=\thumbwidth]{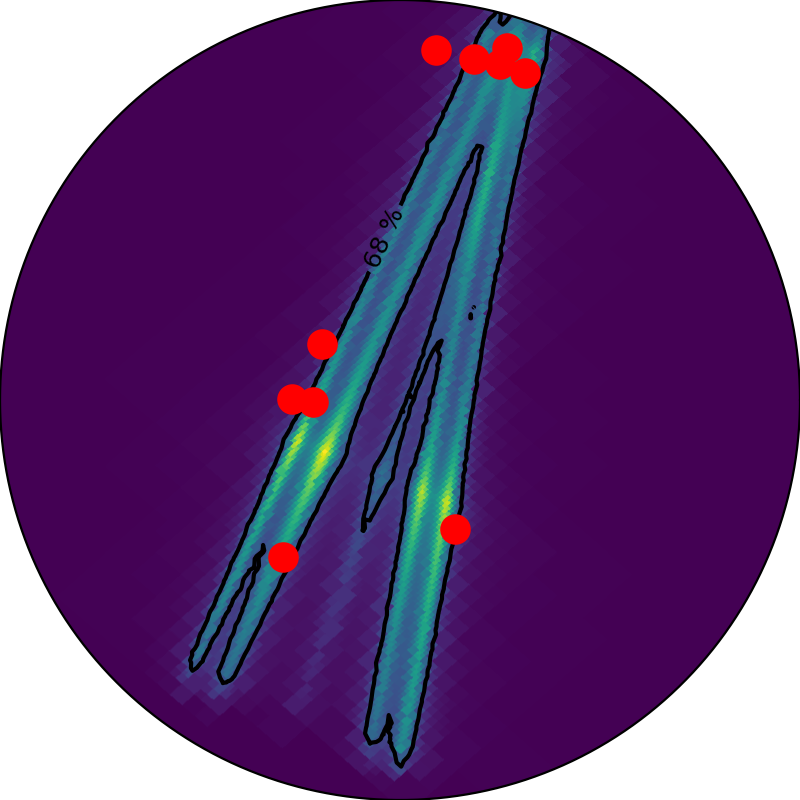} & \includegraphics[width=\thumbwidth]{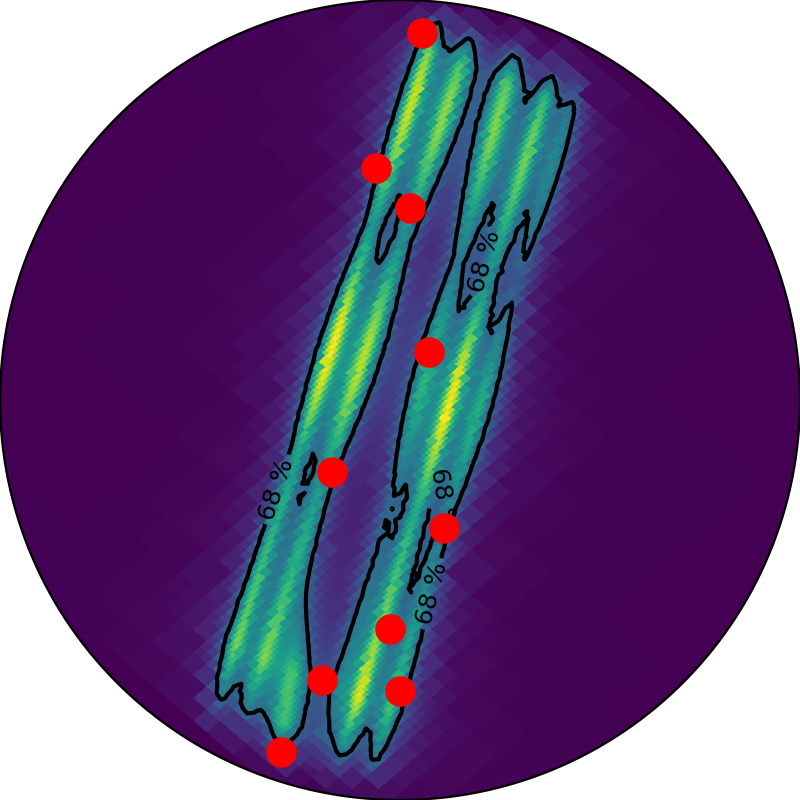} & \includegraphics[width=\thumbwidth]{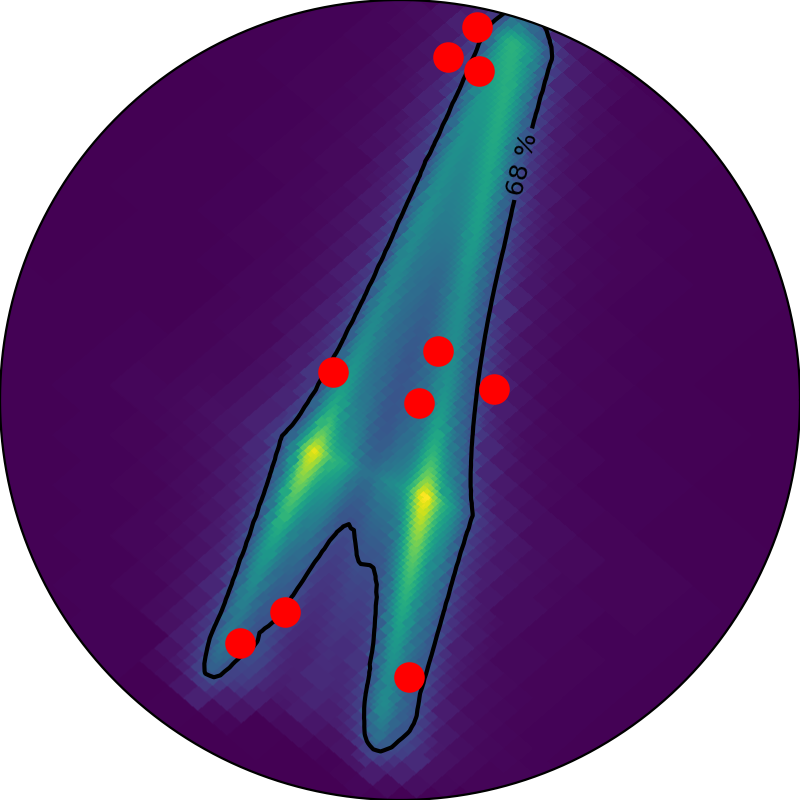} & \includegraphics[width=\thumbwidth]{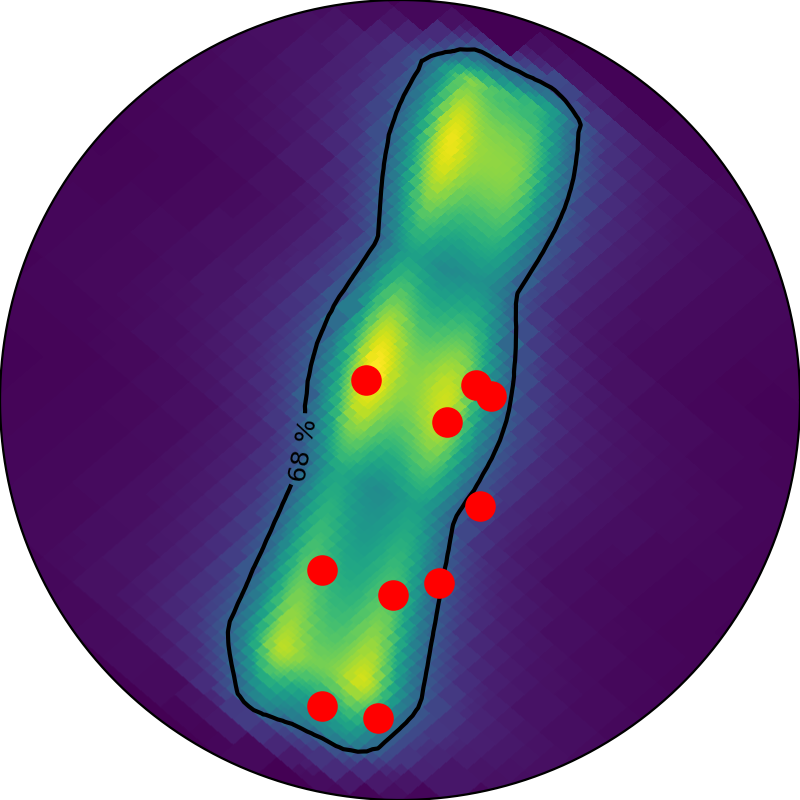} & \includegraphics[width=\thumbwidth]{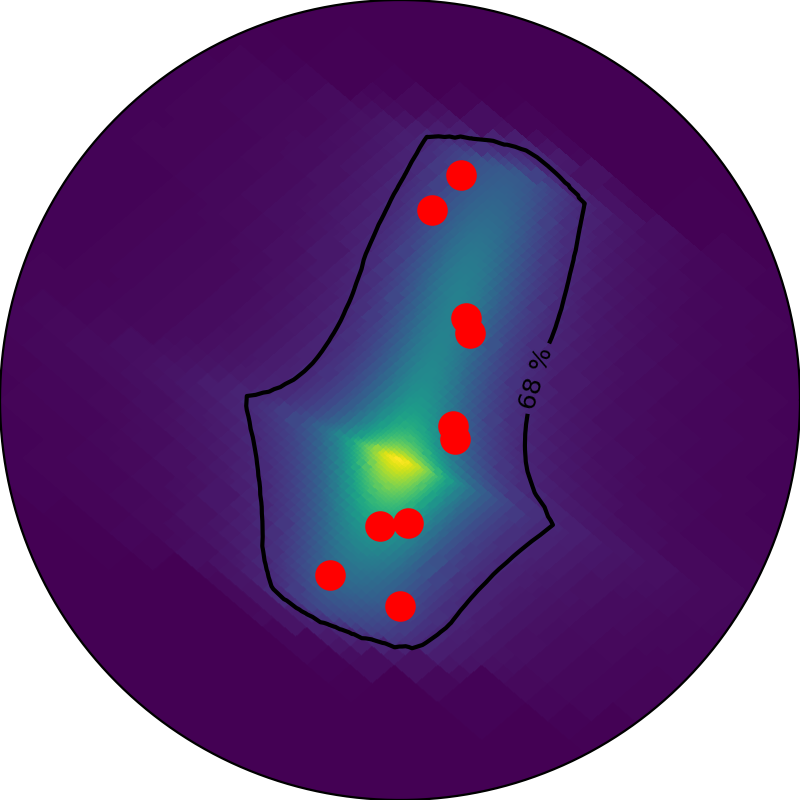} & \includegraphics[width=\thumbwidth]{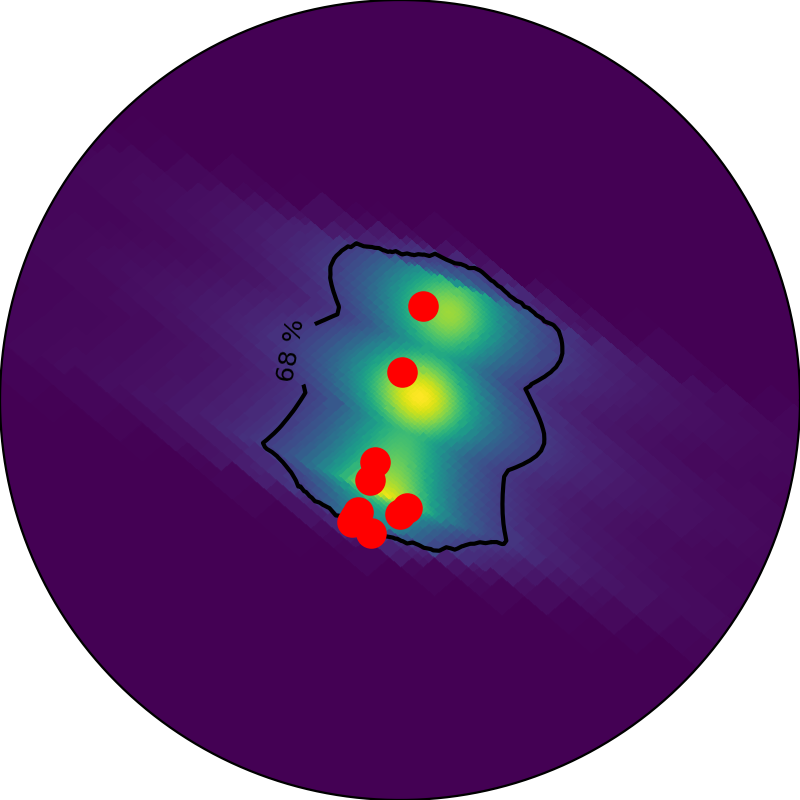} \\
\multicolumn{2}{c}{\textbf{RQS-M (big) + K} } & \includegraphics[width=\thumbwidth]{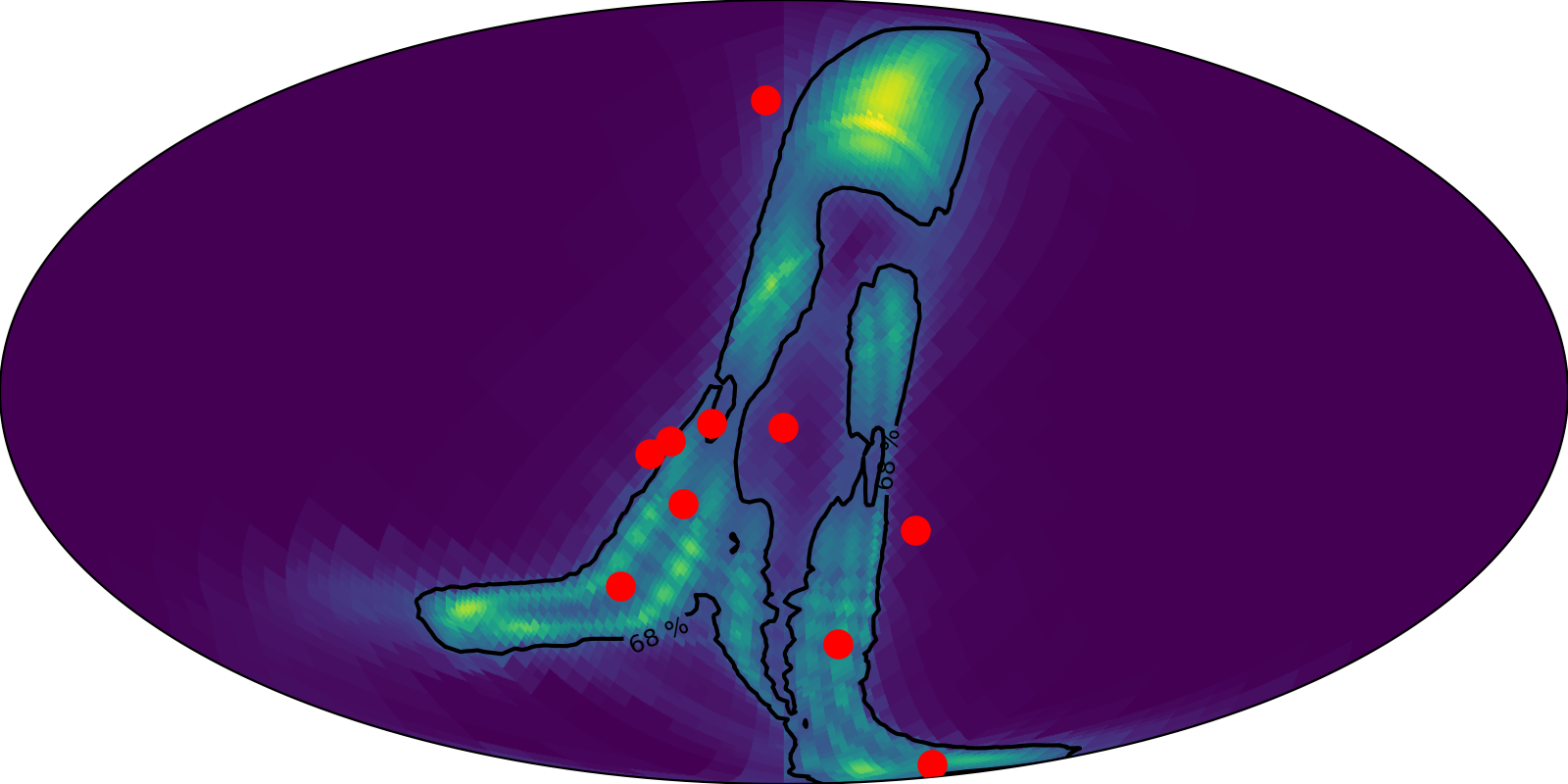} & \includegraphics[width=\thumbwidth]{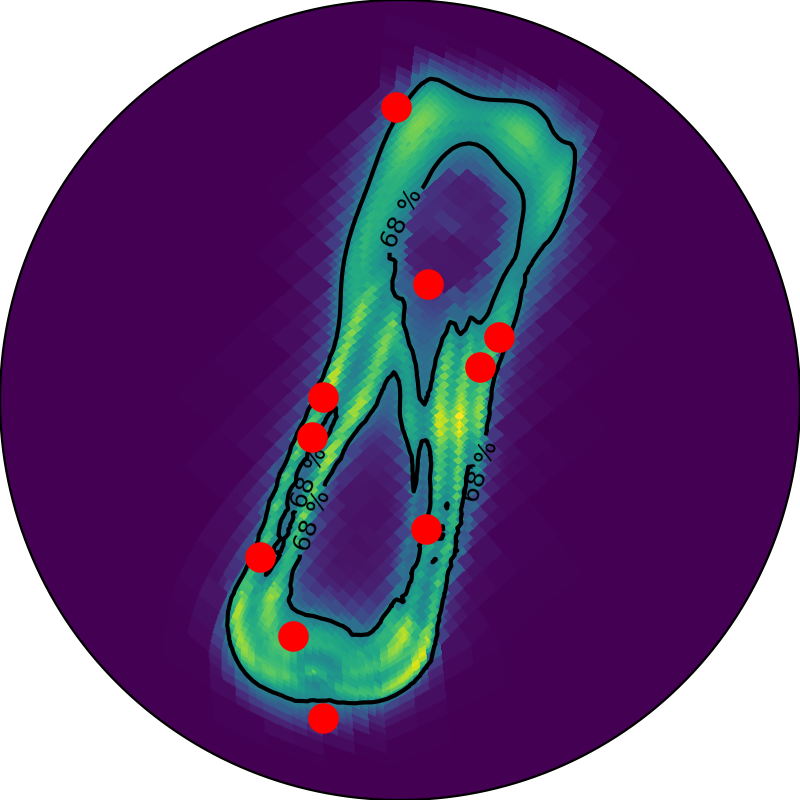} & \includegraphics[width=\thumbwidth]{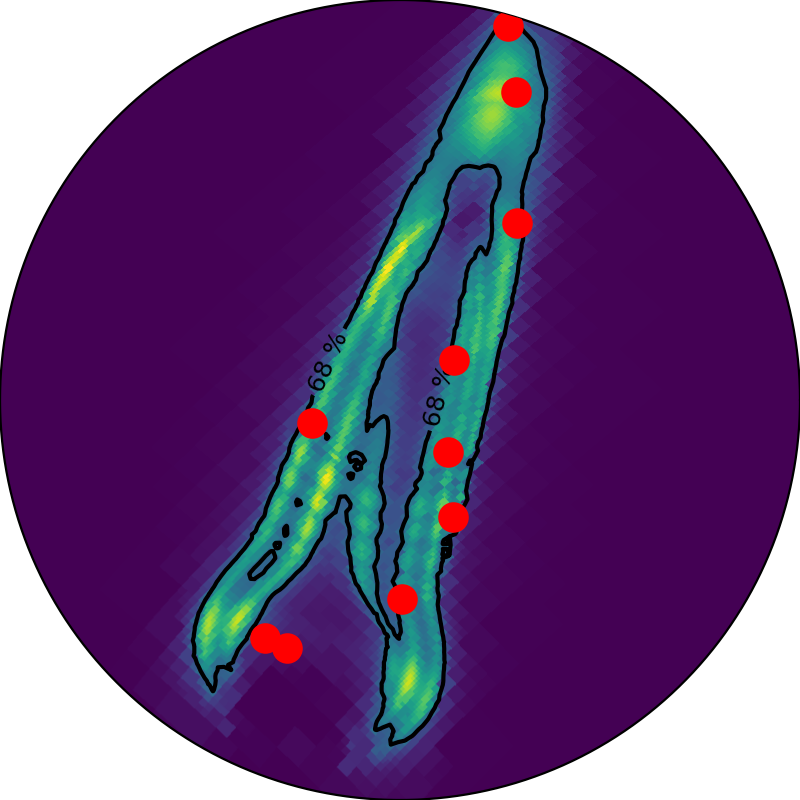} & \includegraphics[width=\thumbwidth]{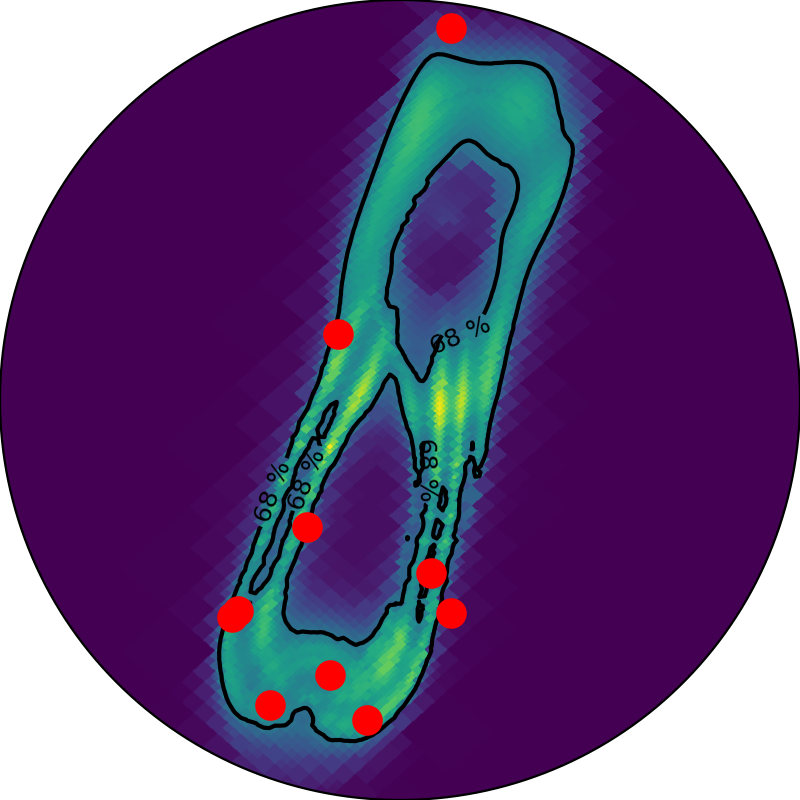} & \includegraphics[width=\thumbwidth]{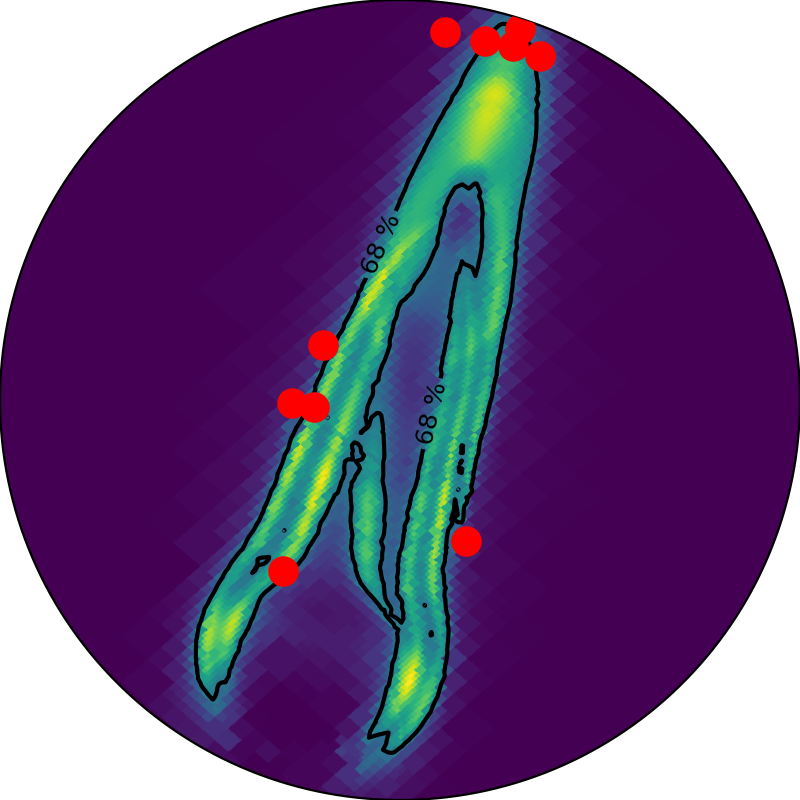} & \includegraphics[width=\thumbwidth]{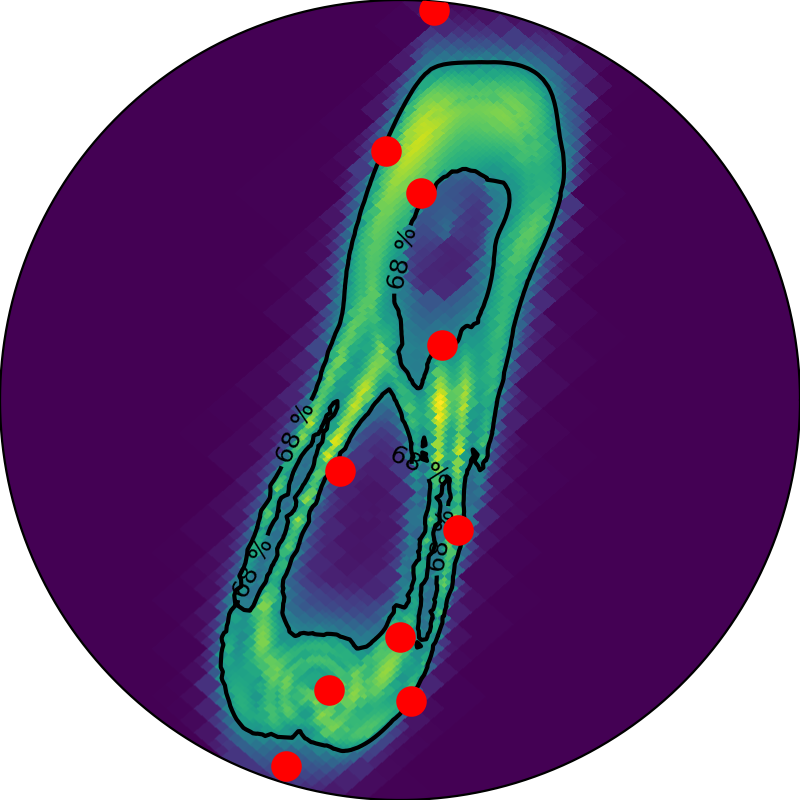} & \includegraphics[width=\thumbwidth]{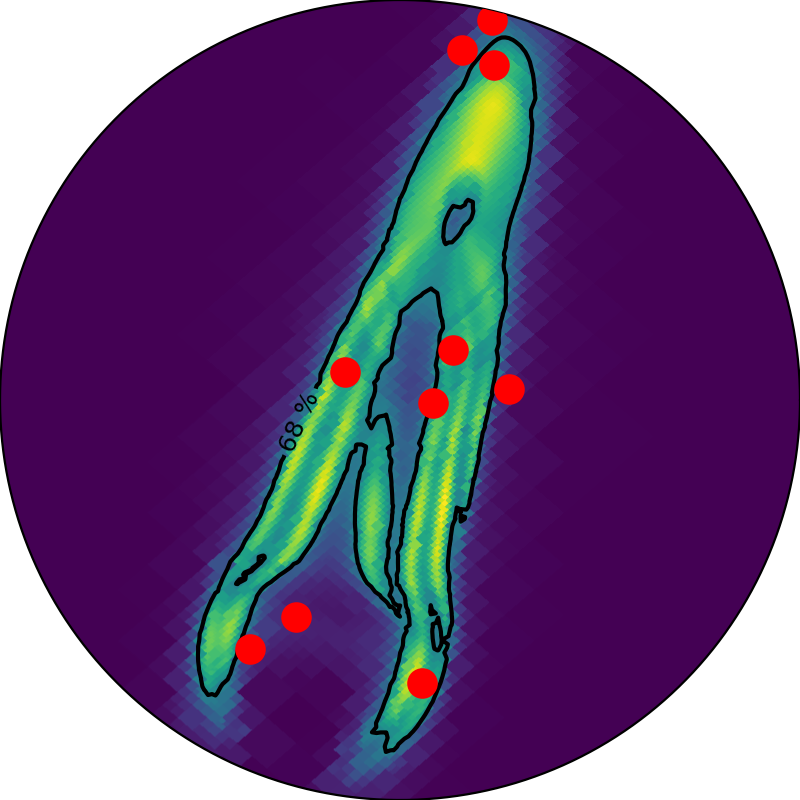} & \includegraphics[width=\thumbwidth]{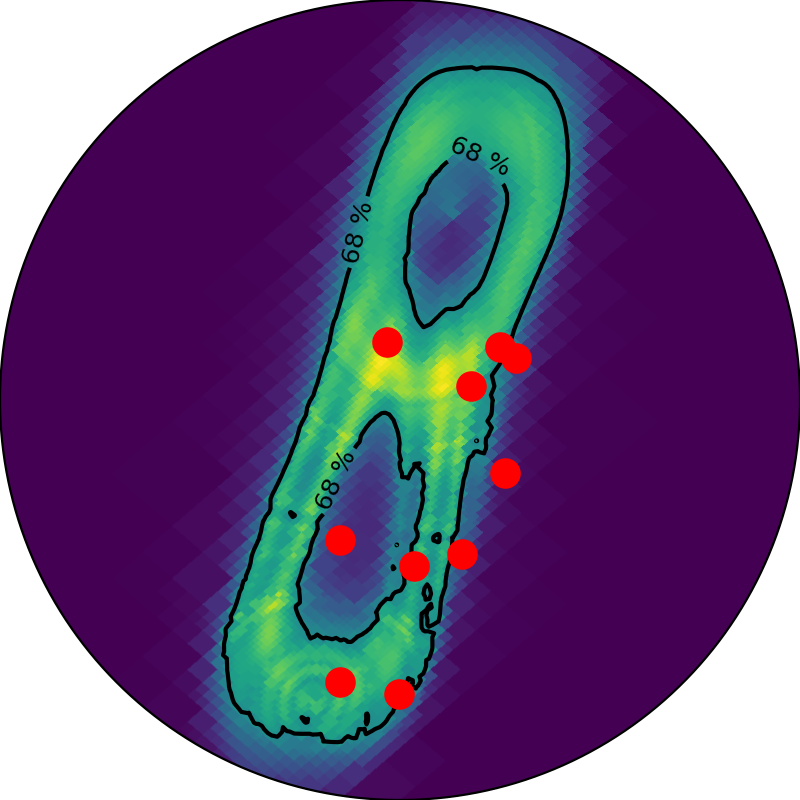} & \includegraphics[width=\thumbwidth]{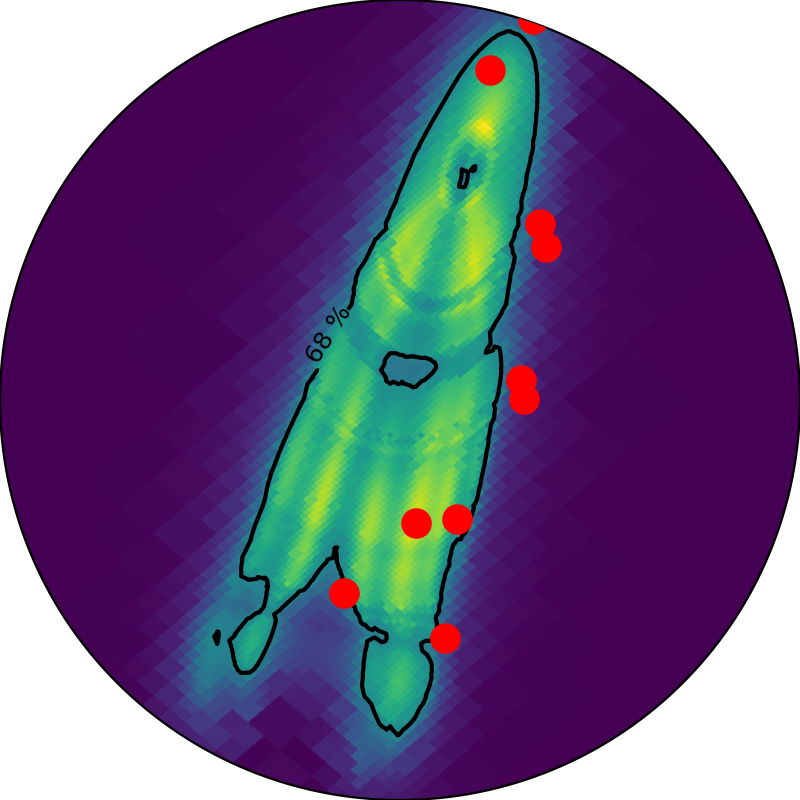} & \includegraphics[width=\thumbwidth]{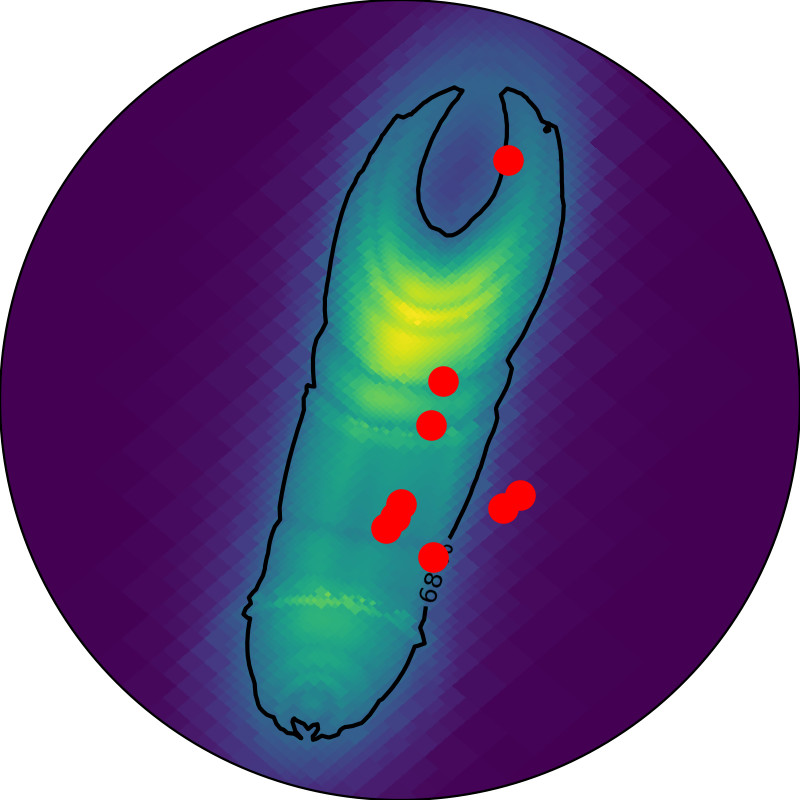} \\
\midrule
\multicolumn{2}{c}{\textbf{EXP-R} }   & \includegraphics[width=\thumbwidth]{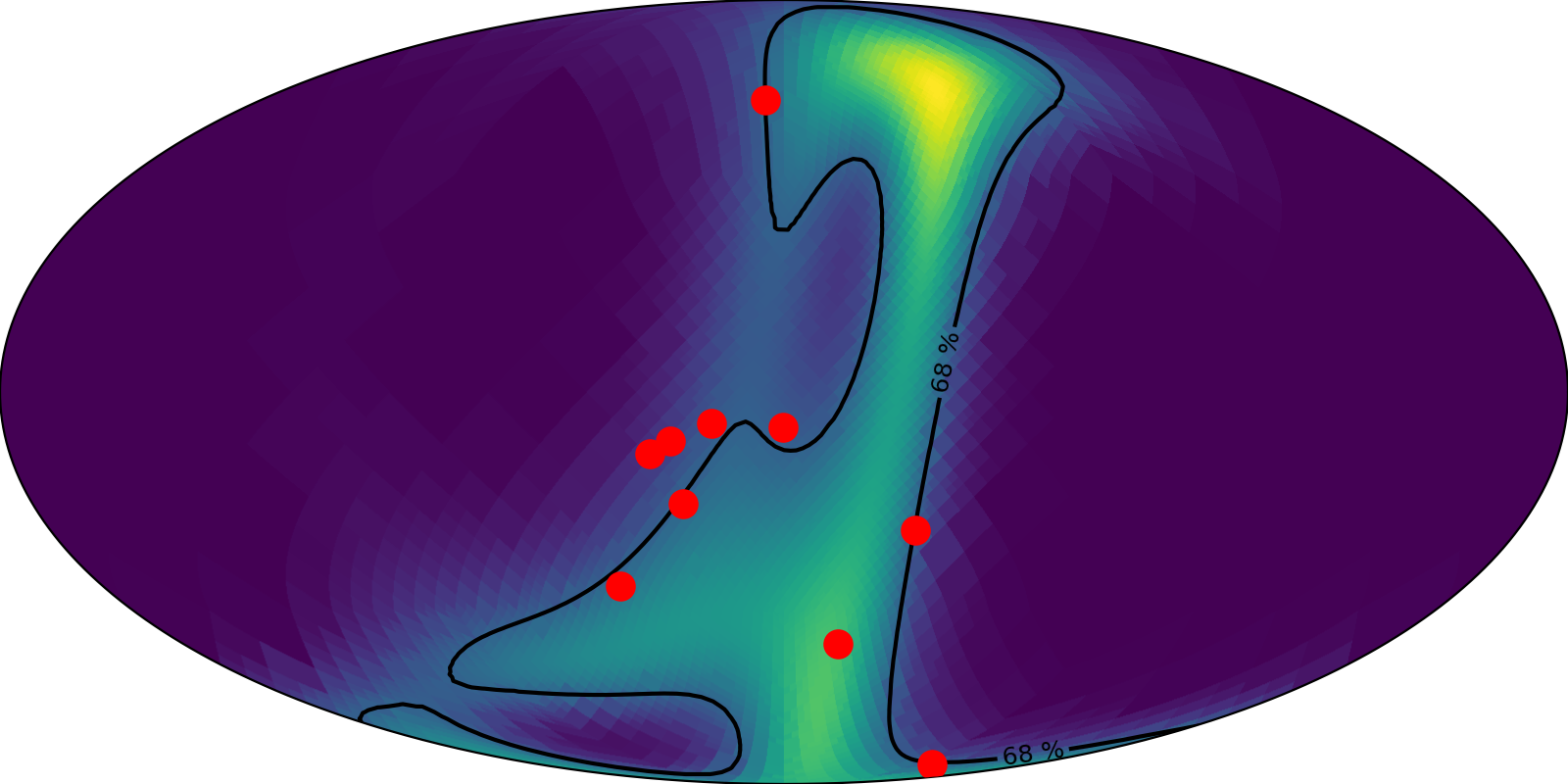} & \includegraphics[width=\thumbwidth]{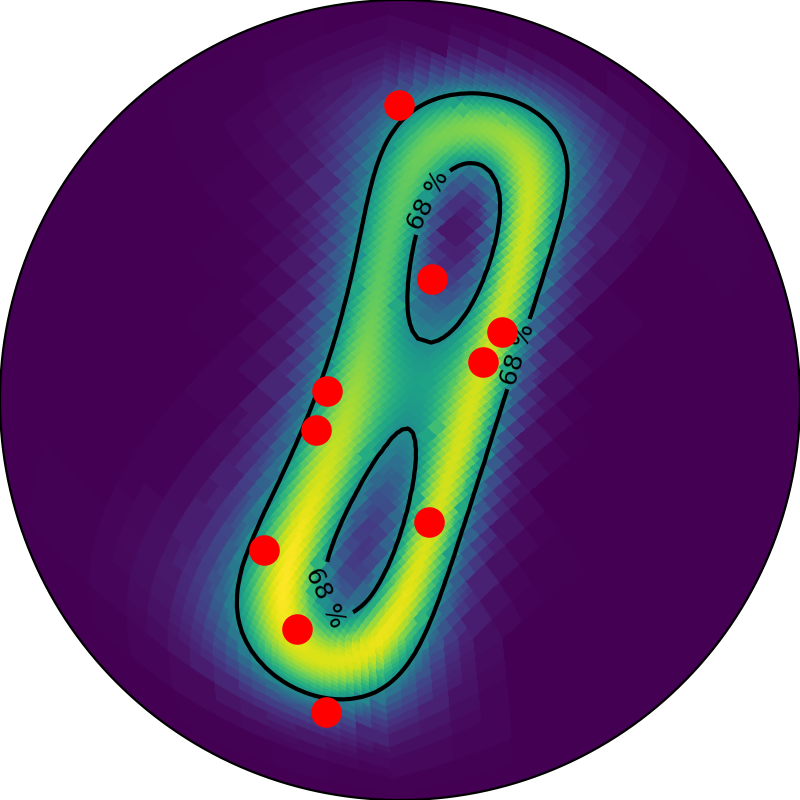} & \includegraphics[width=\thumbwidth]{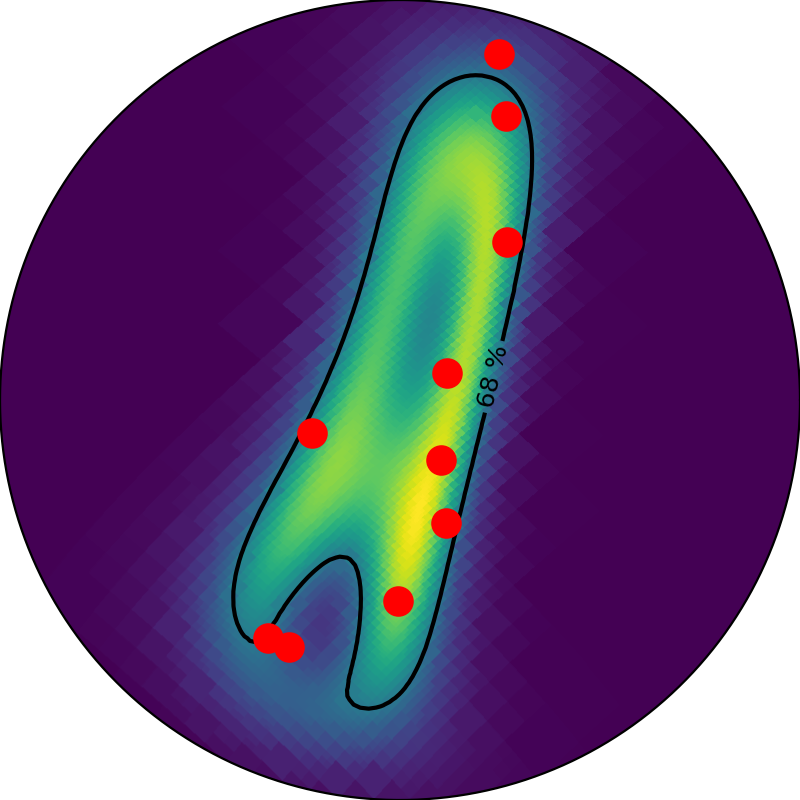} & \includegraphics[width=\thumbwidth]{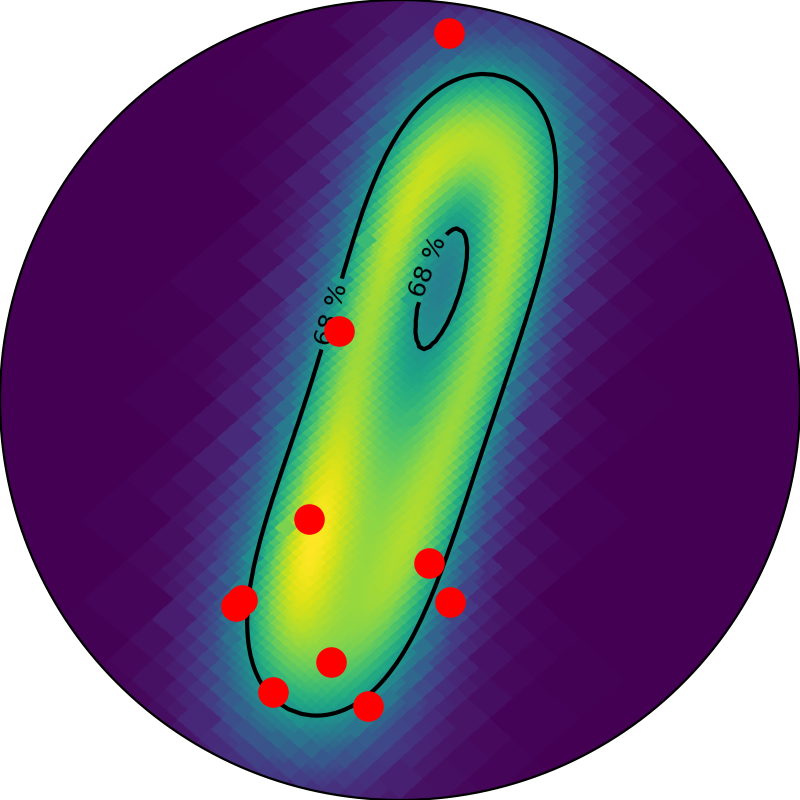} & \includegraphics[width=\thumbwidth]{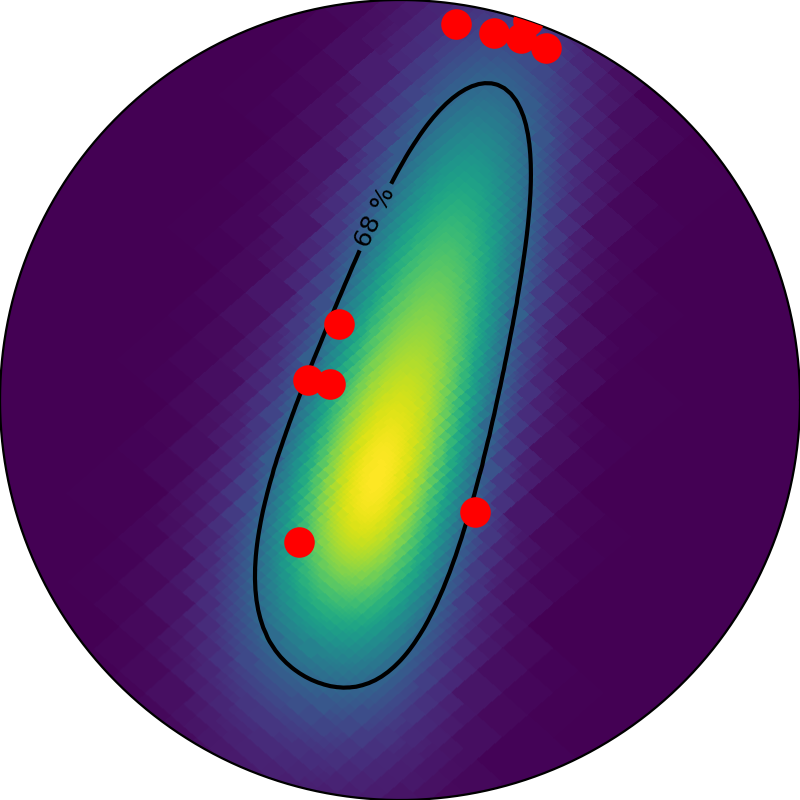} & \includegraphics[width=\thumbwidth]{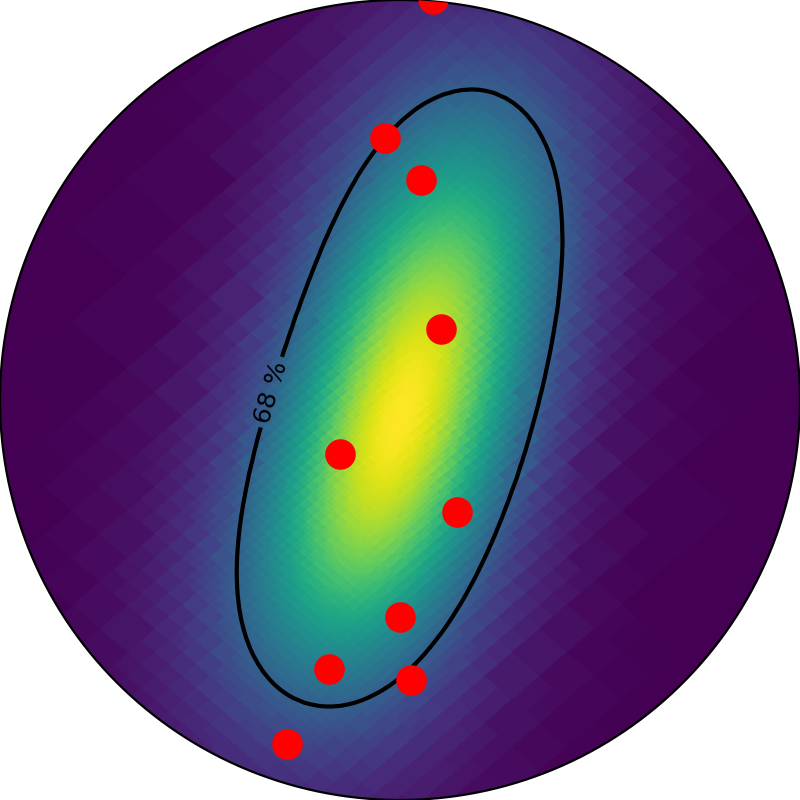} & \includegraphics[width=\thumbwidth]{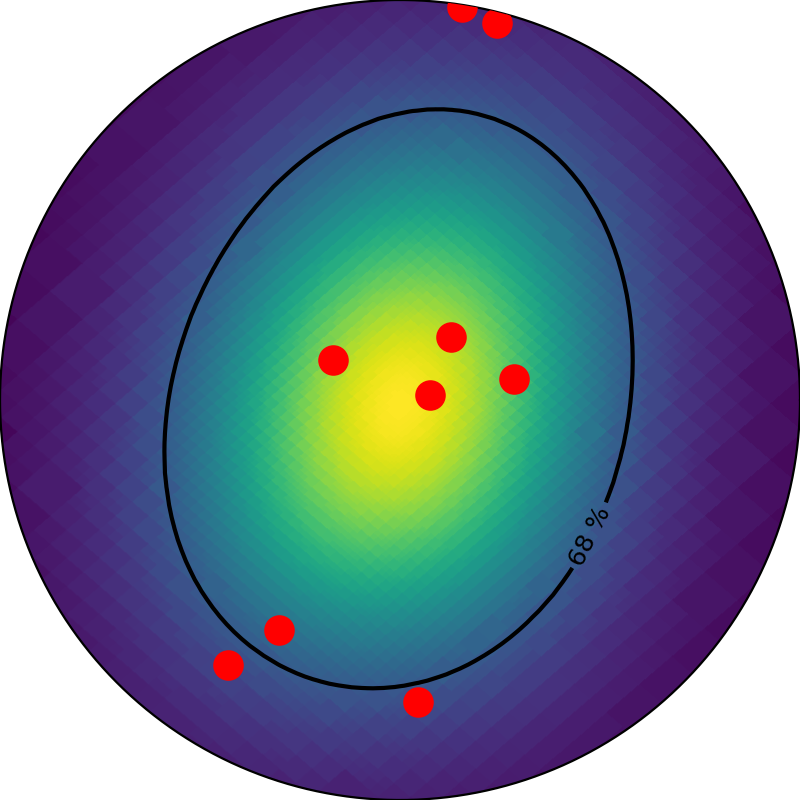} & \includegraphics[width=\thumbwidth]{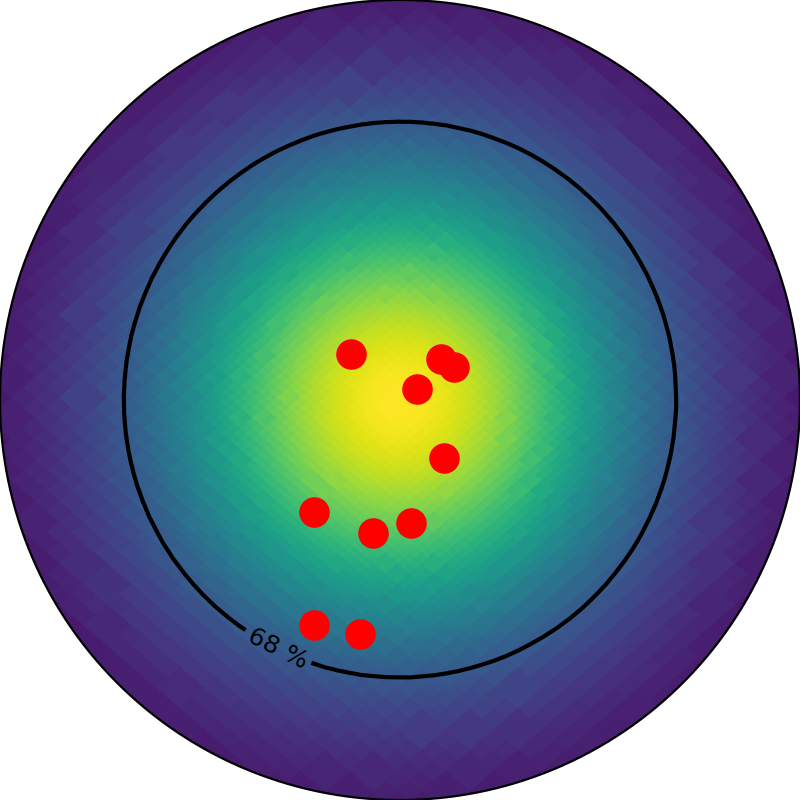} & \includegraphics[width=\thumbwidth]{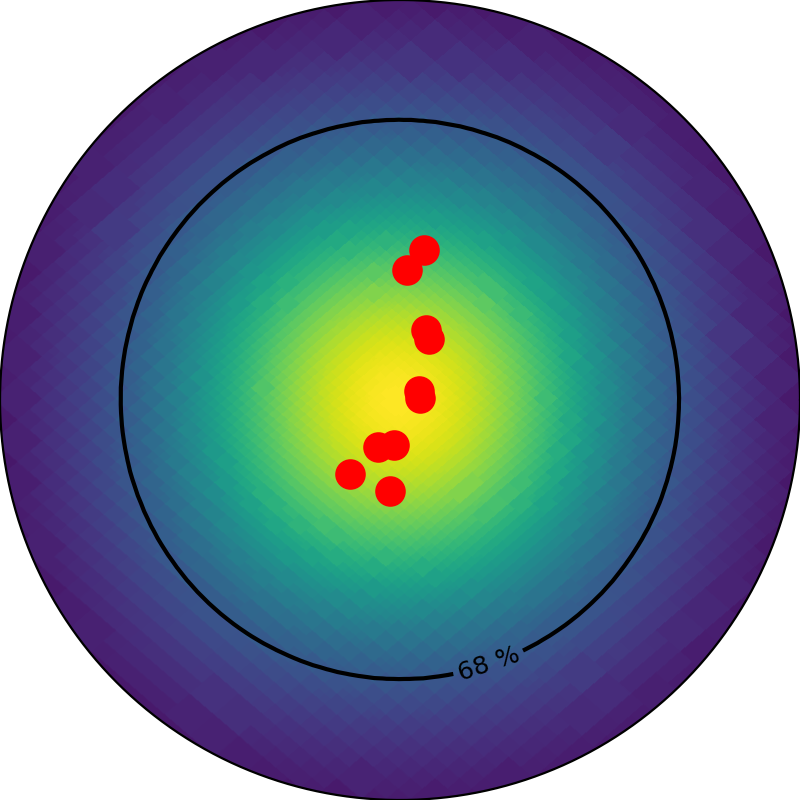} & \includegraphics[width=\thumbwidth]{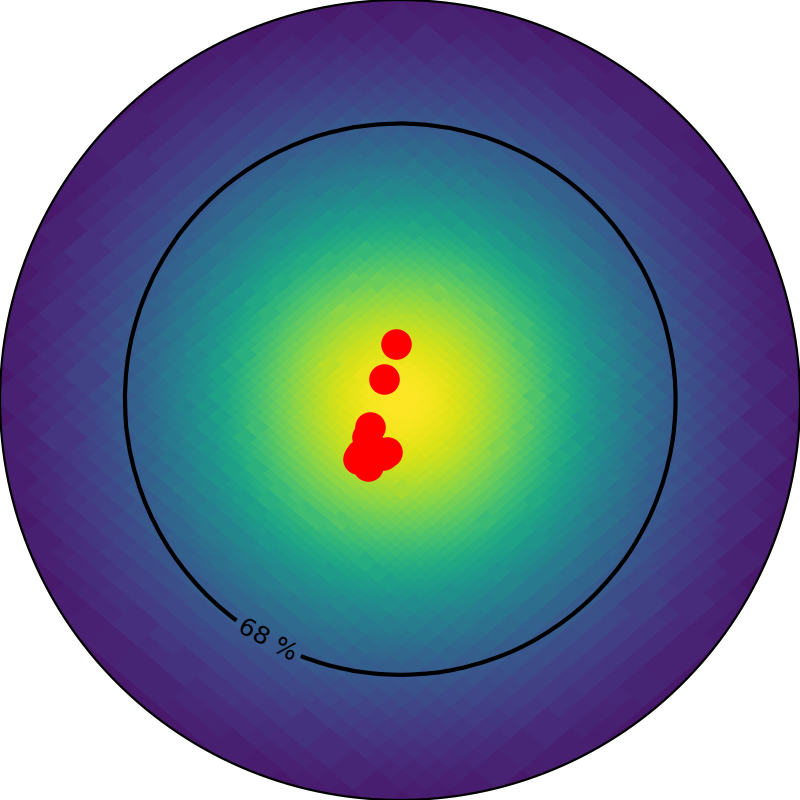} \\
\multicolumn{2}{c}{\textbf{EXP-R + K} } & \includegraphics[width=\thumbwidth]{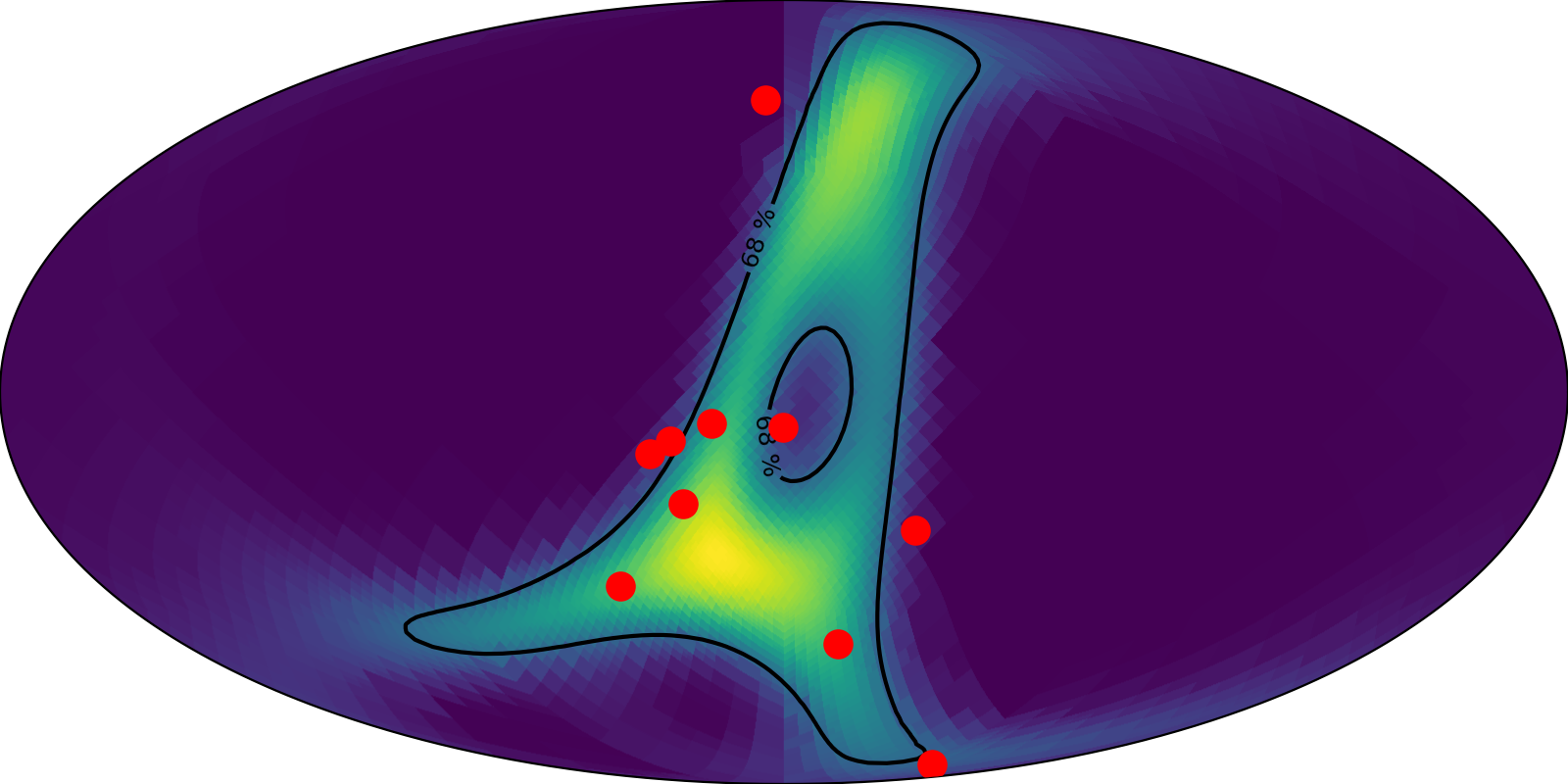} & \includegraphics[width=\thumbwidth]{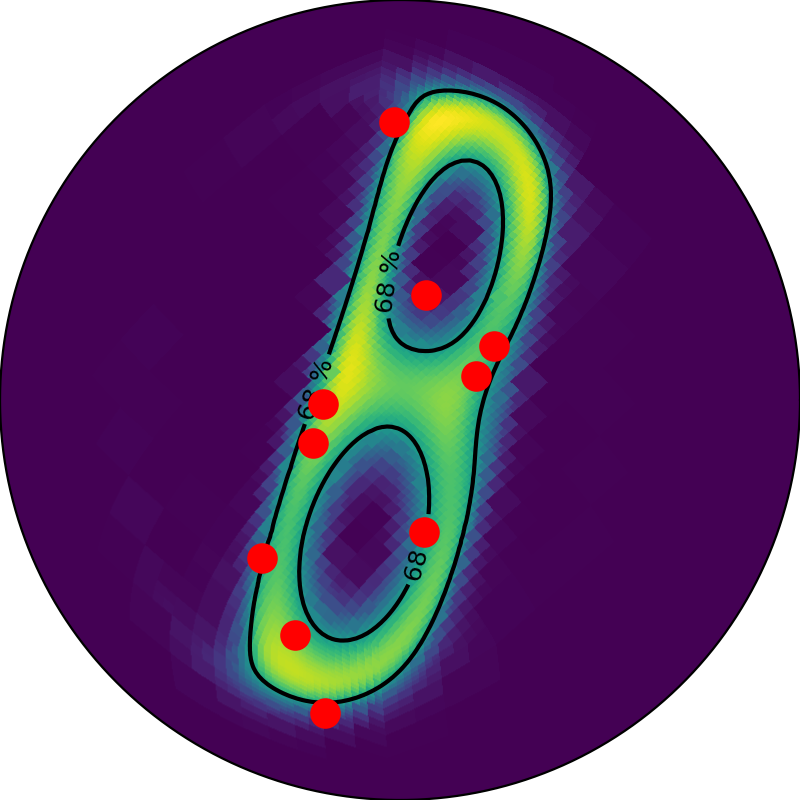} & \includegraphics[width=\thumbwidth]{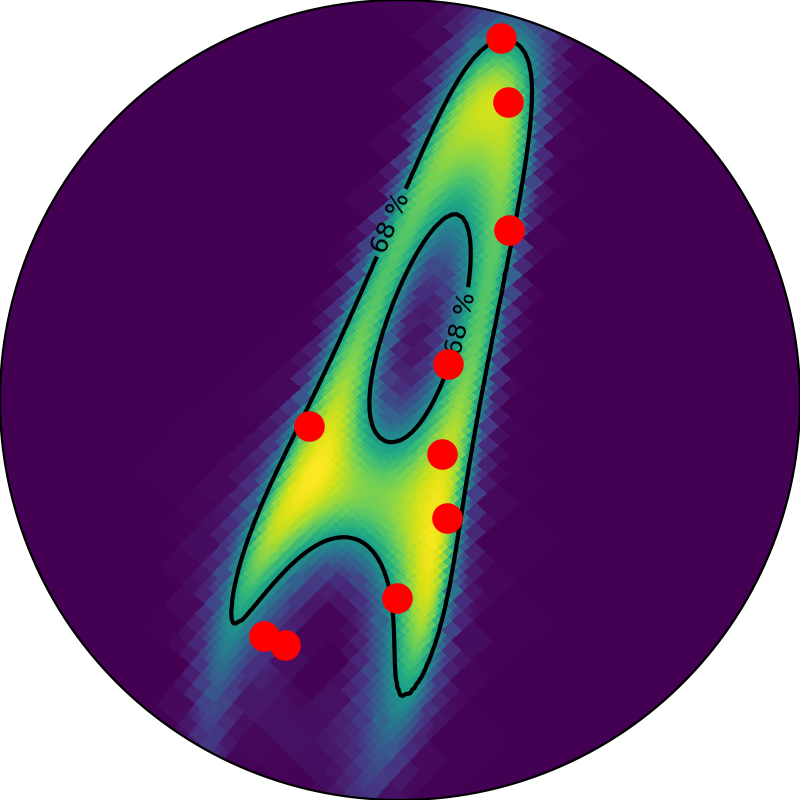} & \includegraphics[width=\thumbwidth]{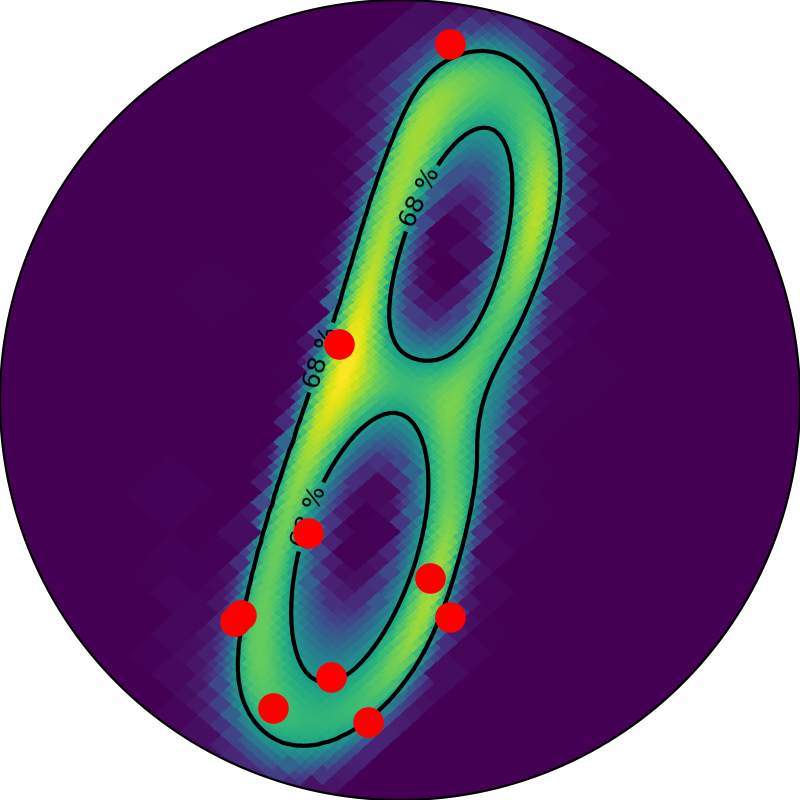} & \includegraphics[width=\thumbwidth]{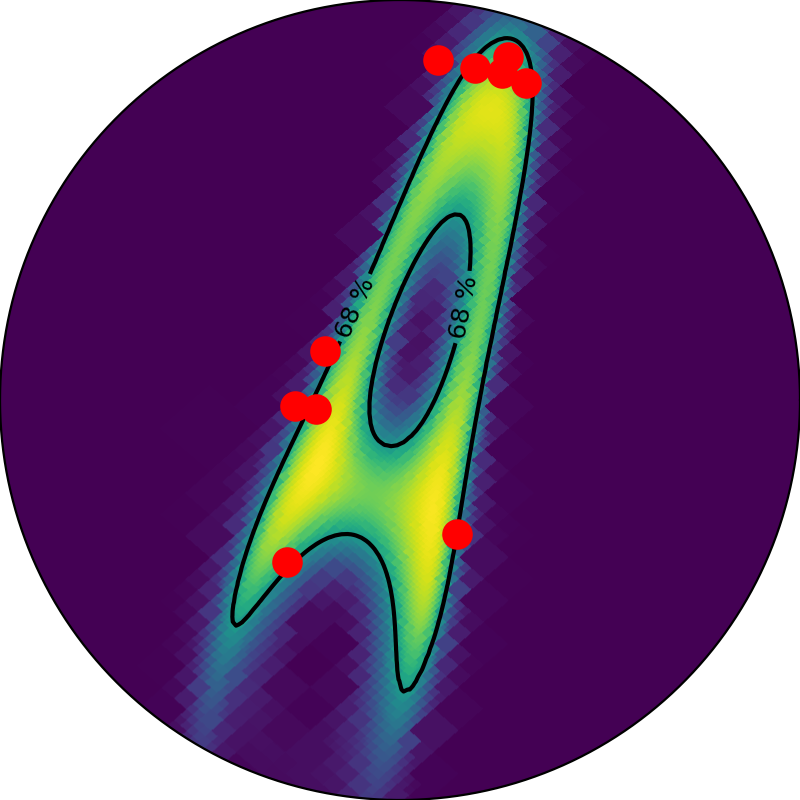} & \includegraphics[width=\thumbwidth]{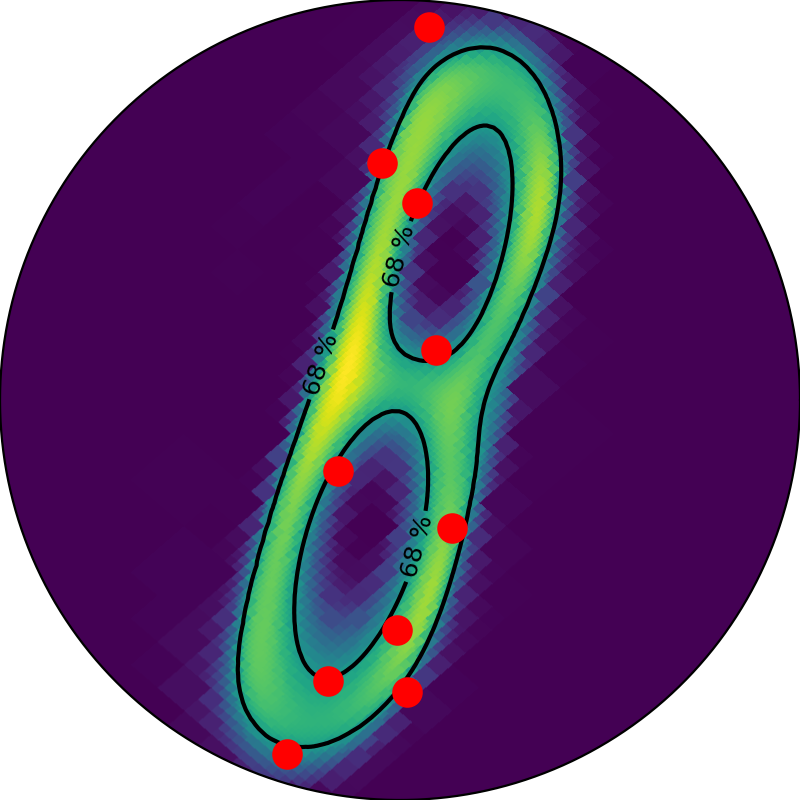} & \includegraphics[width=\thumbwidth]{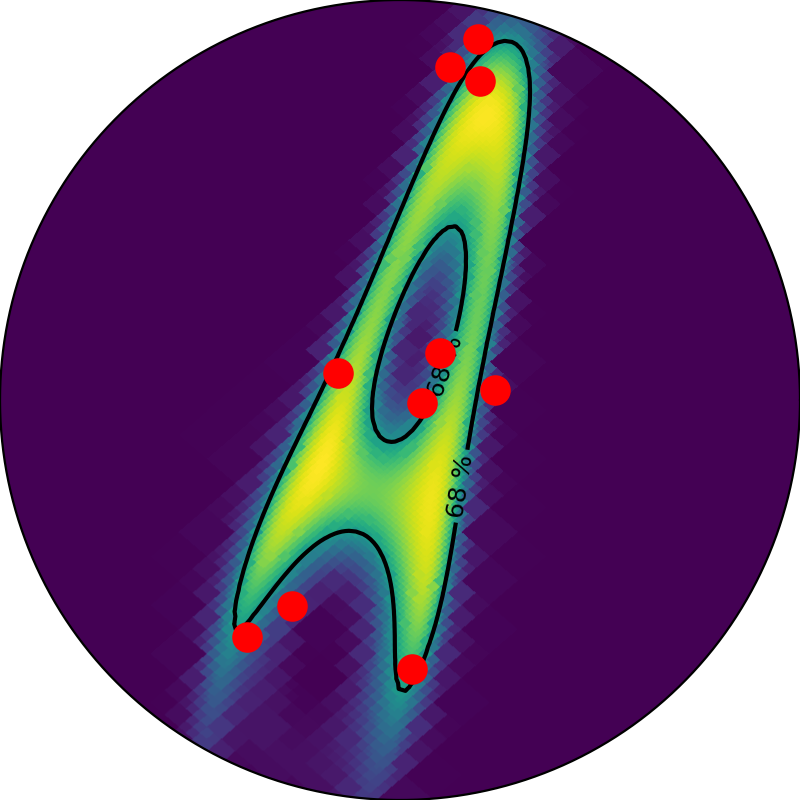} & \includegraphics[width=\thumbwidth]{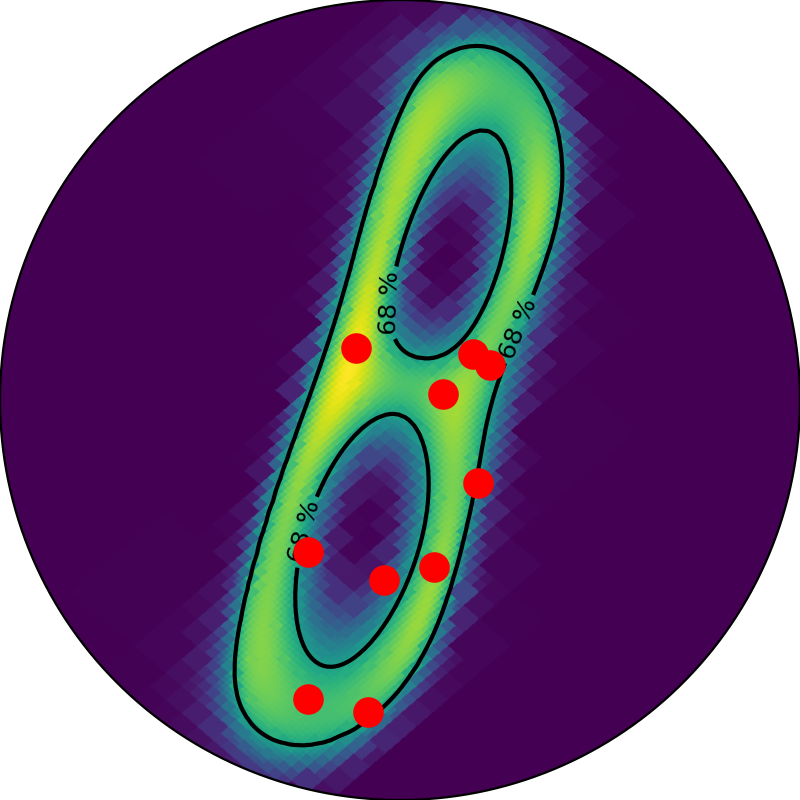} & \includegraphics[width=\thumbwidth]{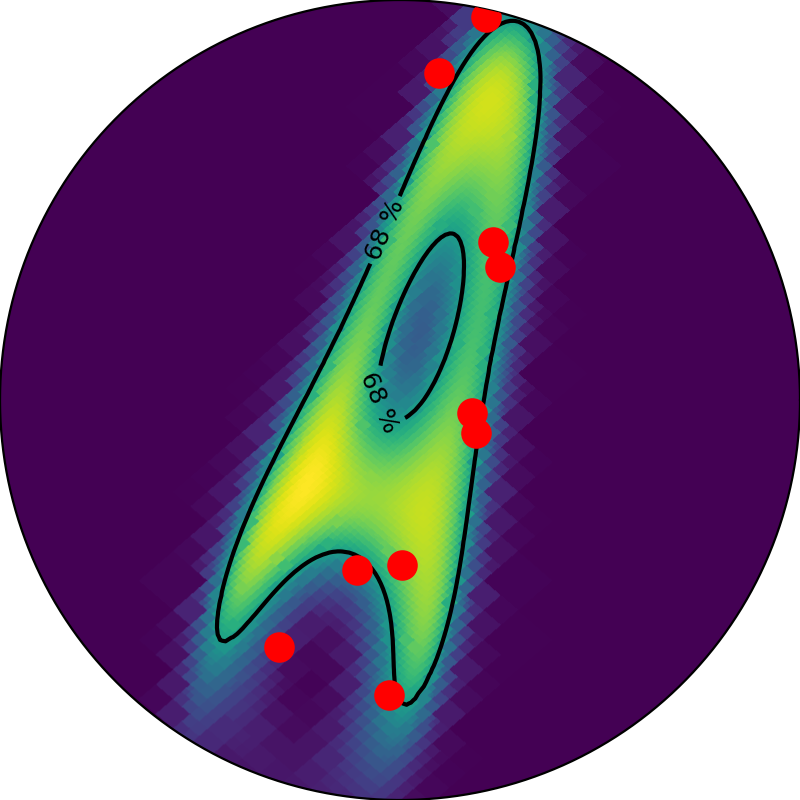} & \includegraphics[width=\thumbwidth]{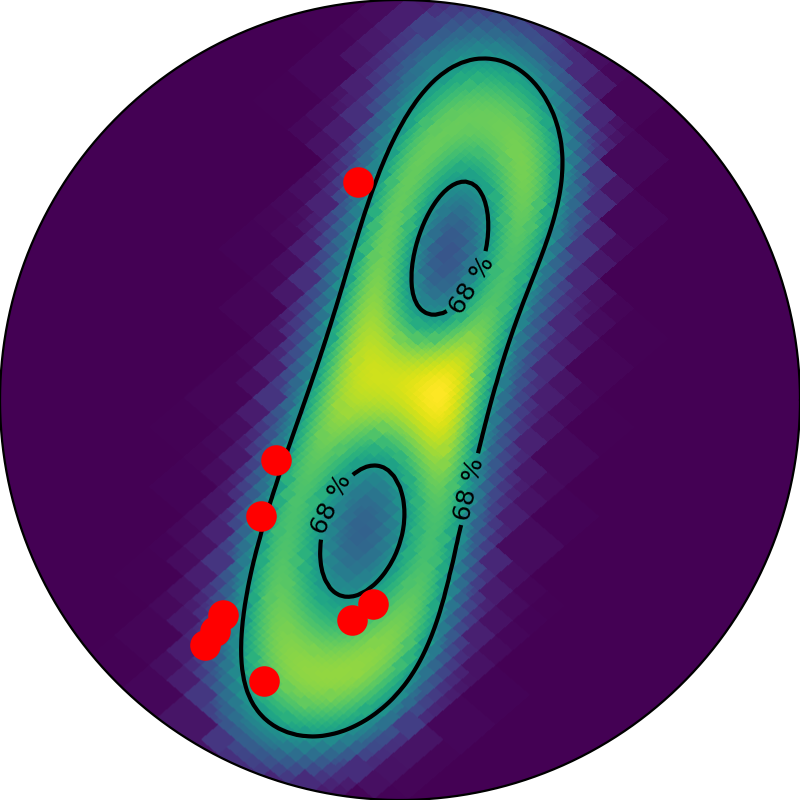} \\

      \end{tabular}
       \caption{Visual depiction of test results for the letters "A" and "B" using ZLP-Fisher flow, the
       Exponential map flow with radial basis function (EXP-R) and rational quadratic splines/Möbius flow (RQS-M). A ZLP-Kent addition as final layer is indicated with "+ K". The first column shows a Mollweide projection of the whole sphere, the others use orthographic projection. In the orthographic projections, the patch size is roughly the same as shown in the column header. Ten samples (red) from the test letter shape are shown for reference aswell.}
      \label{fig:visual_depiction_gain}
    \end{figure*}

\begin{figure}[htb!]
      \centering
      \includegraphics[width=0.49\textwidth]{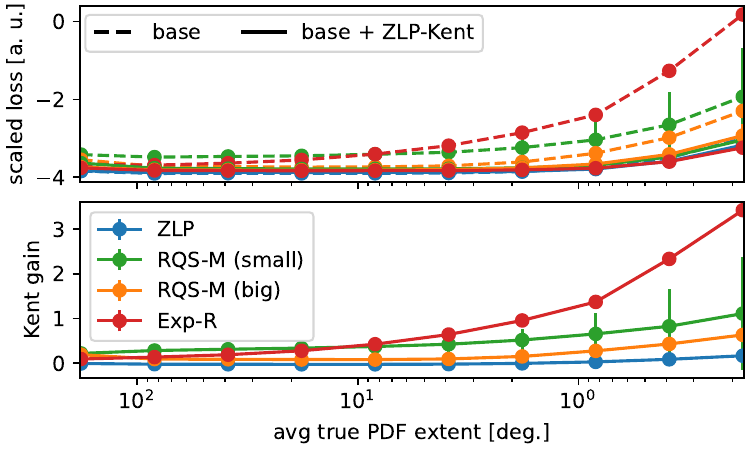}
      \caption{Improvement in test loss due to addition of a ZLP-Kent as the final flow layer. The x-axis shows PDF extents from 180 down to 0.2 degrees (zoom factors 1-1000). The upper plot shows a scaled test loss, where the entropy of a Gaussian of similar extent is subtracted to see the behavior more clearly.}
      \label{fig:ratio_gain}
\end{figure}
We can now ask what happens when we combine the "Fisher zoom" ($\phi_{\mathrm{Z}}$) and the "linear-project" ($\phi_{\mathrm{LP}}$) transformations in combination with rotations ($\phi_{\mathrm{R}}$). This is illustrated in table \ref{tab:equivalent_fb_distributions}. All the depicted flow functions are to be read from right to left, and their inverses from left to right. The vMF and central angular Gaussian, i.e. the Bingham-like density, are trivial since they correspond to the building blocks themselves. 
Using the second parametrization from last section involving $\Phi_{\mathrm{LP,S}}$, we can write the Bingham-like density in a second way that directly generalizes to the $\mathrm{FB}_{4}$-like case, which has a Fisher-zoom $\Phi_{\mathrm{Z}}$ inserted in the middle. It is therefore useful to think of the (bimodal)  $\mathrm{FB}_{4}$ as literally a zoomed-in version along a symmetry axis of the Bingham distribution. In the particular parametrization $\Phi_{\mathrm{FB}_4}(\vec{x})=[\Phi_{\mathrm{R}} \circ \Phi_\mathrm{Z} \circ \Phi_{\mathrm{LP,S}}](x)$, the zoom along a symmetry axis is always fulfilled. Strictly, the $\mathrm{FB}_{4}$ has no bimodal structure, so the bimodal extension we get by allowing free parameters is traditionally captured by the $\mathrm{FB}_{6}$, whose flow variant we describe later. If we do not want to allow for bimodal structure, or multimodal in higher dimensions, we have to further constrain $S_{ii}=\sigma_i \equiv \sigma$ for $i<D$, i.e. a shared scaling factor instead of free ones. Another important case is the $\mathrm{FB}_{5}$-like or Kent-like distribution, which we obtain by switching the order of the linear-project and Fisher zoom and adding appropriate constraints as described in the following theorem. 

\

\begin{theorem}
\label{theorem:kent_limit}[Properties of the ZLP-Fisher Kent flow]
    Let $\Phi_K(\vec{x})$ be
\begin{align}
\Phi_K(\vec{x})=[\Phi_{\mathrm{R}} \circ \Phi_{\mathrm{LP,S}} \circ \Phi_\mathrm{Z} ](x)
\end{align} 
the flow function, which yields a ZLP-Fisher flow on the D-1 sphere with zoom parameter $\kappa$ from the Fisher-zoom and linear-project diagonal matrix $S_{ii} = \sigma_i $ for $i=1 \ldots D-1$ and $S_{ii}=1$ for $i=D$. If $\frac{\sqrt{D}}{\sqrt{\kappa+D}} < \sigma_i < \frac{\sqrt{\kappa+D}}{\sqrt{D}}$ for all $i$ the resulting distribution has a single maximum and a single minimum at the opposite side of the sphere and we call it a "Kent-like" flow. It approximates a D-1 Gaussian distribution in the tangent space with standard deviations $\sigma_{i,t} = \frac{\sigma_i}{\sqrt{\kappa}}$ along its principal axes as $\kappa \rightarrow \infty$ and $\kappa \gg \sigma_i^2$ for all $i$.
\end{theorem}
Proof in appendix. For the 2-sphere, for example, it works well to parametrize a single variable $u$ to follow one side of the constraint and then set $\sigma_1=u$ and $\sigma_2=1/u$. We denote the constrained matrix $S$ that fulfills theorem \ref{theorem:kent_limit} as $S_c$ (see table \ref{tab:equivalent_fb_distributions}).
It turns out the order is crucial. Only by first applying the Fisher zoom followed by a diagonal and constrained linear-project step, do we get the right limiting properties that the distribution approaches a multi-variate normal distribution in tangent space for large $\kappa$. In the other order, these properties are not fulfilled. 
We can combine the $\mathrm{FB}_{4}$  and $\mathrm{FB}_{5}$ flows into an $\mathrm{FB}_{6}$ flow which involves two linear-project steps - one before and one after the zoom (see table \ref{tab:equivalent_fb_distributions}). This yields elliptical generalizations and bimodal densitites on the small circle. However, the first linear-project step and the Fisher zoom still share their symmetry axis. If we allow for an arbitrary first linear-project step, we obtain the full $\mathrm{FB}_{8}$, which includes asymmetries. Finally, since we have a normalizing flow, we can compose several linear-project and Fisher zooms together to yield more complex distributions.

\section{Tests on conditional density estimation}
\label{sec:conditional_density_test}

In the following we test the ZLP-Fisher flows in a conditional density estimation task. The dataset consists of samples drawn from letter shapes of small and capital alphabet letters, which are rotated and scaled on the upper half sphere. During generation of the data, samples are scaled with a factor $s$ between $0.001$ and $1$ and rotated to an arbitrary position on the half sphere,  with three parameters $\mu_1,\mu_2,\mu_3$ specifying the rotation. The input $C_{inp}$ to the conditional density is a 56-d vector, consisting of  a 52-one-hot encoding to determine the letter and 4 numbers specifying scaling and orientation.
\begin{equation}
    C_{inp}=(\underbrace{0,\ldots, 1,\ldots,0}_{52-\mathrm{d \ onehot}}, \mathrm{log}_{10}(s), \mu_1, \mu_2, \mu_3)
\end{equation}
An associated sample from such a scaled and rotated letter $i$ consists of the $x_i,y_i,z_i$ embedding coordinates on the sphere. We then train on negative log-probability as 
\begin{equation}
    L=-\frac{1}{N}\sum\limits_{i=1}^{N} \mathrm{ln}\left(p(x_i,y_i,z_i|C_{inp,i})\right)
\end{equation}
in batched training on 5 million random pairs of a specific letter $C_{inp,i}$ and corresponding samples ${x_i,y_i,z_i}$ to learn end-to-end the conditional PDF of all alphabetic letters with specific orientation and scaling. Training details and validation curves are given in the supplementary material. We compare the ZLP-Fisher flow, rational-quadratic splines with Möbius flows \cite{tori_and_spheres_paper} (large and small model), and exponential map flows with radial basis functions \cite{tori_and_spheres_paper}. We also train Kent-upgrades of these flows. If $\Phi_f(x)$ is a flow function to augment, and $\Phi_K(x)$ is a ZLP-Fisher flow of "Kent" type, the new flow function is then $\Phi_K(\Phi_f(x))$. The results are indicated in Fig. \ref{fig:visual_depiction_gain} for a qualitative comparison and in Fig. \ref{fig:ratio_gain} for a quantitative comparison. Overall, the ZLP-Fisher flow performs well in this setting compared to the alternatives. With a Kent upgrade all flows gain in absolute loss, especially at high zoom factors. The Exponential-map flow benefits the most, while the ZLP flow benefits the least since every ZLP layer contains a Fisher zoom and can in principle mimic a Kent, although we do not enforce a strict parametrization. The exponential map flow has an intrinsic size limit based on the number of flow layers and the RQS-M flow has an instability due to the poles, which makes them challenging to train on this dataset without a Kent upgrade.

\section{Related Work}
\label{sec:related_work}

In \cite{tori_and_spheres_paper}, the authors propose normalizing flows on the sphere involving exponential maps \cite{sei_gradient_map_paper} and using rational-quadratic splines \cite{neural_spline_flows} with conditional Möbius flows. We have used some of their proof results in proposition 3.1. and tested these flows in our toy study. They mention the von-Mises Fisher distribution in the context of "conditioning" approaches, and the angular Gaussian in the context of "projection" approaches, but they do not elaborate further since the general angular Gaussian is not tractable as a flow. They then conclude that in general flows on spheres with these approaches are only tractable in special cases, which makes them usually not very flexible. As we have shown, with the right parametrization and ordering of the vMF and central angular Gaussian flow analogues, we can actually not only generate complex flows, we can also form a Fisher-Bingham-like family of distributions which gives a starting point to build up more complexity as needed.

In neural posterior estimation \cite{npe_paper}, the task is to estimate a posterior from the real data and access to a simulator. Typical density estimators used here include mixture models \cite{npe_paper}, diffusion models \cite{npe_diffusion} or normalizing flows \cite{npe_flows}. Often, the approximations are refined in a sequential fashion, to finetune or to adapt to specific prior beliefs. Here, we are learning all posteriors at the same time without refinement and assume the prior in the simulation is the one we use, a form of amortized NPE. This is motivated by applications in astronomy, where it is sometimes not possible to do re-simulation per datum due to expensive simulators and high data rates. In this setting, we are specifically concerned with learning posteriors on the sphere at varying scales simultaneously - a task that we have not seen described in the literature before.

Existing normalizing flows on spheres usually have issues when used in the conditional setting. Neural spline flows paired with Möbius flows \cite{tori_and_spheres_paper} have numerical instability issues close to the poles. Exponential map flows \cite{sei_gradient_map_paper} with radial basis functions \cite{tori_and_spheres_paper}  are expressive, but usually can not describe arbitrarily localized regions - the barrier how localized depend on the number of flow layers. Continuous manifold normalizing flows \cite{cont_ode_paper} are already slow in the non-conditional setting - conditional continuous manifold flows are even slower. 

\section{Summary and Discussion}
\label{sec:summary}
We formulated how a Fisher-Bingham-like family of normalizing flows - the ZLP-Fisher family - can be constructed from two simple building blocks: the "Fisher-zoom", which corresponds to the flow that defines the vMF distribution, and the "linear-project" flow, which corresponds to the central angular Gaussian. The order of these building blocks matters, and together with specific parameter constraints defines which member of the family is obtained. For the Kent-like member, we showed that the limit of high concentration approximates a multivariate normal in tangent space, similar to the standard Kent distribution. Furthermore, we generalized the vMF diffeomorphism from D=3 to general D which allows to define the entire ZLP-Fisher family in any dimension. For generic D this requires numerical inverses in the forward and backward direction, which can in both cases be efficiently performed via newton iterations in logit space. For odd D the transformation admits finite-sum building blocks, reduces to the known 2-sphere result for D=3 and in tests is stable until at least $D=100$ and large concentrations $\kappa > \num{1e6}$. We leave it for future studies to test the $D \neq 3$  construction in detail, and focused here on the important case of the 2-sphere. In this context, we tested the ZLP-Fisher flow in a conditional density setting of approximating letter shapes on the 2-sphere which span 3 orders of magnitude in extent (from the whole sphere down to arcminutes) with various orientations. This mimics amortized NPE applications in astronomy where posterior scale differences can span orders of magnitudes. A complex ZLP-Fisher flow with many iterative flow layers behaves stable and can achieve good performance in combination with intermittent rotations. Furthermore, the Kent-like ZLP flow can upgrade any existing flow function as a final flow layer, which typically helps with performance at marginal extra cost. In the setting of orders of magnitudes of scale differences, this is a powerful flow upgrade to help with the first and second moment modeling at small scales. 
\section*{Acknowledgement}
We thank Tianlu Yuan for discussions related to the $\mathrm{FB}_8$ family.

\bibliographystyle{dinat}
\bibliography{main}

\begin{thebibliography}{24}
\makeatletter
\newcommand{\dinatlabel}[1]%
{\ifNAT@numbers\else\NAT@biblabelnum{#1}\hspace{2\labelsep}\fi}
\makeatother
\expandafter\ifx\csname natexlab\endcsname\relax\def\natexlab#1{#1}\fi
\expandafter\ifx\csname url\endcsname\relax\def\url#1{\texttt{#1}}\fi

\bibitem[Bingham(1974)]{bingham_distribution_paper}
\dinatlabel{Bingham 1974} \textsc{Bingham}, Christopher:
\newblock An Antipodally Symmetric Distribution on the Sphere.
\newblock In: \emph{The Annals of Statistics}
\newblock 2 (1974), Nr.~6, S.~1201--1225. --
\newblock URL \url{http://www.jstor.org/stable/2958339}. -- Zugriffsdatum: 2025-09-02. --
\newblock ISSN 00905364, 21688966

\bibitem[Bingham und Mardia(1978)]{fb4_paper}
\dinatlabel{Bingham und Mardia 1978} \textsc{Bingham}, Christopher~; \textsc{Mardia}, K.~V.:
\newblock A Small Circle Distribution on the Sphere.
\newblock In: \emph{Biometrika}
\newblock 65 (1978), Nr.~2, S.~379--389. --
\newblock URL \url{http://www.jstor.org/stable/2335218}. -- Zugriffsdatum: 2025-09-02. --
\newblock ISSN 00063444, 14643510

\bibitem[Davidson u.\,a.(2018)Davidson, Falorsi, Cao, Kipf und Tomczak]{hyperspherical_vaes_paper}
\dinatlabel{Davidson u.\,a. 2018} \textsc{Davidson}, Tim~R.~; \textsc{Falorsi}, Luca~; \textsc{Cao}, Nicola~D.~; \textsc{Kipf}, Thomas~; \textsc{Tomczak}, Jakub~M.:
\newblock \emph{Hyperspherical Variational Auto-Encoders}.
\newblock 2018. --
\newblock URL \url{https://arxiv.org/abs/1804.00891}

\bibitem[Durkan u.\,a.(2019)Durkan, Bekasov, Murray und Papamakarios]{neural_spline_flows}
\dinatlabel{Durkan u.\,a. 2019} \textsc{Durkan}, Conor~; \textsc{Bekasov}, Artur~; \textsc{Murray}, Iain~; \textsc{Papamakarios}, George:
\newblock Neural Spline Flows.
\newblock In: \emph{Advances in Neural Information Processing Systems} Bd.~32, Curran Associates, Inc., 2019. --
\newblock URL \url{https://proceedings.neurips.cc/paper_files/paper/2019/file/7ac71d433f282034e088473244df8c02-Paper.pdf}

\bibitem[{Fisher}(1953)]{vmf_distribution}
\dinatlabel{{Fisher} 1953} \textsc{{Fisher}}, Ronald:
\newblock {Dispersion on a Sphere}.
\newblock In: \emph{Proceedings of the Royal Society of London Series A}
\newblock 217 (1953), Mai, Nr.~1130, S.~295--305

\bibitem[Greenberg u.\,a.(2019)Greenberg, Nonnenmacher und Macke]{npe_flows}
\dinatlabel{Greenberg u.\,a. 2019} \textsc{Greenberg}, David~; \textsc{Nonnenmacher}, Marcel~; \textsc{Macke}, Jakob:
\newblock Automatic Posterior Transformation for Likelihood-Free Inference.
\newblock In: \emph{Proceedings of the 36th International Conference on Machine Learning} Bd.~97, PMLR, 09--15 Jun 2019, S.~2404--2414. --
\newblock URL \url{https://proceedings.mlr.press/v97/greenberg19a.html}

\bibitem[Hernandez-Stumpfhauser u.\,a.(2017)Hernandez-Stumpfhauser, Breidt und van~der Woerd]{generic_ag_paper}
\dinatlabel{Hernandez-Stumpfhauser u.\,a. 2017} \textsc{Hernandez-Stumpfhauser}, Daniel~; \textsc{Breidt}, F.~J.~; \textsc{Woerd}, Mark~J. van~der:
\newblock {The General Projected Normal Distribution of Arbitrary Dimension: Modeling and Bayesian Inference}.
\newblock In: \emph{Bayesian Analysis}
\newblock 12 (2017), Nr.~1, S.~113 -- 133. --
\newblock URL \url{https://doi.org/10.1214/15-BA989}

\bibitem[Jakob(2012)]{fisher_sampling}
\dinatlabel{Jakob 2012} \textsc{Jakob}, Wenzel:
\newblock Numerically stable sampling of the {von Mises Fisher} distribution on $\mathbb{S}^2$ (and other tricks).
\newblock In: \emph{EPFL Technical Report}
\newblock (2012). --
\newblock URL \url{https://infoscience.epfl.ch/handle/20.500.14299/128111}

\bibitem[Kent(1982)]{fb5_paper}
\dinatlabel{Kent 1982} \textsc{Kent}, John~T.:
\newblock The Fisher-Bingham Distribution on the Sphere.
\newblock In: \emph{Journal of the Royal Statistical Society. Series B (Methodological)}
\newblock 44 (1982), Nr.~1, S.~71--80. --
\newblock URL \url{http://www.jstor.org/stable/2984712}. -- Zugriffsdatum: 2025-09-02. --
\newblock ISSN 00359246

\bibitem[Kingma und Welling(2014)]{vae_paper}
\dinatlabel{Kingma und Welling 2014} \textsc{Kingma}, D.P.~; \textsc{Welling}, Max:
\newblock {Auto-Encoding Variational Bayes}.
\newblock In: \emph{International Conference on Learning Representations}, 2014
\newblock (ICLR)

\bibitem[Kingma und Dhariwal(2018)]{glow_paper}
\dinatlabel{Kingma und Dhariwal 2018} \textsc{Kingma}, Durk~P.~; \textsc{Dhariwal}, Prafulla:
\newblock Glow: Generative Flow with Invertible 1x1 Convolutions.
\newblock In: \emph{Advances in Neural Information Processing Systems} Bd.~31, Curran Associates, Inc., 2018. --
\newblock URL \url{https://proceedings.neurips.cc/paper_files/paper/2018/file/d139db6a236200b21cc7f752979132d0-Paper.pdf}

\bibitem[Lou u.\,a.(2020)Lou, Lim, Katsman, Huang, Jiang, Lim und De~Sa]{cont_ode_paper}
\dinatlabel{Lou u.\,a. 2020} \textsc{Lou}, Aaron~; \textsc{Lim}, Derek~; \textsc{Katsman}, Isay~; \textsc{Huang}, Leo~; \textsc{Jiang}, Qingxuan~; \textsc{Lim}, Ser-Nam~; \textsc{De~Sa}, Christopher:
\newblock Neural manifold ordinary differential equations.
\newblock In: \emph{Proceedings of the 34th International Conference on Neural Information Processing Systems}.
\newblock Red Hook, NY, USA~: Curran Associates Inc., 2020
\newblock (NIPS '20). --
\newblock ISBN 9781713829546

\bibitem[Mardia(1975)]{mardia_fb8_1975}
\dinatlabel{Mardia 1975} \textsc{Mardia}, K.~V.:
\newblock Statistics of Directional Data.
\newblock In: \emph{Journal of the Royal Statistical Society. Series B (Methodological)}
\newblock 37 (1975), Nr.~3, S.~349--393. --
\newblock URL \url{http://www.jstor.org/stable/2984782}. -- Zugriffsdatum: 2025-09-15. --
\newblock ISSN 00359246

\bibitem[Paine u.\,a.(2018)Paine, Preston, Tsagris und Wood]{ag_kent_paper}
\dinatlabel{Paine u.\,a. 2018} \textsc{Paine}, P.~J.~; \textsc{Preston}, S.~P.~; \textsc{Tsagris}, M.~; \textsc{Wood}, Andrew T.~A.:
\newblock An elliptically symmetric angular Gaussian distribution.
\newblock In: \emph{Statistics and Computing}
\newblock 28 (2018), May, Nr.~3, S.~689--697. --
\newblock URL \url{https://doi.org/10.1007/s11222-017-9756-4}. --
\newblock ISSN 1573-1375

\bibitem[Papamakarios und Murray(2016)]{npe_paper}
\dinatlabel{Papamakarios und Murray 2016} \textsc{Papamakarios}, George~; \textsc{Murray}, Iain:
\newblock Fast $\varepsilon$-free inference of simulation models with Bayesian conditional density estimation.
\newblock In: \emph{Proceedings of the 30th International Conference on Neural Information Processing Systems}.
\newblock Red Hook, NY, USA~: Curran Associates Inc., 2016
\newblock (NIPS'16), S.~1036–1044. --
\newblock ISBN 9781510838819

\bibitem[Papamakarios u.\,a.(2021)Papamakarios, Nalisnick, Rezende, Mohamed und Lakshminarayanan]{flows_paper}
\dinatlabel{Papamakarios u.\,a. 2021} \textsc{Papamakarios}, George~; \textsc{Nalisnick}, Eric~; \textsc{Rezende}, Danilo~J.~; \textsc{Mohamed}, Shakir~; \textsc{Lakshminarayanan}, Balaji:
\newblock Normalizing Flows for Probabilistic Modeling and Inference.
\newblock In: \emph{Journal of Machine Learning Research}
\newblock 22 (2021), Nr.~57, S.~1--64. --
\newblock URL \url{http://jmlr.org/papers/v22/19-1028.html}

\bibitem[Rezende u.\,a.(2020)Rezende, Papamakarios, Racaniere, Albergo, Kanwar, Shanahan und Cranmer]{tori_and_spheres_paper}
\dinatlabel{Rezende u.\,a. 2020} \textsc{Rezende}, Danilo~J.~; \textsc{Papamakarios}, George~; \textsc{Racaniere}, Sebastien~; \textsc{Albergo}, Michael~; \textsc{Kanwar}, Gurtej~; \textsc{Shanahan}, Phiala~; \textsc{Cranmer}, Kyle:
\newblock Normalizing Flows on Tori and Spheres.
\newblock In: \emph{International Conference on Machine Learning} Bd.~119, 13--18 Jul 2020, S.~8083--8092

\bibitem[Rivest(1984)]{rivest_fb6}
\dinatlabel{Rivest 1984} \textsc{Rivest}, Louis-Paul:
\newblock On the Information Matrix for Symmetric Distributions on the Hypersphere.
\newblock In: \emph{The Annals of Statistics}
\newblock 12 (1984), Nr.~3, S.~1085--1089. --
\newblock URL \url{http://www.jstor.org/stable/2240982}. -- Zugriffsdatum: 2025-09-15. --
\newblock ISSN 00905364, 21688966

\bibitem[Sei(2013)]{sei_gradient_map_paper}
\dinatlabel{Sei 2013} \textsc{Sei}, Tomonari:
\newblock A Jacobian Inequality for Gradient Maps on the Sphere and Its Application to Directional Statistics.
\newblock In: \emph{Communications in Statistics - Theory and Methods}
\newblock 42 (2013), Nr.~14, S.~2525--2542

\bibitem[Sharrock u.\,a.(2024)Sharrock, Simons, Liu und Beaumont]{npe_diffusion}
\dinatlabel{Sharrock u.\,a. 2024} \textsc{Sharrock}, Louis~; \textsc{Simons}, Jack~; \textsc{Liu}, Song~; \textsc{Beaumont}, Mark:
\newblock Sequential Neural Score Estimation: Likelihood-Free Inference with Conditional Score Based Diffusion Models.
\newblock In: \emph{Proceedings of the 41st International Conference on Machine Learning} Bd.~235, PMLR, 21--27 Jul 2024, S.~44565--44602. --
\newblock URL \url{https://proceedings.mlr.press/v235/sharrock24a.html}

\bibitem[Tyler(1987)]{ag_bingham}
\dinatlabel{Tyler 1987} \textsc{Tyler}, David~E.:
\newblock Statistical Analysis for the Angular Central Gaussian Distribution on the Sphere.
\newblock In: \emph{Biometrika}
\newblock 74 (1987), Nr.~3, S.~579--589. --
\newblock URL \url{http://www.jstor.org/stable/2336697}. -- Zugriffsdatum: 2025-09-02. --
\newblock ISSN 00063444

\bibitem[Watson(1965)]{watson_distribution_great_dircle}
\dinatlabel{Watson 1965} \textsc{Watson}, G.~S.:
\newblock Equatorial Distributions on a Sphere.
\newblock In: \emph{Biometrika}
\newblock 52 (1965), Nr.~1/2, S.~193--201. --
\newblock URL \url{http://www.jstor.org/stable/2333824}. -- Zugriffsdatum: 2025-09-02. --
\newblock ISSN 00063444, 14643510

\bibitem[Watson(1983)]{watson_book}
\dinatlabel{Watson 1983} \textsc{Watson}, Geoffrey~S.:
\newblock \emph{Statistics on Spheres}.
\newblock Wiley, 1983

\bibitem[Yuan(2021)]{fb8_paper}
\dinatlabel{Yuan 2021} \textsc{Yuan}, Tianlu:
\newblock The 8-parameter Fisher--Bingham distribution on the sphere.
\newblock In: \emph{Computational Statistics}
\newblock 36 (2021), Mar, Nr.~1, S.~409--420. --
\newblock URL \url{https://doi.org/10.1007/s00180-020-01023-w}. --
\newblock ISSN 1613-9658

\end{thebibliography}

\end{document}